\pdfoutput=1
\documentclass[11pt]{article}

\usepackage[final]{acl}

\usepackage{times}
\usepackage{latexsym}

\usepackage[T1]{fontenc}
\usepackage[utf8]{inputenc}

\usepackage{microtype}

\usepackage{inconsolata}

\usepackage{framed}

\usepackage{graphicx}

\usepackage{url}            % simple URL typesetting
\usepackage{float}
\usepackage{placeins}
\usepackage{booktabs}       % professional-quality tables
\usepackage{array}
\usepackage{rotating}
\usepackage{multirow}
\usepackage{amsfonts, amsmath, amsthm, amssymb, mathtools} % math
\usepackage{subfigure}
\usepackage{wrapfig}
\usepackage{nicefrac}       % compact symbols for 1/2, etc.
\usepackage[most]{tcolorbox} % The 'many' library is loaded for additional options
\usepackage{xcolor}       % colors
\usepackage[inline]{enumitem}
\usepackage{algorithm}
\usepackage{algpseudocode}
\usepackage{pgfplots}
\usepackage{svg}
\pgfplotsset{compat=1.17}

\definecolor{customblue}{HTML}{1F77B4}
\definecolor{customdarkblue}{HTML}{0000FF}
\definecolor{customdarkred}{HTML}{FF0000}
\definecolor{customorange}{HTML}{FF7F0E}
\definecolor{customgreen}{HTML}{2CA02C}
\definecolor{custom_green}{HTML}{2CA02C}
\definecolor{custom_brown}{HTML}{8C564B}
\definecolor{custom_purple}{HTML}{9467BD}
\definecolor{custom_red}{HTML}{D62728}

\newcommand{\revision}[1]{\textcolor{black}{#1}}

\newcommand{\blue}{\raisebox{2pt}{\tikz{\draw[customblue, solid, line width=2.3pt](0,0) -- (5mm,0);}}}
\newcommand{\orange}{\raisebox{2pt}{\tikz{\draw[customorange, solid, line width=2.3pt](0,0) -- (5mm,0);}}}
\newcommand{\green}{\raisebox{2pt}{\tikz{\draw[custom_green, solid, line width=2.3pt](0,0) -- (5mm,0);}}}
\newcommand{\brown}{\raisebox{2pt}{\tikz{\draw[custom_brown, solid, line width=2.3pt](0,0) -- (5mm,0);}}}
\newcommand{\red}{\raisebox{2pt}{\tikz{\draw[custom_red, solid, line width=2.3pt](0,0) -- (5mm,0);}}}
\newcommand{\purple}{\raisebox{2pt}{\tikz{\draw[custom_purple, solid, line width=2.3pt](0,0) -- (5mm,0);}}}

\newlength{\myheight}
\newcommand{\coloredsquare}[1]{%
  \settoheight{\myheight}{X}% You can replace 'X' with any text whose height you want to match.
  \tikz \fill [#1] (0,0) rectangle (\myheight,\myheight);%
}

\newtcbox{\highlight}[1][customblue]{
    on line,
    arc=0pt, % Set to 0pt for sharp corners
    colback=#1!35!white,
    colframe=#1!35!white,
    boxsep=0pt,
    left=1pt,
    right=1pt,
    top=2pt,
    bottom=2pt,
    boxrule=0pt,
    nobeforeafter
}

\author{Baturay~Saglam\thanks{Corresponding author: \texttt{baturay.saglam@yale.edu}}, Xinyang~Hu, Zhuoran~Yang, \\ {\bf Dionysis~Kalogerias,} {\bf Amin~Karbasi} \\ Yale University}

\title{Learning Task Representations from In-Context Learning}

\begin{document}
\maketitle
\begin{abstract}
    Large language models (LLMs) have demonstrated remarkable proficiency in in-context learning (ICL), where models adapt to new tasks through example-based prompts without requiring parameter updates. However, understanding how tasks are internally encoded and generalized remains a challenge. To address some of the empirical and technical gaps in the literature, we introduce an automated formulation for encoding task information in ICL prompts as a function of attention heads within the transformer architecture. This approach computes a single task vector as a weighted sum of attention heads, with the weights optimized causally via gradient descent. Our findings show that existing methods fail to generalize effectively to modalities beyond text. In response, we also design a benchmark to evaluate whether a task vector can preserve task fidelity in functional regression tasks. The proposed method successfully extracts task-specific information from in-context demonstrations and excels in both text and regression tasks, demonstrating its generalizability across modalities.
\end{abstract}

\section{Introduction}
\label{sec:intro}
Large language models (LLMs) based on the transformer architecture \citep{transformer} have seen dramatic improvements in recent years. A notable feature of these models, such as GPT-3 \citep{gpt3}, is their capability for \emph{in-context learning} (ICL). This process involves the model receiving a prompt that includes demonstrations of a task in the form of input-output pairs. When presented with a new query input, the model can generate the appropriate output by extrapolating from the provided examples. For instance, after being prompted with a few examples, these models are capable of producing the antonyms of given input words. A concrete example is 
\begin{equation}
\label{eq:icl_language_example}
\text{
\small$
    \underbrace{{\color{blue}\text{vanish} \rightarrow \text{appear}, \text{ short} \rightarrow \text{tall},}}_{\text{examples}} \underbrace{{\color{blue}\text{increase} \rightarrow}}_{\text{query}} \underbrace{~{\color{red}\text{decrease}}}_{\text{completion}}$
\normalsize},
\end{equation}
where the blue text is the \textcolor{blue}{prompt} and the red text is the \textcolor{red}{completion} provided by the model. Not limited to linguistic tasks, transformers can also in-context learn a general class of functions \( \mathcal{F} \) \citep{garg_icl}. Specifically, for any function \( f \in \mathcal{F} \), the model is capable of approximating \( f(x_\text{query}) \) for a new query input \( x_\text{query} \). This class may include linear or nonlinear relationships, which can be represented by various machine learning models such as linear regression, multi-layer ReLU networks, and decision trees. 

The ICL capability is particularly intriguing as it allows models to adapt to a wide range of downstream tasks on-the-fly—\emph{i.e.}, without requiring any parameter updates post-training \citep{gpt3}. This indicates that models can extract abstract task information (the relationship between input-output pairs) from the prompt and use it to generate accurate responses for query inputs. However, due to the complex nature of the transformer architecture, the computational mechanisms that facilitate how models internally extract and encode task information are not well understood.

\textit{Task vectors} have been identified as key mechanisms for encoding task information during ICL \citep{fv, dongfang, icl_creates_task_vectors, icv}. These vectors capture how models process prompts and embed the abstract in-context task as a numerical representation. While the effectiveness of these vectors is empirically well-supported, the precise computation of a task vector within the transformer architecture remains debated. Early findings suggest that layer or attention activations significantly influence ICL performance \citep{icl_creates_task_vectors, icv, fv}. 

\revision{However, there is no consensus on the optimal way to conceptualize or compute these vectors, resulting in varied methodologies that often excel in linguistics but fail to generalize to other tasks, such as functional regression. Since these methods rely on the internal components of transformers and architectural differences due to output modality (e.g., language head or numeric outputs) do not change their operation, any meaningful task representation should, in principle, be agnostic to output modality. Poor non-language performance therefore suggests that they may be capturing modality-specific patterns rather than a generalizable task representation.} Additionally, these methods require adaptation to individual transformer architectures. For example, the layers over which the task vector is computed must be explicitly identified for each model \citep{dongfang, fv, icl_creates_task_vectors}.

These limitations raise questions about the generalizability and practical utility of existing task vector formulations. To pursue a more standardized and automated pipeline for task encodings, \revision{we build on the recent finding that attention heads are the primary driver for vectorizing ICL tasks and hypothesize that:}
\begin{center}
    \begin{framed}
    \emph{In-context learning tasks can be effectively represented as a weighted sum of all attention heads in a transformer, with the weights learned causally through gradient descent.}
    \end{framed}
\end{center}

\subsection{Contributions and Findings}
\label{sec:contributions}

The key contributions and findings of our study are outlined as follows.

\paragraph{An Automated Task Formulation Independent of Modality and Architecture} 
We propose a framework that approximates the internal weights transformers implicitly use to represent ICL tasks. By optimizing these weights through causal gradient descent, the framework effectively captures the model’s inherent task representations, supporting our hypothesis. Unlike existing methods that require prior model analysis—such as manually selecting hidden layers to perturb—our approach is fully automated. Given any dataset and autoregressive model, it seamlessly computes the task formulation without model-specific adjustments.

\paragraph{A New Benchmark for Regression Tasks}  
While prior work has explored where in-context tasks are represented, most studies focus solely on language, with limited attention to other modalities like functional regression. We introduce a new benchmark to assess task encodings in regression. This setup provides an intuitive framework for analyzing task encodings and ICL capabilities in regression by leveraging an out-of-distribution condition.

\paragraph{Performance Gains}  
Our method shows effective task encoding on the proposed regression benchmark and language datasets. Ablation studies indicate that this performance stems from aligning the distribution of the last hidden state with that of an optimally in-context-learned model, offering insights into how task steering is achieved.

For reproducibility, we provide the code in our GitHub repository\footnote{\url{https://github.com/baturaysaglam/ICL-task-repr}}.

\section{Related Work}
\label{sec:related_work_task_repr}
Here, we discuss studies that specifically focus on learning task representations in ICL. A more comprehensive review of related works is available in Appendix \ref{app:related_work}.

Initial studies on developing task representations for transformers are documented by \citet{lampinen_task_repr, shao_2023, mu_2023, panigrahi_2023, ilharco_2023}. These works proposed methods to create compositional task encodings through model perturbations in the parameter space: ``soft'' prompts, codebooks, and meta-mappings. Notably, the term \emph{task vectors} was first introduced by \citet{ilharco_2023}. In contrast, through the application of causal mediation analysis \citep{direct_and_indirect_effects, vig, meng_factual_assoc, wang_2023a, geva_2023}, \textit{function vectors} were discovered to exist inherently within the transformer architecture and to exhibit strong causal effects \citep{fv}. This finding aligns with research on RNNs, where it was demonstrated that RNN hidden states can be grouped based on task similarities \citep{lake_2018, hill_2018}.

Efforts to represent ICL tasks often derive a task vector from the layer activations associated with the dummy token following the input query \citep{icl_creates_task_vectors}. This approach was later refined by averaging the layer activations of dummy tokens across a few-shot prompt and optimizing them \citep{dongfang}. However, these methods were primarily designed for language tasks, where input-output pairs are explicitly separated by specific tokens (\emph{e.g.}, ``\(\rightarrow\)''). This limitation restricts their applicability to broader domains, such as regression tasks. Another approach leverages the first principal component of the difference in layer activations to guide the ICL task, resulting in \textit{In-Context Vectors} (ICVs) \citep{icv}. Notably, attention heads have been argued to play a critical role in transferring information between token positions \citep{transformer, math_of_transformers}. \revision{Similarly, \citet{fv} found that attention heads are primarily responsible for encoding ICL behavior. Their proposed formulation, \textit{Function Vector} (FV), is computed as the sum of activations from a specific subset of attention heads, selected based on an \emph{indirect metric} derived from causal inference literature \citep{direct_and_indirect_effects}.} In this study, we aim to develop a structured and automated method for extracting function vectors from transformer architectures, expanding on the approach of \citet{fv}.

\section{Technical Preliminaries}
\revision{We focus on decoder-only autoregressive transformer architectures.}

\subsection{Transformer Architecture}
The transformer architecture uses self-attention mechanisms to process sequences of data. Initially, input data is tokenized into a sequence, where each token represents data units such as segments of words. In this work, we consider autoregressive transformers denoted by \(M_\theta\) and parameterized by \(\theta\). The model predicts the next element in a sequence based on previous outputs. It consists of \(L\) layers, each transforming encoded token vectors of dimension \(d\) through linear and nonlinear operations. Our focus is on the computation at the last token position within these layers, where each layer \( \ell \leq L\) generates a vector representation \( \mathbf{h}_\ell \in \mathbb{R}^d \) from its preceding layer’s output.

Self-attention in the transformer architecture employs multi-head attention at each layer:
\begin{multline*}
    \operatorname{MultiHead}_\ell(Q_\ell, K_\ell, V_\ell) = \operatorname{Concat}(\operatorname{head}_{\ell, 1}, \\
    \ldots, \operatorname{head}_{\ell, J})W_\ell^O,
\end{multline*}
where \(\mathbb{R}^{q} \ni \operatorname{head}_{\ell, j} \coloneqq \operatorname{Attn}_{\ell, j} = \operatorname{softmax}\big(Q_{\ell, j} K^\top_{\ell, j } / \sqrt{d_k} ) V_{\ell, j}\). Here \(W_\ell^O \in \mathbb{R}^{Jq \times d}\) is the output projection matrix, and \(Q_{\ell, j}\), \(K_{\ell, j}\), and \(V_{\ell, j}\) are the query, key, and value matrices for each attention head \(j \leq J\) at layer \(\ell\). The term \(\sqrt{d_k}\) normalizes the softmax operation for stability, where \(d_k\) is the dimension of the key matrix. This multi-head approach allows the model to dynamically adjust its focus across different parts of the input based on the context.

Note that each head at layer $\ell$ operates in a low-dimensional space $q < d$, distinct from the main hidden state residual stream of the transformer. As observed by \citet{transformer_circuits}, the original transformer formulation \citep{transformer} can be seen as dividing the matrix into a block form $[W_{\ell, 1}, \ldots, W_{\ell, J}]$, followed by directly projecting each $\operatorname{head}{\ell, j}$ into the residual space. Consequently, the attention head output $a{\ell, j}$ can be defined using \citeposs{transformer} and \citeposs{transformer_circuits} notations as:
\begin{equation*}
    a_{\ell, j} = \operatorname{head}_{\ell, j}(Q_{\ell, j}, K_{\ell, j}, V_{\ell, j})W_{\ell, j}^O \in \mathbb{R}^d.
\end{equation*}

\subsection{In-Context Learning} 
\label{sec:background_icl}
A prompt \( p^t \), corresponding to task \( t \), comprises a sequence of tokens including \( T \) input-output exemplar pairs \(\{(x_i, y_i)\}_{i=0}^T\). We may refer to the length of prompts by the number of demonstrations they contain. Each pair demonstrates the execution of the same underlying task \( t \). This set defines a functional mapping between each input \( x_i \) and its corresponding output \( y_i \). In addition to these exemplar pairs, each prompt includes a specific query input \(x_{\text{query}}\) that follows the demonstrations. The LM predicts the target response \(y_{\text{query}} \) by extrapolating from the demonstrations. In our study, we investigate the in-context learning of two distinct modalities: functional regression and language tasks.

\paragraph{Functional Regression Tasks}
\emph{Regression tasks} refer to a function class \(\mathcal{F}\) that is learned through ICL. We closely follow the formulation proposed in \citep{garg_icl} for training and testing models on regression data. For each prompt, a random function \(f\) from \(\mathcal{F}\) is sampled according to a distribution \(\mathcal{D}_\mathcal{F}\), and a set of random inputs \(x_i \in \mathbb{R}^m\) for \(i = 1, \ldots, T\) is drawn from \(\mathcal{D}_\mathcal{X}\). These inputs are then evaluated by \(f\) to produce the prompt \(p^f = \{x_1, f(x_1), \ldots, x_T, f(x_T), x_\text{query}\}\). We denote the model's output as \(M_{\theta}(p^f)\).

A function class \(\mathcal{F}\) can be either linear or nonlinear. For linear functions, inputs are drawn from an isotropic Gaussian distribution \(\mathcal{N}(0, I_m)\), and a random function is defined by sampling a weight vector \(w\) from \(\mathcal{N}(0, I_m)\), setting \(f(x) = w^\top x\). Here, the task is characterized by the weight vector \(w \in \mathbb{R}^m\). For nonlinear functions, possible forms of \(f\) include multi-layer ReLU networks or decision trees. We employ the models pre-trained by \citet{garg_icl}, with the training procedure and additional details provided in Appendix \ref{app:model_training_synth}.

\paragraph{Language Tasks}
We focus on straightforward NLP applications such as antonym generation, with an example shown in \eqref{eq:icl_language_example}. The model \(M_\theta\) processes each prompt \(p^t\) and produces a next-token distribution \(M_\theta(\cdot \mid p^t)\) over the vocabulary \(\mathcal{V}\). In our evaluation of task representations, we follow \citet{fv}, who \emph{corrupted} prompts \(\Tilde{p}_i^t\) by shuffling labels within each prompt. This shuffling disrupts the connection between inputs \(x_{i, k}\) and outputs \(\Tilde{y}_{i, k}\), rendering the examples uninformative about the task and effectively ``blocking'' ICL.

\begin{figure} % Right wrap, half of the text width
    \centering
    \begin{equation*}
        \text{\small {\blue} Transformer}  \quad \text{\small {\green} Trans. +FV} \quad \text{\small {\purple} Trans. +ICV}
    \end{equation*}
	\subfigure[Linear]{
            \includegraphics[width=0.31\linewidth]{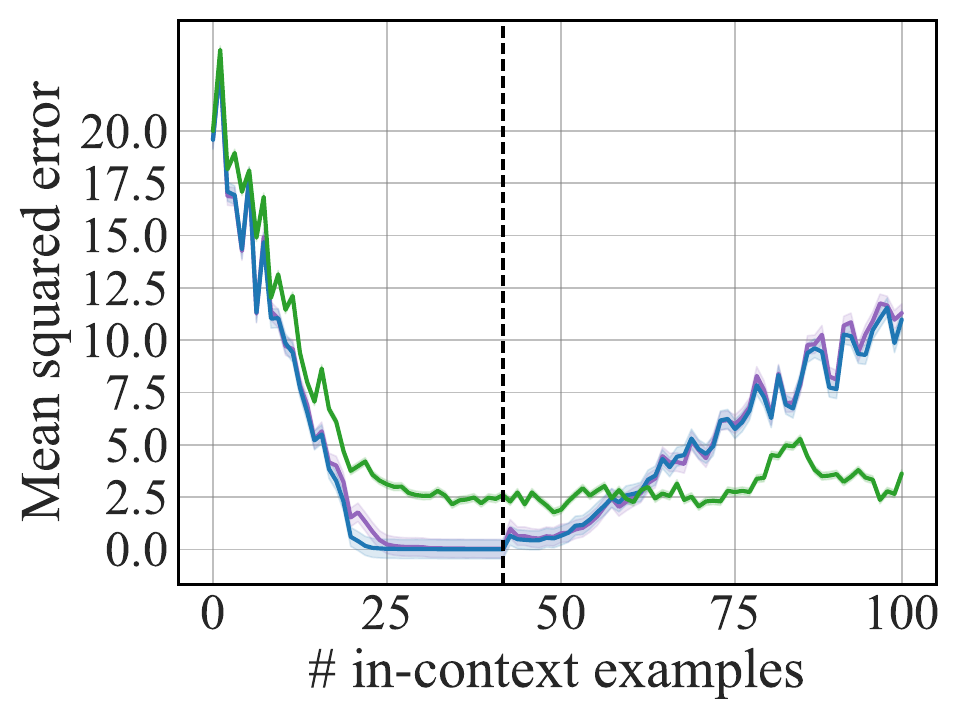}%
        }
        \subfigure[Sparse linear]{
            \includegraphics[width=0.31\linewidth]{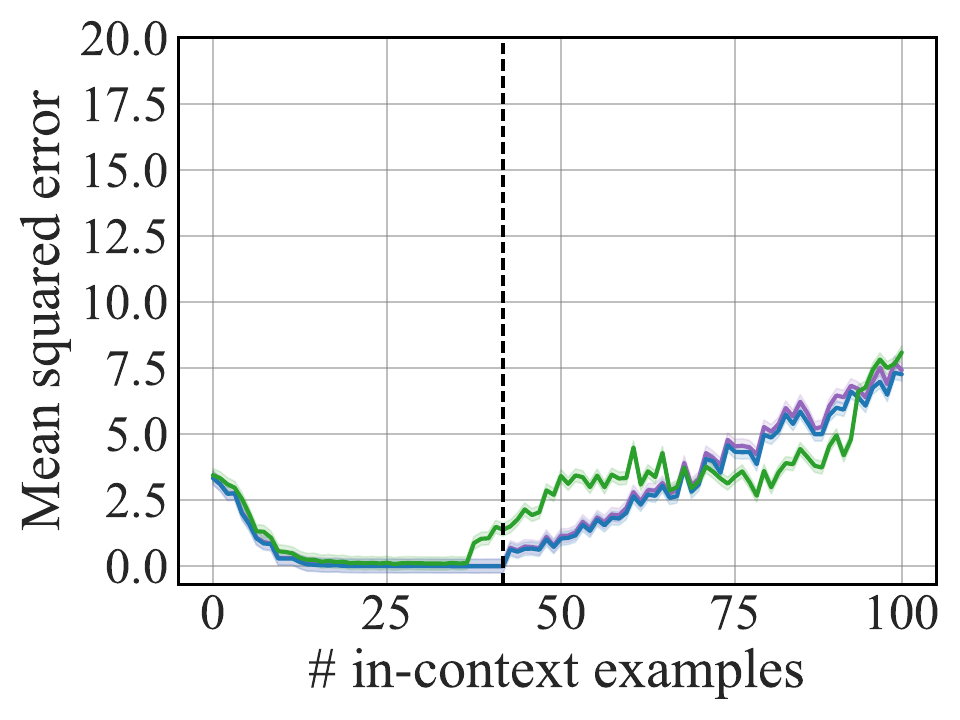}%
        }
        \subfigure[2-layer NN]{
            \includegraphics[width=0.31\linewidth]{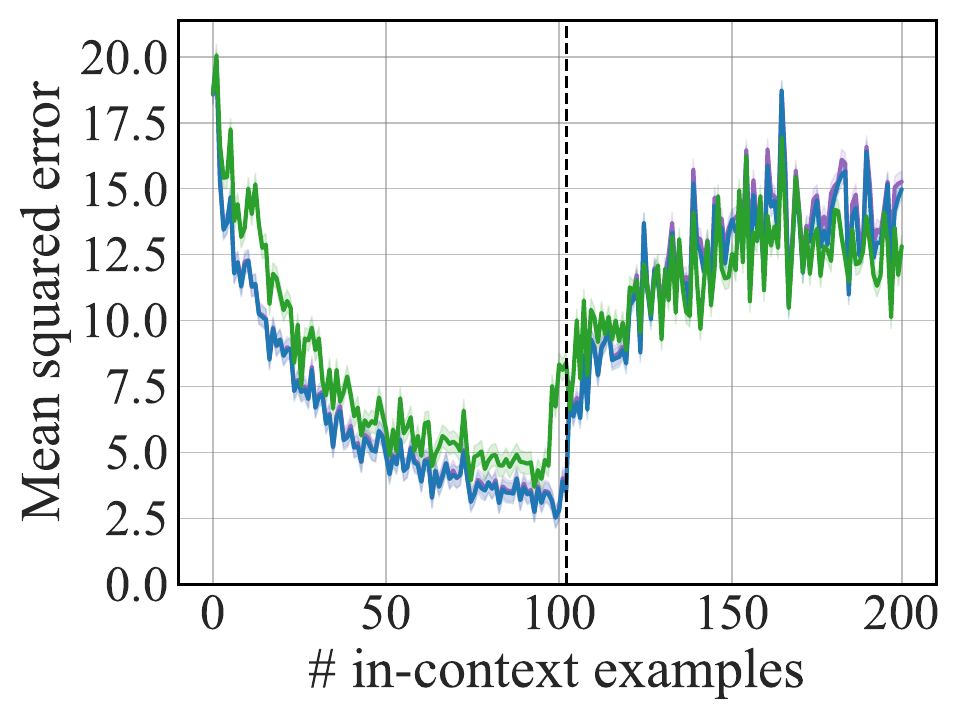}%
        }
\caption{Squared error on the query input as a function of the number of demonstrations in the prompts, where lower error \((f(x_{\text{query}}) - M_{\theta}(x_{\text{query}} \mid p^f))^2\) indicates better ICL performance. We evaluate three different function classes: (a) linear functions \(f(x_\text{query}) = w^\top x_\text{query}\), (b) sparse linear functions \(f(x_\text{query}) = w_s^\top x\), and (c) 2-layer ReLU neural networks (NNs) \(f(x_\text{query}) = W_2 \operatorname{ReLU}(W_1 x_\text{query})\). Results are averaged over a batch of 256 tasks randomly selected from the same function class. The shaded area represents the 95\% confidence interval over the sampled prompts. The dashed line indicates the maximum sequence length used during the positional encoding training, \(T_\text{train}\).}
\label{fig:transformers_failure}
\end{figure}

\section{Learning Task Representations}

\subsection{Motivational Observations}
\label{sec:motiv_obs}

Our study begins by noting that existing ICL task formulations are either inapplicable \citep{dongfang, icl_creates_task_vectors} (see Appendix \ref{app:other_baselines}) or fail to generalize to regression tasks \citep{icv, fv}. To demonstrate this limitation, we design a new benchmark for functional regression.

\revision{
The motivation for this benchmark is straightforward and follows standard practices from prior work on language tasks (see Section \ref{sec:background_icl}). We first prevent the model from extracting task understanding from in-context examples. If ICL performance is restored when the task vector is integrated into the model, it suggests that the task vector encodes the task information. However, techniques like shuffling labels or probing are unsuitable for regression tasks due to fundamental differences in how input-output relationships are represented and learned (see Appendices \ref{app:stop_icl_shuffling} and \ref{app:probing_ineffective}).
}

\revision{To suppress task-relevant information in regression prompts, we introduce an out-of-distribution (OOD) case. The model is initialized and pre-trained to handle sequences up to a maximum length \(T_\text{max}\), ensuring it can process longer sequences. However, during pre-training, the model is exposed only to sequences of length up to \(T_\text{train} < T_\text{max}\), leaving positional encodings for positions beyond \(T_\text{train}\) untrained.}

\revision{At test time, we evaluate the model on sequences longer than \(T_\text{train}\) but within \(T_\text{max}\). The focus is not on downstream accuracy but on robustness under OOD conditions. Since positional encodings beyond \(T_\text{train}\) are untrained, the model cannot infer where the input and outputs are located within the prompt, and thus cannot extract the desired input-output relationship any further—the importance of positional information in ICL has also been considered by \citet{pe_icl}. However, if a representation contains task-relevant information (specific to the ideal function \(f\)), it should maintain functional task behavior and generalize to prompts longer than \(T_\text{train}\).}

\revision{The results, shown in Figure \ref{fig:transformers_failure}, reveal that the tested approaches fail to generalize across the evaluated class of functions, despite decent performance in the language domain. Because these methods rely on the internal components of transformers and are unaffected by architectural differences related to output modality (e.g., language heads or numeric outputs), any meaningful task representation should, in principle, remain consistent across output modalities.}

\begin{figure*}[!htbp]
  \centering
  \includegraphics[width=\linewidth]{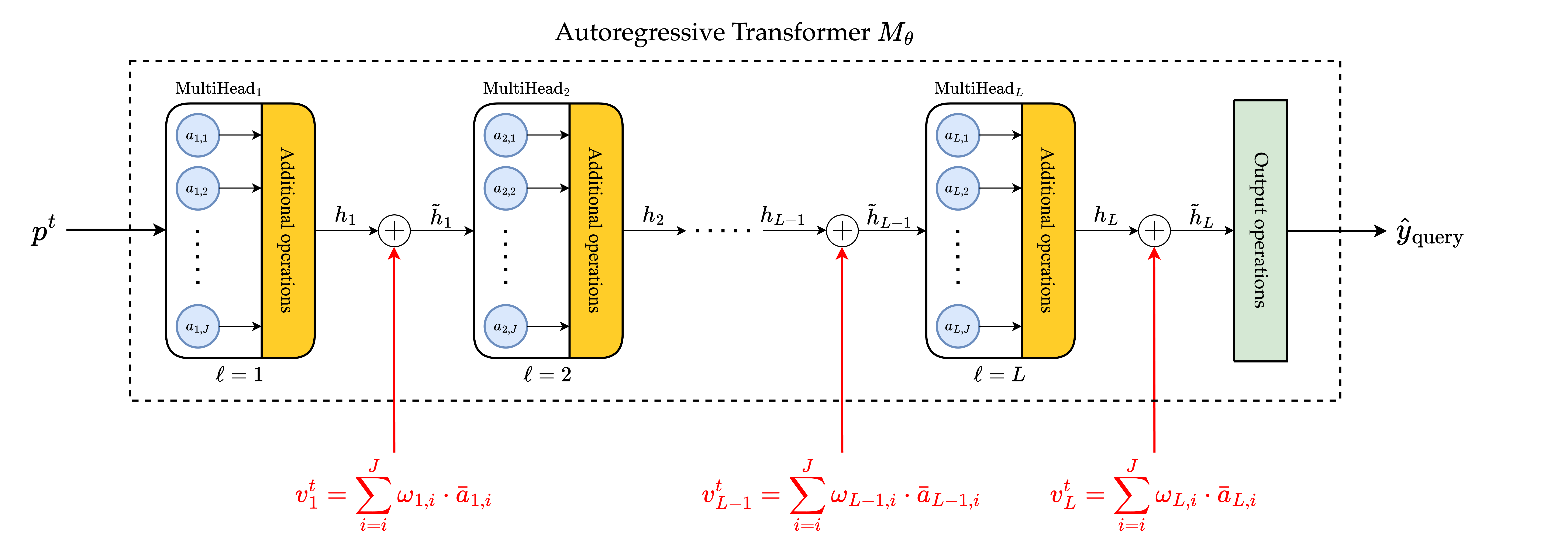}
    \caption{Illustration of the operation. \textit{Additional} and \textit{output} operations may include residual connections, normalization, feedforward, or prediction layers, depending on the architecture. LTV is added sequentially to each layer, allowing the effects of the integrated LTV to be progressively observed across subsequent layers.}
  \label{fig:ltv_inference}
\end{figure*}

As our investigation deepens, we articulate the following central \underline{research questions}:
\begin{enumerate}[label=(\textit{\roman*})]
    \item How can we develop a task representation in a principled manner that generalizes across modalities such as language and mathematical functions?
    \item \revision{How can this representation be leveraged to preserve task fidelity on downstream tasks?}
\end{enumerate}

\subsection{Learnable Task Vector}
Our approach addresses key limitations of previous methods, such as Function Vectors, by automatically computing task formulations without the need for prior model analysis or manual adjustments.

\paragraph{Variability in Contributions of Attention Heads}
\revision{The FV encoding is based on strong causal evidence that attention heads are the main carriers of in-context task information \citep{fv}. This formulation assumes equal contributions from all attention heads to the task representation by summing their activations. However, we argue that the influence of each head varies—some contribute significantly to task encoding, while others have a lesser impact.} Constraining weights to unity is therefore impractical, and focusing exclusively on a subset of attention heads risks overlooking subtle contributions from others. To address this, attention heads should be weighted in the summation to accurately capture their varying contributions.

\paragraph{Layer-Specific Task Vectors}
Previous work typically integrates task encodings into models by adding vectors to the outputs of specific hidden layers through simple vector summation \citep{fv, icl_creates_task_vectors, dongfang}. However, this approach assumes uniform applicability across layers, ignoring that transformers' hidden states represent progressively refined representations of the input. To overcome this limitation, we compute task vectors uniquely for each hidden layer, aligning them with the evolving representations. While initializing a unique set of attention head weights for each of the \(L\) layers is an option, it would be computationally expensive. Instead, we compute layer-specific task vectors as a weighted sum of the attention heads within each layer:
\begin{equation}
\boxed{
    v^t_\ell = \sum_{j = 0}^{J} \omega_{\ell, j} \cdot \Bar{a}_{\ell, j},
}
\end{equation}
where \(\omega_{\ell, j} \in \mathbb{R}\) represents the weight parameters assigned to attention heads, organized in the parameter vector \(\Phi \in \mathbb{R}^{LJ}\), and \(\bar{a}\) is the attention activations averaged on a separate sample set of prompts demonstrating task \(t\). This method ensures that each layer-wise FV is composed solely of the \(J\) heads within that specific layer, avoiding the aggregation of attention across all \(L \times J\) heads.

Although excluding attention heads from layers \(\ell^\prime \neq \ell\) in the latter modification might seem contradictory, this is counterbalanced by the transformer's feedforward design. Including attention from earlier layers could introduce redundancy and complicate learning, as the hidden state at layer \(\ell\) already encapsulates all transformed information in layers \(\ell^\prime < \ell\). Furthermore, integrating attention heads from future layers would conflict with the transformer’s sequential processing that avoids forward-looking capabilities. 

We refer to the resulting formulation as \emph{Learnable Task Vector} (LTV) and illustrate its operation during inference in Figure \ref{fig:ltv_inference}. Next, we describe the methodology for learning the weights \(\Phi\) of LTV.

\begin{algorithm*}[!htbp]
\caption{Optimizing Learnable Task Vector (LTV)}
    \label{alg:ltv_training}
    \begin{algorithmic}[1]
    \Statex \textbf{Input:} Model $M_\theta$, task $t$, and number of samples $N$
    \Statex \textbf{Initialize and freeze:} Model parameters $\theta$ (no updates)
    \State Collect $N$ prompts $\{p^t_i\}_{i=1}^N$ demonstrating task $t$
    \Repeat
        \For{each $p^t_i$ in $\{p^t_i\}_{i=1}^N$}
            \State Compute LTV: $\mathbf{v}_\Phi^t = \{v^t_\ell\}_{\ell=1}^L$, where $v^t_\ell = \sum_{j=0}^{J} \omega_{\ell, j} \cdot \Bar{a}_{\ell, j}$
            \State Obtain the LTV-integrated model output: $\hat{y}_{\text{query}, i} \gets M_\theta(p^t_i \mid \mathbf{v}_\Phi^t)$
            \State Compute the loss: $\mathcal{L}(y_{\text{query}, i}, \hat{y}_{\text{query}, i})$ based on \eqref{eq:ltv_train_synthetic} or \eqref{eq:ltv_train_language}
            \State Compute gradients and update $\omega_{\ell, j}$ (model parameters $\theta$ remain frozen)
        \EndFor
    \Until{\textbf{convergence}}
\end{algorithmic}
\end{algorithm*}

\subsection{How to Learn Task Representations}
We optimize LTV weights through gradient steps to approximate the ``true'' weights the model uses to represent ICL tasks, aiming to validate this formulation as a \textit{proof-of-concept}. Consequently, this approach is \textit{not} intended to directly address issues like length generalization shown in Figure \ref{fig:transformers_failure}. \revision{Instead, LTV serves as a mechanistic probe into the model's internal structure, demonstrating that transformer architectures admit $d$-dimensional vectorization of task information that can be manipulated to steer behavior.}

An effective strategy for learning the parameters $\Phi$ involves optimizing the LTV in scenarios where the model (with frozen parameters) fails to perform ICL, isolating the causal contribution of $\Phi$ to the task performance. This approach ensures our method is both \emph{data-driven} and \emph{causal}, as it directly attributes improvements to \(|\Phi| = L \times J\) parameters, facilitating application to downstream tasks. The overall pipeline for training an LTV is summarized in Algorithm \ref{alg:ltv_training}.

\paragraph{Regression Tasks} The LTV is computed over prompts longer than the maximum length encountered during pre-training (\(T_\mathbf{v} > T_\text{train}\)). Its integration into the model modifies the transformer's output. Subsequently, backpropagation is performed over \(\Phi\) through the transformer to minimize the loss on the query input \(x_\text{query}\):
\begin{equation}
\label{eq:ltv_train_synthetic}
    \underset{\Phi}{\text{min}} \text{ } \mathbb{E}_{f \sim \mathcal{D}_{\mathcal{F}}, x \sim \mathcal{D}_\mathcal{X}} \bigg[\Big(\Tilde{M}_\theta \big(p^f \big .\mid \mathbf{v}_\Phi^f \big) - f(x_\text{query})\Big)^2\bigg],
\end{equation}
where \(\mathbf{v}_\Phi^f = \{v_{\ell}^f\}_{\ell=1}^L\) represents the set of layer-wise LTVs computed for function \(f\), \(\Tilde{M}_\theta(\cdot \mid \mathbf{v}_\Phi^f)\) denotes the transformer output modified by integrating \(\mathbf{v}_\Phi^f\) into the corresponding hidden layers, and the regression prompt $p^f$ is in the structure:
\begin{equation*}
    p^f = \{x_1, f(x_1), \ldots, x_{T_\mathbf{v}}, f(x_{T_\mathbf{v}}), x_\text{query}\}.
\end{equation*}

\paragraph{Language Tasks} Given the true query output \(y_\text{query}\), the LTV is trained in a supervised manner to minimize the cross-entropy loss on shuffled prompts:
\begin{equation}
\label{eq:ltv_train_language}
    \underset{\Phi}{\text{min}} \text{ } \mathbb{E}_{\Tilde{p}^t \sim \Tilde{P}^t} \Big[- \log \Big( \Tilde{M}_\theta\big(y_\text{query} \mid \Tilde{p}^t; \mathbf{v}_\Phi^t\big) \Big)\Big],
\end{equation}
where \( \Tilde{M}_\theta(y_\text{query} \mid \Tilde{p}^t; \mathbf{v}_\Phi^t)\) denotes the probability predicted by the model for the true class \(y_\text{query}\) given the shuffled prompt \(\Tilde{p}^t\).

\section{Experiments}
\label{sec:experiments}

\paragraph{Models} 
We employ decoder-only autoregressive transformers: GPT-2 \citep{gpt2} for regression tasks and GPT-J \citep{gptj} for tasks in the language domain, using the default configurations described by \citet{garg_icl} and \citet{fv}, respectively. GPT-2 is configured with 9.5M parameters (256 hidden dimension) across 12 layers and 8 attention heads per layer, while GPT-J features 6B parameters, 28 layers, and 16 attention heads per layer. The GPT-2 models pre-trained on functional regression data follow the training procedure outlined by \citet{garg_icl}.

\paragraph{Benchmarking}
For the regression data, we evaluate performance on sequences longer than the maximum length seen during pre-training but within the model's capacity: \(T_\text{train} < T_\textbf{v} \leq T_\text{max}\), as described in Section \ref{sec:motiv_obs}. \revision{This setup provides a controlled environment to assess the functional restoration of task behavior through causal interventions.} In the language domain, we adopt the approach of shuffling labels of in-context examples to diminish ICL \citep{fv}, ensuring fair comparisons with \citet{fv, icl_creates_task_vectors, dongfang}.

\renewcommand{\thesubfigure}{}

\begin{figure*}[!tbh]
    \centering
    % \vspace{-27.5pt}
    \begin{equation*}
        \text{\footnotesize {\blue} Transformer} \quad \text{\footnotesize {\red} Trans. +LoRA} \quad \text{\footnotesize {\green} Trans. +FV (optimized)} \quad \text{\footnotesize {\purple} Trans. +ICV (tuned)} \quad \text{\footnotesize {\orange} Trans. +LTV}
        \vspace{-14bp}
    \end{equation*}
    \subfigure[\small (a) Linear functions (\(T_\text{train} = 41\))]{
            \subfigure[\phantom{(a)} \(T_\mathbf{v} = 56\)]{
                \includegraphics[width=0.14\linewidth]{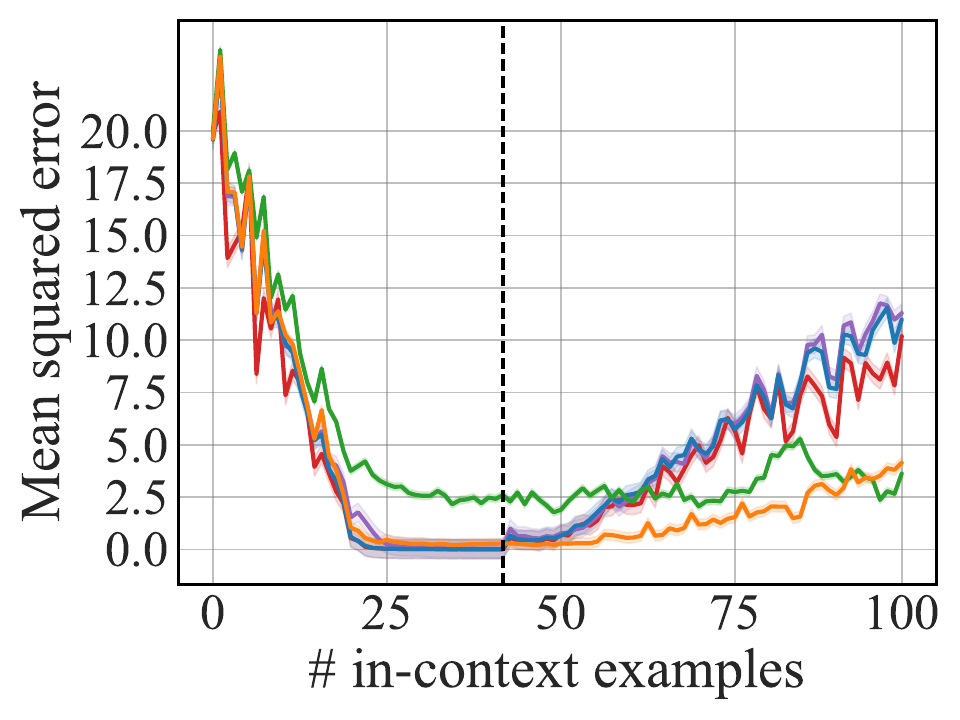}%
            }
            \subfigure[\phantom{(a)} \(T_\mathbf{v} = 71\)]{
                \includegraphics[width=0.14\linewidth]{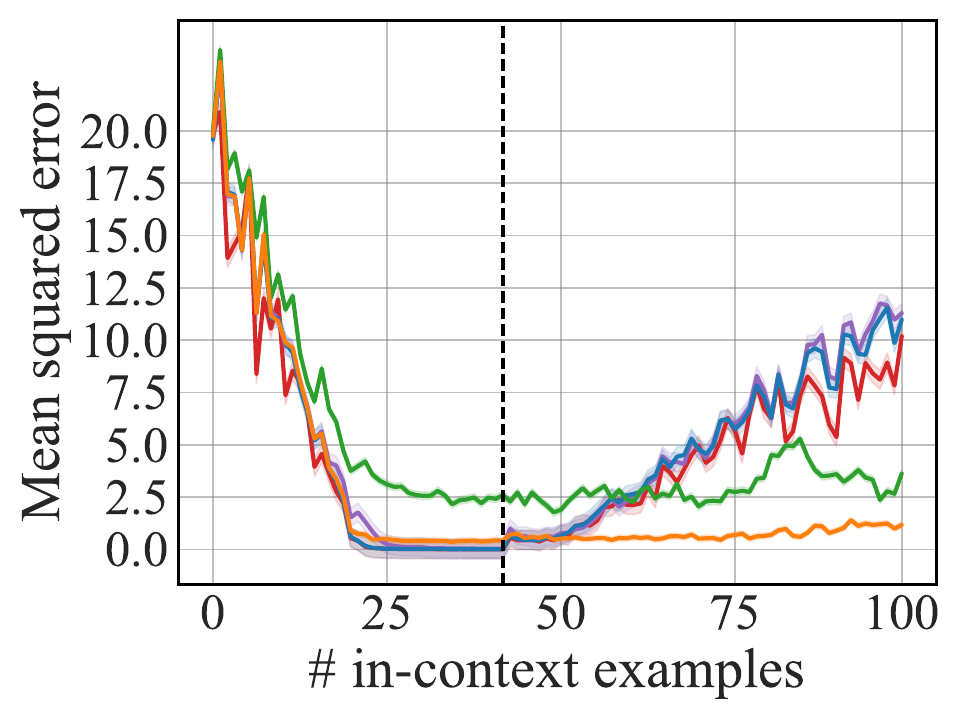}%
            }
        }
	\subfigure[\small (b) Sparse linear functions (\(T_\text{train} \hspace{-2bp}=\hspace{-1bp} 41\))]{
            \subfigure[\phantom{(a)} \(T_\mathbf{v} = 56\)]{
                \includegraphics[width=0.14\linewidth]{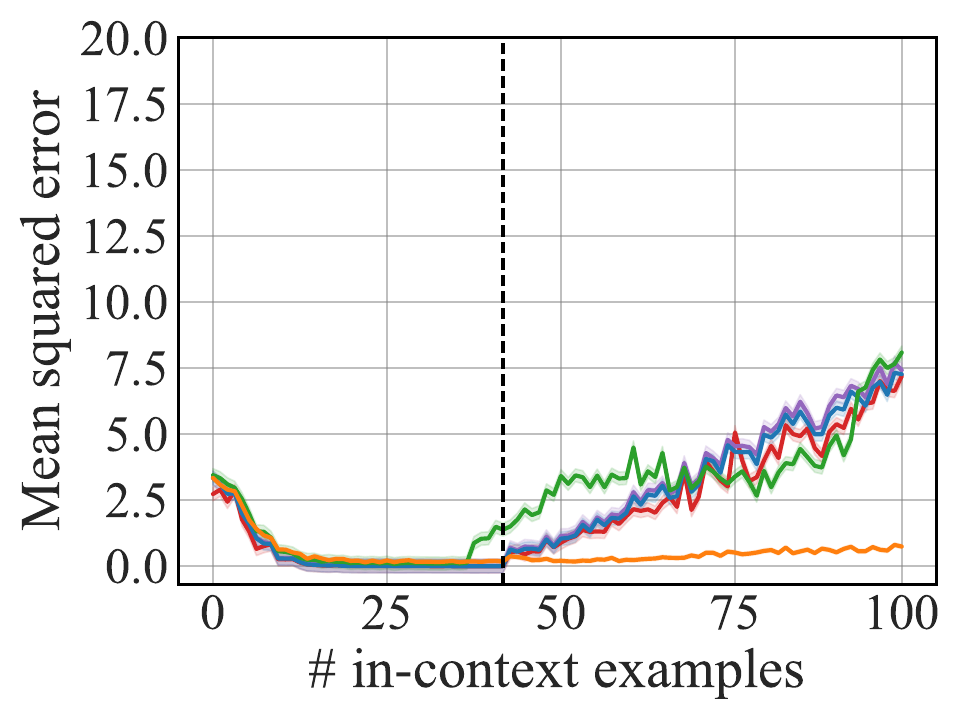}%
            }
            \subfigure[\phantom{(a)} \(T_\mathbf{v} = 71\)]{
                \includegraphics[width=0.14\linewidth]{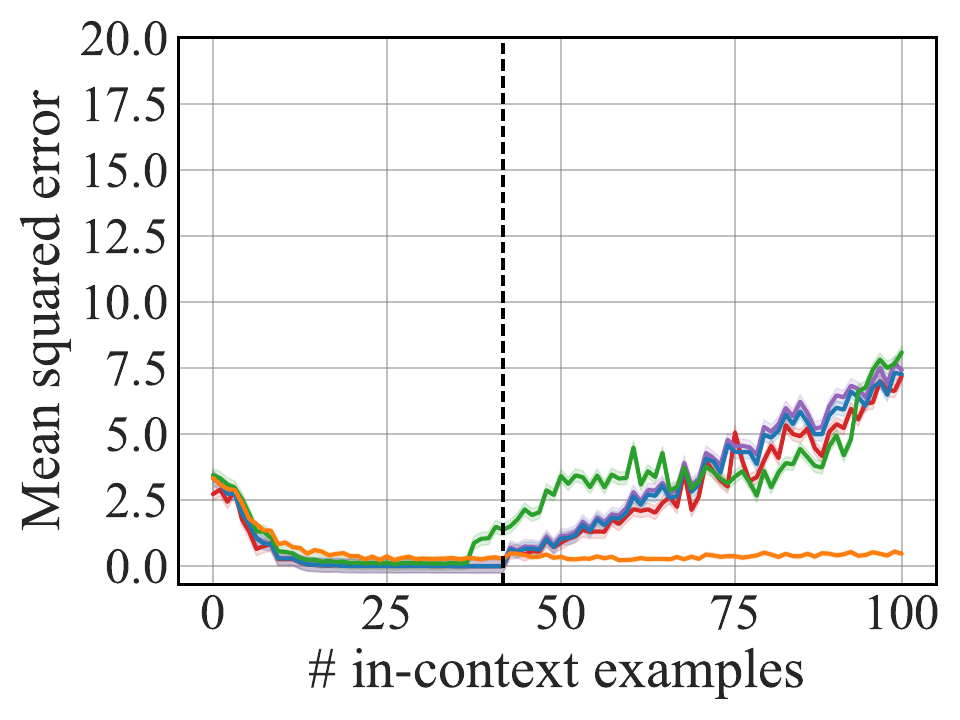}%
            }
        }
        \subfigure[\small (c) 2-layer ReLU NN (\(T_\text{train} = 101\))]{
            \subfigure[\phantom{(a)} \(T_\mathbf{v} = 126\)]{
                \includegraphics[width=0.14\linewidth]{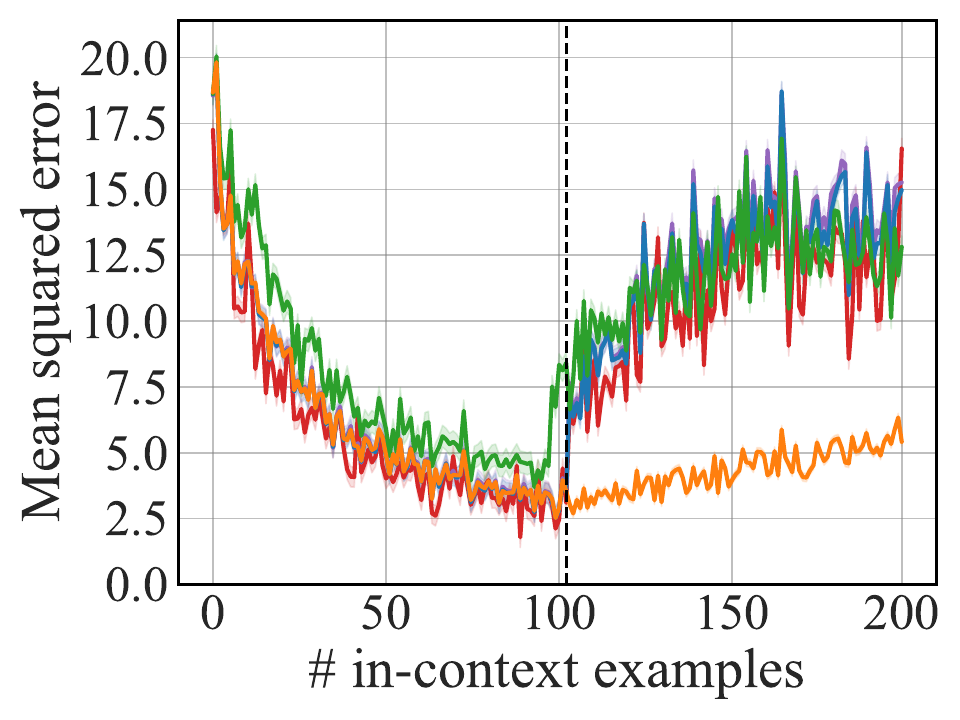}%
            }
            \subfigure[\phantom{(a)} \(T_\mathbf{v} = 151\)]{
                \includegraphics[width=0.14\linewidth]{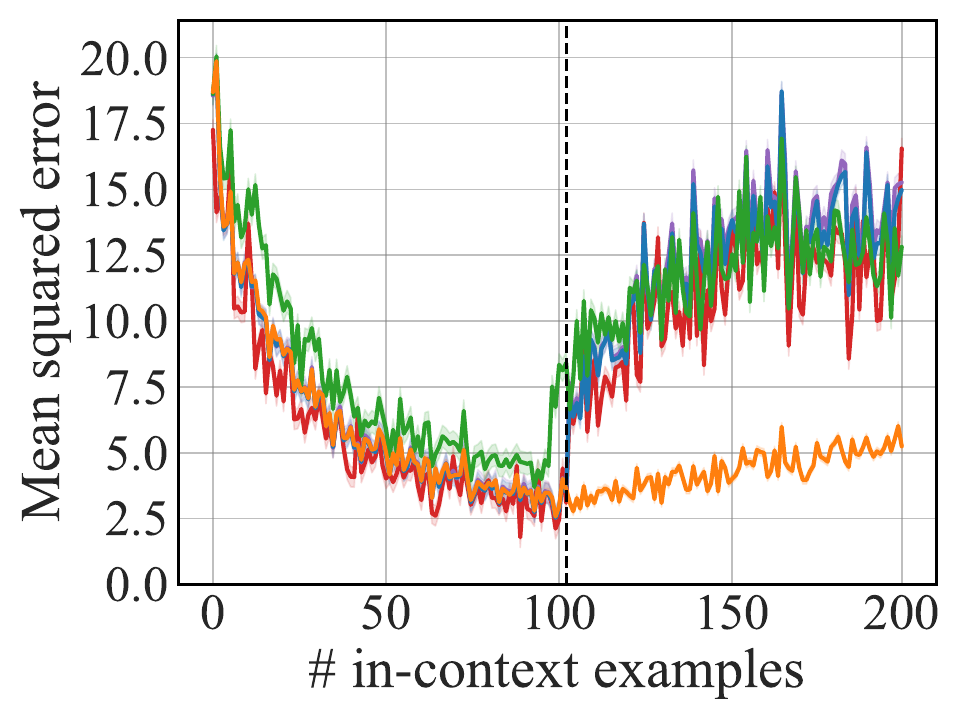}%
            }
        }
\vspace{-5.5bp}
\caption{Squared error on the query input, averaged over a batch of 256 tasks randomly selected from the same function class. The shaded area represents the 95\% confidence interval over the sampled prompts. The dashed line indicates the number of examples the transformer was trained with, and \(T_\mathbf{v}\) denotes the prompt length used in LTV training. Complete results for different \(T_\mathbf{v}\) values are provided in Figures \ref{fig:complete_results_lin}, \ref{fig:complete_results_sparse_lin}, and \ref{fig:complete_results_relu} in Appendix \ref{app:complete_regression_evals}.}
\label{fig:synthetic_results}
\end{figure*}

\paragraph{Tasks}
{For regression data, as illustrated in Figure \ref{fig:transformers_failure}, we evaluate the models on classes of linear functions, sparse linear functions, and 2-layer ReLU networks \citep{garg_icl}. While decision trees have also been explored as a function class, they exhibit less sensitivity to positional encodings, which we discuss in Appendix \ref{app:sampling_functions}.

For NLP tasks, we focus on single-token generation, as considered in previous work \citep{icl_creates_task_vectors, fv}. The tasks are derived from widely studied natural language corpora, including SST-2 \citep{sst_2} and CoNLL-2003 \citep{ner}. Further details are provided in Appendix \ref{app:language_tasks}.

\paragraph{Baseline Methods} 
For functional regression, we include FV \citep{fv} and ICV \citep{icv} as baselines for comparison. The other methods discussed in Appendix \ref{app:related_work} were primarily designed for language tasks and are not applicable to regression tasks (see Appendix \ref{app:other_baselines} for details). Although FV and ICV were originally proposed for linguistic applications, we adapted them for regression tasks with several optimizations. Additionally, we consider Low-Rank Optimization (LoRA) \citep{lora} for further discussion. For language tasks, we also include Task Vector (TV) \citep{icl_creates_task_vectors} and State Vector (SV) with inner optimization \citep{dongfang} alongside FV and ICV.

\paragraph{Learnable Task Vector}
For each task, we curated a dataset of 10,000 samples to demonstrate the underlying task. LTVs were trained for approximately 2,000 iterations on regression tasks and 120,000 iterations on language tasks, using a lower learning rate for the latter. Each iteration involved sampling a batch of 256 input prompts and performing gradient descent based on the objectives in \eqref{eq:ltv_train_synthetic} and \eqref{eq:ltv_train_language} for regression and language tasks, respectively.

Details of our experimental setup are provided in Appendix \ref{app:exp_details}. The full set of results is available in Appendix \ref{app:complete_results}, including: \textit{(i)} LTV trained with various $T_\mathbf{v}$, optimized and tested on chain-of-thought (CoT) prompts, and evaluations under distributional shift; \textit{(ii)} accuracy scores on 15 NLP benchmarks; and \textit{(iii)} complete ablation results.

\subsection{Evaluation on Regression Tasks}
The loss curves for ICL inference are shown in Figure \ref{fig:synthetic_results}. While FV offers some benefits in linear regression, ICV negatively impacts the vanilla model's performance. Despite extensive optimizations, these results indicate that task encoding formulations, originally crafted for language, do not generalize well to other modalities.

\begin{table*}[!t]
\centering
\resizebox{\textwidth}{!}{%
\begin{tabular}{@{}c lcccccccc@{}}
\toprule
\multicolumn{1}{l}{} & Dataset       & Prompt        & Transformer & +LoRA & +FV   & +ICV  & +SV   & +TV   & +LTV (ours) \\ 
\midrule
\multirow{6}{*}{\begin{sideways}Abstractive\end{sideways}} 

    & \multirow{2}{*}{AG News} & Zero-shot & 0.0 $\pm$ 0.0 & 9.1 $\pm$ 0.2 & 30.1 $\pm$ 5.7 & 33.6 $\pm$ 5.6 & 31.7 $\pm$ 5.7 & 34.5 $\pm$ 5.0 & \highlight[customorange]{\textbf{46.9 $\pm$ 6.2}} \\
     & & Few-shot & \highlight[customorange]{67.2 $\pm$ 5.8} & \highlight[customorange]{69.3 $\pm$ 3.8} & \highlight[customorange]{66.0 $\pm$ 5.8} & 64.7 $\pm$ 5.4 & 66.1 $\pm$ 4.5 & 57.9 $\pm$ 5.8 & \highlight[customorange]{\textbf{71.1 $\pm$ 5.6}} \\
     & \multirow{2}{*}{CommonsenseQA} & Zero-shot & 10.2 $\pm$ 3.7 & 24.8 $\pm$ 4.1 & 23.0 $\pm$ 5.2 & 28.9 $\pm$ 5.2 & 21.7 $\pm$ 1.1 & 22.0 $\pm$ 4.0 & \highlight[customorange]{\textbf{41.0 $\pm$ 6.1}} \\
     & & Few-shot & \highlight[customorange]{18.0 $\pm$ 4.7} & \highlight[customorange]{18.6 $\pm$ 5.5} & \highlight[customorange]{20.3 $\pm$ 5.0} & \highlight[customorange]{20.0 $\pm$ 4.6} & \highlight[customorange]{20.3 $\pm$ 2.1} & \highlight[customorange]{19.5 $\pm$ 3.4} & \highlight[customorange]{\textbf{21.1 $\pm$ 5.0}} \\
     & \multirow{2}{*}{Sentiment Analysis} & Zero-shot & 0.0 $\pm$ 0.0 & 0.7 $\pm$ 0.6 & 0.0 $\pm$ 0.0 & 4.8 $\pm$ 1.9 & 0.0 $\pm$ 0.0 & 0.0 $\pm$ 0.0 & \highlight[customorange]{\textbf{10.9 $\pm$ 3.8}} \\
     & & Few-shot & 74.6 $\pm$ 5.4 & 75.9 $\pm$ 4.1 & 69.5 $\pm$ 5.7 & 72.5 $\pm$ 3.2 & 74.2 $\pm$ 2.5 & 77.1 $\pm$ 3.4 & \highlight[customorange]{\textbf{94.5 $\pm$ 2.8}} \\
     \midrule
    
\multirow{6}{*}{\begin{sideways}Extractive\end{sideways}} 
    & \multirow{2}{*}{NER-person} & Zero-shot & 5.5 $\pm$ 2.8 & 15.7 $\pm$ 1.1 & 48.8 $\pm$ 6.2 & 49.9 $\pm$ 5.8 & 46.3 $\pm$ 1.3 & 47.8 $\pm$ 5.3 & \highlight[customorange]{\textbf{58.2 $\pm$ 6.1}} \\
     & & Few-shot & 11.7 $\pm$ 4.0 & 12.1 $\pm$ 1.6 & 56.2 $\pm$ 6.1 & 65.9 $\pm$ 5.2 & 53.7 $\pm$ 3.3 & 62.6 $\pm$ 3.9 & \highlight[customorange]{\textbf{79.3 $\pm$ 5.0}} \\
     & \multirow{2}{*}{NER-location} & Zero-shot & 6.6 $\pm$ 3.1 & 6.5 $\pm$ 2.4 & \highlight[customorange]{\textbf{37.9 $\pm$ 6.0}} & 26.4 $\pm$ 5.1 & \highlight[customorange]{34.6 $\pm$ 3.9} & 27.2 $\pm$ 4.7 & 27.3 $\pm$ 5.5 \\
     & & Few-shot & 23.4 $\pm$ 5.2 & 29.5 $\pm$ 5.5 & 43.4 $\pm$ 6.1 & 47.4 $\pm$ 5.9 & 43.5 $\pm$ 4.4 & 41.9 $\pm$ 5.6 & \highlight[customorange]{\textbf{57.4 $\pm$ 6.1}} \\
     & \multirow{2}{*}{NER-organization} & Zero-shot & 21.9 $\pm$ 5.1 & 21.9 $\pm$ 6.6 & 48.8 $\pm$ 6.2 & 52.5 $\pm$ 6.0 & 46.1 $\pm$ 1.2 & 46.6 $\pm$ 4.3 & \highlight[customorange]{\textbf{64.1 $\pm$ 5.9}} \\
     & & Few-shot & 16.8 $\pm$ 4.6 & 19.2 $\pm$ 3.8 & 50.8 $\pm$ 6.2 & 54.1 $\pm$ 5.0 & 53.5 $\pm$ 2.6 & 51.1 $\pm$ 4.1 & \highlight[customorange]{\textbf{75.0 $\pm$ 5.3}} \\
\bottomrule
\end{tabular}}
\caption{Accuracy scores (\%) for zero-shot and few-shot (5-shot) predictions, averaged across 256 random seeds. \(\pm\) represents the margin of error at a 95\% confidence level. The highest accuracy is marked in \textbf{boldface}, and the statistically best-performing method is \highlight[customorange]{highlighted}. The complete results are provided in Table \ref{tab:language_results_full} in Appendix \ref{app:nlp_results}, along with references to the benchmark datasets.}
\label{tab:language_results}
\end{table*}

\paragraph{\revision{LTV preserves task information in the OOD case.}}  
LTV shows minimal performance difference when trained with prompt lengths close to \(T_{\text{train}}\). However, as \(T_{\mathbf{v}}\) increases, LTV's impact becomes more prominent, achieving optimal performance comparable to the vanilla model at \(T = T_{\text{train}}\) (see Appendix \ref{app:complete_regression_evals} for detailed plots). While this might raise concerns about the validity of the results, as LTV is causally optimized in the OOD cases (\(T > T_{\text{train}}\)), training LTV with prompt lengths slightly exceeding \(T_{\text{train}}\)—specifically, \(T_{\mathbf{v}} = 1.37 \times T_{\text{train}}\) for linear functions and \(T_{\mathbf{v}} = 1.25 \times T_{\text{train}}\) for 2-layer ReLU networks—proves sufficient to maintain performance. Extending \(T_{\mathbf{v}}\) further offers no significant improvements, indicating that only a small portion of the OOD case is needed for LTV to generalize. This suggests that the attention head weights naturally converge to their optimal values with sufficient training data.

\paragraph{Benefits of LTV come from effective task encoding, not fine-tuning.} {The effects of LTV in regression tasks may resemble those of parameter-efficient fine-tuning (PEFT) methods, as only a small number of parameters (\(|\Phi| = L \times J\)) are trained to adapt the model to previously unlearned scenarios. To evaluate this, we compare LTV with LoRA (rank 8). While LoRA fine-tunes approximately 196K of the 9.5M GPT-2 parameters, LTV optimizes only 96. Despite LoRA fine-tuning a significantly larger number of parameters on sequences with a maximum length \(T_\text{max}\), no notable performance improvement is observed. Since LoRA depends on the pre-trained model's existing positional representations, it cannot address untrained positional encodings. This observation indicates that the task fidelity achieved by LTV is not a result of sole fine-tuning but rather of effective task encoding that can steer the behavior of the model.

\paragraph{Interpretation of the Learned LTV Weights} No regularization, optimization techniques, dropout, or specific weight initialization were applied to learn the LTV weights. The weights are unbounded and sampled from the standard Gaussian distribution. Visualizing the learned weights as heatmaps revealed they consistently lie within the interval [-3, 3] across tasks. However, no meaningful patterns were observed though, suggesting that any underlying structure may be subtle and challenging to detect visually. \revision{Notably, some weights converge to near-zero, supporting the idea that only a subset of attention heads encodes task information. Our method thus provides a principled way to identify and weigh this subset, rather than assuming uniform importance or selecting it manually.}

\subsection{Evaluation on Language Tasks}
The accuracy scores are reported in Tables \ref{tab:language_results} and \ref{tab:language_results_full}. While \citet{fv} primarily considered ``filtered'' accuracies, which take into account only the test queries where at least one model responds correctly, we present unfiltered accuracies as a fairer metric, counting all samples regardless of model performance.

\paragraph{LTV is also superior in the language domain.} 
In 29 out of 30 evaluations (15 tasks with zero- and few-shot prompts), LTV achieves the highest accuracy and ranks among the statistically best-performing methods (see Table \ref{tab:language_results_full} in Appendix \ref{app:nlp_results} for complete results). Furthermore, in 19 out of 30 evaluations, LTV's performance is statistically superior to all other techniques. We also observe a notable performance by ICV, particularly when the vanilla transformer performs poorly in the zero-shot setting. This effectiveness is attributed to ICV's approach of crafting unique task vectors for each layer, similar to our method. \revision{Lastly, while evaluating LTV with higher shot counts may reveal diminishing gains, we expect the gains to remain nonzero and notable, as the vanilla model may not achieve further contextualization beyond a saturation point.} Overall, these findings confirm that LTV excels in both regression and challenging linguistic tasks, supporting our hypothesis about how ICL tasks are formulated within the transformer architecture.

\subsection{Ablation Studies}
We aim to understand how LTVs sustain task behavior when ICL is not performed. To investigate this, we use the regression data benchmark and analyze the last hidden states of an LTV-integrated model and a vanilla model at \(T = T_\text{train}\) (the optimal-performing model). We measure the distributional similarities of the last hidden states using KL divergence to assess how closely the LTV-integrated model aligns with the vanilla model.

The KL divergence scores are reported in Table \ref{tab:ablations} in Appendix \ref{app:ablation_results}. From the baselines, we only include FV, as it is the only method that does not degrade the vanilla transformer. The corresponding probability densities of the last hidden state distributions, visualized through histograms, are also included in Appendix \ref{app:ablation_results}. Additionally, a detailed description of our experimental methods, along with a figure illustrating the pipeline, is provided in Appendix \ref{app:ablation_details}.

\paragraph{LTV maintains the last hidden state distribution with that of the optimal-performing model.} 
We observe that the divergence between the last hidden state distributions of the LTV-integrated model and the optimal-performing model is generally low and significantly decreases when the LTV training length matches or exceeds the middle length. This suggests that LTV's effectiveness primarily stems from its ability to dominate the last hidden state and align it with that of the optimal-performing model, thereby facilitating task steering. This observation further implies that task steering may \textit{generally} be achieved through this mechanism, not exclusively by our method. However, a deeper analysis is required to fully understand this process, which is beyond the scope of this study.

\section{Conclusion}
We investigated the internal representation of in-context learning (ICL) tasks in large language models (LLMs). We propose the \textit{Learnable Task Vector} (LTV) framework, which encodes ICL tasks as a weighted sum of attention head activations optimized through causal gradient descent. LTV generalizes across both text and regression tasks, addressing the limitations of existing representations. Through empirical validation using a novel benchmark designed to analyze task encodings, we demonstrate that LTV preserves task fidelity when models fail to extract task information from demonstrations. Our findings further suggest that this preservation is achieved by maintaining the last hidden state distribution similar to that of an optimally in-context-learned model, bringing us closer to understanding how task steering works in general.

\section*{Limitations}
Although not explicitly stated in the main body, our study assumes access to the models' hidden states, limiting the scope to open- or white-box models. Additionally, we acknowledge that optimizing LTV can be time- and resource-intensive, \revision{e.g., requiring 120K iterations over 10K samples for regression. LTV training serves only to learn a fixed task representation vector while keeping model weights frozen, and it is not intended to replace ICL or few-shot prompting.} A dataset for LTV training must first be generated or collected. For functional regression, this is relatively straightforward, as data and regression coefficients can be sampled from distributions (e.g., zero-mean Gaussian). However, for language data, it must be collected or generated using a generative AI tool. As a reminder, we utilized data collected by \citet{fv} based on widely studied datasets such as SST-2 \citep{sst_2} and CoNLL-2003 \citep{ner}. Finally, because hidden layer access is required, deploying an open-source model on GPUs may demand significant computational resources.

\section*{Ethical Considerations}  
We do not identify any direct application of our study for unethical purposes.

\bibliography{custom}

\appendix

\section{Extended Related Work}
\label{app:related_work}

A substantial body of research on ICL has been ongoing since its discovery. We review previous studies through different facets of ICL. While our study shares similarities with others, it most closely aligns with the studies described in Section \ref{sec:related_work_task_repr}, where complementary details are provided in the final paragraph of this section.

\paragraph{In-Context Learning}
The ICL capabilities of LLMs were first identified by \citet{gpt3}. Since then, ICL has been extensively studied from various angles. \citet{min_2022, yoo_2022} have examined the effects of different ICL prompts styles. ICL during inference time has also been explored through meta-learning analyses \citep{akyurek, dai_2023, transformers_learn_icl_by_grad_des, transformers_as_algos, garg_icl}. In addition, investigations into ICL task inference from a Bayesian perspective have been conducted \citep{xie_2022, wang_2023c, wies_2023, zhuoran_icl}. Lastly, the scalability of ICL across different model sizes has been examined by \citet{wei_2023, wang_2023c, pan_2023}. While these studies primarily focus on the externally observable behaviors of models during inference and ICL, our study delves into the internal mechanisms of transformers to encode ICL tasks.

\paragraph{The Role of Attention Mechanism in Explaining Model Behavior}
Past analyses of the attention mechanism \citep{voita_2018, bert_attention, voita_2019, bert_dark_secrets, reif_2019, lin_2019, htut_2019, kobayash_2020} have revealed that attention weights often align with linguistic structures. However, these studies primarily focused on explaining the behavior of bidirectional architectures. Moreover, attention scores alone have not been found to fully explain the model's outputs \citep{attn_not_explanation, attn_not_not_explanation, bibal_2022}. In our work, we aim to deepen the understanding of the role of multi-head self-attention in ICL. Specifically, we investigate the contribution of each attention head to the model's internal representation of the ICL task.

\paragraph{Mechanisms to Explain Task Performance in In-Context Learning}
The components of transformers during ICL inference have been investigated to identify the origins of incorrect predictions and false statements \citep{merullo_2024, halawi_2024}. Similarly, numerous studies have adjusted attention mechanisms or activations of hidden layers during inference to steer model behavior \citep{li_2023a, subramani_2022, turner_2023, rimsky_2024, icv}. It was observed that tokens representing labels in an ICL prompt might hold the semantic information crucial for the language model’s final prediction \citep{wang_2023a}. Moreover, it was suggested that certain neurons within pre-trained transformers are highly predictive of task labels and empirically demonstrate that these neurons encode task-specific skills \citep{wang_2023b}. {A similar effort has been made to identify the critical layers where task information is stored \citep{where_icl_happens}. It has been shown that critical layers vary across tasks and can even differ within the same task, such as between subtasks like English-to-French and French-to-English translation. In contrast, our work aims to develop a principled conceptualization—such as a function of the transformer's components—that can effectively \emph{represent} and \emph{automatically differentiate} across a wide range of tasks, regardless of task modality. We further leverage this conceptualization to guide the language model's behavior across diverse tasks, aligning with prior works discussed next.

\paragraph{Tasks Representations in In-Context Learning}
Efforts to represent ICL tasks typically start by deriving a task vector from the layer activations \citep{icl_creates_task_vectors}, referred to as the \textit{state}, at the position of a dummy query's separator token. This token separates inputs and outputs in the in-context examples using specific symbols such as ``$\rightarrow$''. The dummy query is strategically placed immediately before the actual query to ensure that the task vector remains independent of the query itself. This method was further refined by averaging the states at each of the $N$ separator tokens in an $N$-shot prompt, followed by optimization using inner and momentum techniques \citep{dongfang}. However, these approaches are limited to ICL prompts that incorporate a separator token (\emph{e.g.}, ``$\rightarrow$''), restricting their broader applicability. Moreover, it has been shown that the principal direction of layer activation differences can effectively guide the ICL task \citep{icv}, leading to the task representation known as \textit{In-Context Vectors} (ICVs). Nonetheless, it has been recently argued that the focus should be on attention heads \citep{fv}, as these are crucial for transferring information between token positions \citep{transformer, math_of_transformers}.

Deriving from this finding, the \textit{Function Vector} (FV) is computed as the sum of the outputs from a selectively chosen subset of attention heads based on an \emph{indirect metric} \citep{fv}, derived from causal inference literature \citep{direct_and_indirect_effects}. To the best of our knowledge and based on our preliminary analyses, the most effective empirical representation of tasks in ICL are FVs. Therefore, our approach starts by deriving from \citet{fv}, in contrast to methods the outlined by \citet{icl_creates_task_vectors, icv}. Instead of merely using raw activations, we optimize weights assigned to these heads to enhance transformer performance in scenarios where ICL cannot be performed. Ultimately, this leads to a more formalized conceptualization that can be adapted to various models and tasks, whether functional regression or linguistic.

\section{Discussions of Alternative Approaches to Assess Task Representations}
Section \ref{app:stop_icl_shuffling} explains why shuffling labels in regression tasks does not prevent the model from extracting task information. Section \ref{app:probing_ineffective} discusses why probing is ineffective for assessing the precision of task representations.

\subsection{Why Shuffling Labels is not a Valid Approach for Functional Regression}
\label{app:stop_icl_shuffling}  
We analyze the effect of label shuffling on in-context learning performance for two example tasks: linguistic antonym generation and linear regression. While label shuffling disrupts the model’s ability to infer relationships in the linguistic task, its impact on the numeric task differs fundamentally due to the task's nature and the model's pre-training, particularly if the model was pre-trained on similar functional regression tasks \citep{garg_icl}.

\subsubsection{Linguistic Antonym Generation}  
In the linguistic task, the model is provided with prompts such as:  
\begin{equation*}  
\text{cold} \rightarrow \text{hot}, \text{ happy} \rightarrow \text{sad}, \text{ fast} \rightarrow \text{ ?}  
\end{equation*}  
The model’s goal is to generate the antonym of ``slow'' based on in-context examples. Since LMs are pre-trained on vast text corpora without specific emphasis on antonym pairs, they rely heavily on these examples to infer the relationship. Shuffling the labels in the examples disrupts the pattern, preventing the model from identifying the intended relationship (antonymy) and generating the correct output. Any potential performance degradation in this case is directly linked to the blockage of ICL caused by label shuffling.

\subsubsection{Numeric Linear Regression}  
In numeric tasks, the model processes sequences where each scalar output \(y_i\) is generated from a vector input \(x_i\) through a linear relationship \(y_i = w^\top x_i\):  
\begin{gather*}  
    [x_{1,1}, x_{1, 2}, \dots, x_{1, m}] \rightarrow y_1, \\ [x_{2, 1}, x_{2, 2},  \dots, x_{2, m}] \rightarrow y_2\text{ }, \\ \vdots \\ [x_{T, 1}, x_{T, 2}, \dots, x_{T, m}] \rightarrow \text{?},
\end{gather*}
where the \(m\)-dimensional input vector is flattened before being fed into the model, following our setup. The model's goal is to infer the coefficient vector \(w\) from the examples to predict the missing output. Pre-training embeds a strong bias toward linear mappings in the model's parameters, giving it a default assumption of linearity. In-context examples refine the \textit{exact coefficients} but are not the sole source of understanding this relationship. Consequently, shuffling the labels conflicts with the model's pre-trained understanding of input-output relationships (\emph{i.e.}, its inherent linear bias), leading to confusion rather than simply blocking ICL. In such cases, the shuffled prompt would still need to define a linear relationship, which is non-trivial. In contrast, for language tasks, the model retains an understanding of individual word semantics even when labels are shuffled, aligning with its pre-trained understanding of language.

\subsection{Why Probing Would be Ineffective to Assess Task Representations}
\label{app:probing_ineffective}
We outline a potential approach for probing experiments. One method is to assess whether the LTV encodes task information by using it to predict task-relevant quantities—in this case, quantities in regression tasks or correct labels in language tasks. This prediction is performed by training a linear predictor or a two-layer neural network.

However, we argue that even if the probe achieves high accuracy, this does not demonstrate that the LTV (or any input used for probing) truly encodes task information. Probing offers only indirect evidence; it shows that a mapping exists from the input to the target quantity, not that the input itself is a meaningful representation of the task. It simply indicates that the input contains enough information to predict the output.

In an extreme case, we could apply an arbitrary invertible transformation to the prompt and use that as input to the probe. The probe would likely achieve good performance because the prompt contains sufficient information to predict the task-relevant output. However, this input would clearly not be a meaningful representation, as it is too coarse.

In our case, the LTV is a representation specific to the LLM architecture. Moreover, we are not concerned with whether it can directly predict the task-relevant output—this is not how these vectors are intended to be used. Instead, they are applied to latent layers of the LLM to steer output generation, which is exactly how we evaluate their effectiveness.

\section{Experimental Details}
\label{app:exp_details}

\subsection{Implementation}
For all models, we use the \texttt{huggingface} implementations \citep{huggingface}. For regression tasks and GPT-2 pre-training, we rely on the code provided by the authors on GitHub\footnote{\url{https://github.com/dtsip/in-context-learning}}. Additionally, we use the authors' code\footnote{\url{https://github.com/ericwtodd/function_vectors}\label{fv_repo}} to include FV into our experiments. The remaining baseline methods—ICV, TV, and SV—were implemented by us.

\subsection{Experiments on Regression Tasks}
\label{app:synthetic_exp}
We closely follow the experimental setup established by \citet{garg_icl}. For completeness, we provide the relevant details here, with additional information available in the cited reference.

\subsubsection{Model}
The GPT-2 model processes sequences of vectors in the embedding space and outputs a sequence in the same space. However, the tasks we examine involve functions mapping a lower-dimensional vector space (\emph{e.g.}, 20 dimensions) to a scalar value. To construct a prompt such as \(p = \{x_1, f(x_1), x_2, f(x_2), \ldots, x_\text{query}\}\), we must map \(x_i\) and \(f(x_i)\) into the embedding space. This mapping involves first converting the scalar values \(f(x_i)\) into vectors of the same dimension as \(x_i\) by appending zeros, followed by applying a learnable linear transformation to all these vectors into the embedding space. The model's output vector is then transformed into a scalar value through a dot product with a learnable vector.

We consider the model’s prediction at the position corresponding to \(x_i\) (\emph{i.e.}, the absolute position \(2i - 1\)) as the prediction of \(f(x_i)\). Due to the model's structure, this prediction relies solely on the pairs \(\{x_j, f(x_j)\}\) for \(j < i\) and \(x_i\) itself. We disregard the model predictions at positions corresponding to \(f(x_i)\).

The GPT-2 models use absolute, learnable positional encodings and were trained to accommodate up to 101 examples for linear and sparse linear functions, and up to 201 examples for 2-layer ReLU neural networks in a prompt. While it is possible to feed the model prompts with more examples by adjusting the initialization, this would exceed our computational resources. We also did not want to alter the nature of their experimental process.

\subsubsection{Training}
\label{app:model_training_synth}
We train a model from scratch (\emph{i.e.}, no pre-trained weights are loaded) to predict \( f(x_i) \) for a given \( x_i \), using the set of examples as reference. Each training prompt is generated by randomly sampling a function \(f\) from the function class of interest, followed by sampling inputs \(x_i\) from an isotropic Gaussian distribution \(N(0, I_m)\). The prompt is constructed as \(p = \{x_1, f(x_1), \ldots, x_k, f(x_k)\}\). For each input \(i \leq k\) within a prompt, the model predictions \(\hat{y}_i = M_\theta(x_i \mid p^f = \{x_1, f(x_1), \ldots, x_{i - 1}, f(x_{i - 1})\})\) are obtained, and the loss is computed across all prompt prefixes:
\begin{equation*}
\text{\small$ 
    \underset{\theta}{\text{min}} \text{ } \mathbb{E}_{f \sim \mathcal{D}_{\mathcal{F}}, x \sim \mathcal{D}_\mathcal{X}} \bigg[\frac{1}{T+1} \sum_{i = 0}^{T} \Big(M_\theta \big(p^{f, i}\big) - f(x_{i+1})\Big)^2\bigg],
    $ \normalsize}
\end{equation*}
where \( \mathcal{L} \) is the loss function, typically chosen to be mean squared error, and we have \(x_{T+1} = x_\text{query}\).

During training, we average the loss across a batch of randomly generated prompts, each with different functions and inputs, and update the model parameters. The Adam optimizer \citep{adam} is employed and trained for a total of 500,000 steps with a batch size of 64, using a learning rate of \(10^{-4}\) for all function classes and models.

\paragraph{Curriculum Learning}
The training procedure is accelerated through curriculum learning. The model starts by observing prompt inputs \(x_i\) within a smaller dimensional subspace and with fewer inputs per prompt. Both the subspace dimension and the number of examples are increased gradually. Specifically, all of the coordinates except the first \(T_\text{max, cur}\) of \(x_i\) are zeroed out by sampling prompts of size \(T_\text{cur}\). For the function classes of linear and sparse linear functions, \(T_\text{max, cur} = 5\) and \(T_\text{cur} = 11\) are used initially, and \(T_\text{max, cur}\) and \(T_\text{cur}\) are increased by 1 and 2, respectively, every 2000 steps until reaching \(T_\text{max, cur} = T_\text{max}\) and \(T_\text{cur} = 2m + 1\). A different schedule is applied for 2-layer neural networks to accommodate the need for more inputs; starting from \(T_\text{max, cur} = 5\) and \(T_\text{cur} = 26\), \(T_\text{max, cur}\) and \(T_\text{cur}\) are incremented by 1 and 5 respectively, every 2000 steps until \(T_\text{max, cur} = T_\text{max}\) and \(T_\text{cur} = 5m + 1\).

Consequently, in the curriculum-based training approach, a training prompt 
\begin{equation*}
    p^f = \{x_1, f(x_1), \ldots, x_{T_\text{cur}}, f(x_{T_\text{cur}})\}
\end{equation*}
is generated by sampling a random function \(f\) from the function class and drawing inputs \(x_i\) by sampling i.i.d. from \(\mathcal{N}(0, I_m)\), with all but the first \(T_\text{max, cur}\) coordinates zeroed out. Given the model predictions \(\hat{y}_i\), the loss is computed as
\begin{equation*}
    \frac{1}{T_\text{cur}} \sum_{i = 1}^{T_\text{cur}} \big(\hat{y}_i - f(x_i)\big)^2.
\end{equation*}

\subsubsection{Sampling the Functions}
\label{app:sampling_functions}
For the class of linear functions, we sample the random function \(f(x) = w^\top x\) by drawing \(w \sim \mathcal{N}(0, I_m)\). In the case of sparse linear functions, \(w\) is also sampled from \(\mathcal{N}(0, I_m)\), but we then randomly zero out the first \(T_\text{cur}\) coordinates within the first \(T_\text{max, cur}\) coordinates. For these linear functions, we set \(m = 20\) for all experiments, with a sparsity level of 3. For 2-layer neural networks, we sample \(W_1\) from \(\mathcal{N}(0, I_m)\) and \(W_2\) from \(\mathcal{N}(0, 2 / r)\), where \(f(x) = W_2 \operatorname{ReLU}(W_1 x)\). Here, we set the dimensions \(m = 20\) and the ratio \(r = 100\).

We also experimented with decision trees as a function class, as done by \citet{garg_icl}. However, we found that decision trees exhibit less sensitivity to positional encodings. This is because their feature-based reasoning relies on discrete hierarchical rules rather than point-by-point mappings. By grouping multiple inputs under the same label, decision trees largely bypass positional dependencies and are minimally influenced by untrained positional encodings. 

\subsubsection{Evaluation}
To assess performance, we sample a prompt with a maximum length of \(T_\text{max}\), which is equal to 101 for linear and sparse linear functions and 201 for 2-layer networks. We then trim the prompt to \(T_i \leq T_\text{max}\) demonstrations and track the prediction errors for each \(i \leq T_\text{max}\). Consequently, each point in our error curves corresponds to the error at a specific prompt length \(i\). This analysis is conducted over batches of 256 prompts, with the average error reported. We have determined that batches larger than 256 prompts do not significantly alter the results, confirming that 256 prompts are sufficient to produce generalized results.

\subsubsection{Optimizing the Baseline Methods}
\label{app:fv_opt}
Function Vector (FV) and In-Context Vector (ICV) were originally designed for language data. Preliminary results showed that FV had a notable impact on regression data, with potential for further improvement through fine-tuning. In contrast, ICV slightly underperformed the vanilla model on regression tasks when using the default scaling parameter \(\lambda = 0.1\) from language experiments. This parameter, dominated by activation magnitudes, required tuning for regression data.

Our approach extends the original by integrating FV across multiple GPT-2 layers, leading to improved performance. To minimize output disruption and reinforce tasks, we added \emph{dummy} examples (\emph{e.g.}, \{0, 0\} pairs) at specific prompt positions. Increasing the number of attention heads for FV computation from 10 to 35 further enhanced performance. In contrast, ICV already integrates the task vector at every token position and computes a distinct task vector for each layer, similar to our method. Thus, the only optimization for ICV was tuning the scaling parameter. We found that FV required no scaling (\emph{i.e.}, 1.0). However, ICV often degraded the vanilla model's performance, with a scale of 1.5 being the best option to mitigate this issue. Key baseline optimizations are summarized below:

\begin{enumerate}
    \item \textbf{Function Vector added to multiple layers:} Distributing FV across multiple layers normalized its impact on activations. Layers 6, 7, and 8 of GPT-2 were the most effective.
    \item \textbf{Dummy tokens:} Placing dummy tokens at 0.1, 0.25, 0.5, 0.75, and 0.9 fractions of the prompt length optimized performance for FV.
    \item \textbf{Number of attention heads for FV computation:} Increasing to 35 attention heads maximized GPT-2 performance, with diminishing returns beyond this point.
    \item \textbf{Scaling:} FV performed best with a scaling factor of 1.0, aligning with \citet{fv} for language tasks. ICV required scaling (\emph{e.g.}, 1.5) to counterbalance the dominance of hidden state activations and prevent degradation of the vanilla model's performance.
\end{enumerate}

\subsubsection{Other Baseline Methods}
\label{app:other_baselines}
As explained in Appendix \ref{app:related_work}, \citeposs{icl_creates_task_vectors} and \citeposs{dongfang} methods cannot be included in our experiments because they require in-context prompts to include a separator token (\emph{e.g.}, ``$\rightarrow$'') in each example. This requirement makes them incompatible with our functional regression dataset, as these tasks explicitly avoid tokenization and operate directly on raw floating-point numbers (see Appendix \ref{app:synthetic_exp} for details). Introducing separator tokens fundamentally changes the task by imposing a symbolic or tokenized representation of input-output relationships, contradicting \citeposs{garg_icl} setup. Moreover, such methods rely on the tokenized format, bypassing the challenge of inferring functional relationships from raw numeric data, which renders the comparison invalid.

\subsection{Experiments on Language Tasks}
We closely adhere to the experimental methods described by \citet{fv}, providing all relevant details to ensure our report is self-sufficient.

\subsubsection{Datasets and Tasks}
\label{app:language_tasks}
We provide a detailed overview of the natural language tasks used to evaluate the task formulations. References to the original data sources for each task are included, along with details on how they were refined and transformed into word pairs by \citet{fv}. The final datasets used in our evaluations are provided in the authors' GitHub repository\footref{fv_repo}.

\paragraph{AG News}  
This text classification dataset consists of news headlines and the first few sentences of articles as inputs, with labels indicating the article's category. Categories include Business, Science/Technology, Sports, and World \citep{zhang}.

\paragraph{Antonym and Synonym}  
The dataset is \citeposs{antonym_synonym_dataset} and was further refined by \citet{fv}. Initially, all adjective, noun, and verb pairs from the dataset splits were combined, with duplicate entries removed. The dataset was then modified to include only word pairs where both words are tokenized as single tokens. This process resulted in 2,398 antonym pairs and 2,881 synonym pairs, with a vocabulary size of \(|\mathcal{V}| = 50,400\).  

These datasets originally included multiple outputs for single inputs, \emph{e.g.}, ``increase'' \(\rightarrow\) ``decrease'' and ``increase'' \(\rightarrow\) ``reduce.'' However, handling such cases would require a more powerful model \citep{fv}. Therefore, the dataset has been simplified to ensure a one-to-one mapping between terms.

\paragraph{CommonsenseQA}  
This question-answering dataset requires a model to select the correct answer from five options, each labeled with a letter. The model generates the letter corresponding to the correct answer. For instance, given the question ``Where is a business restaurant likely to be located?'' and the options ``A) town, B) hotel, C) mall, D) business sector, E) yellow pages,'' the model should generate ``D'' \citep{talmor}.

\paragraph{Landmark-Country}  
The Landmark-Country dataset contains entries pairing the name of a landmark with its corresponding country. The data pairs are sourced from \citet{hernandez}.

\paragraph{Person-Instrument}  
The Person-Instrument dataset contains entries pairing the name of a professional musician with the instrument they play. The data pairs are sourced from \citet{hernandez}.

\paragraph{Person-Occupation}  
The Person-Occupation dataset, sourced from \citet{hernandez}, contains entries pairing the names of well-known individuals with their respective occupations.

\paragraph{Person-Sport}  
The Person-Sport dataset, sourced from \citet{hernandez}, contains entries pairing the names of professional athletes with the sports they play.

\paragraph{Product-Company}  
The Product-Company dataset, curated from \citet{hernandez}, contains entries pairing the names of commercial products with the companies that sell them.

\paragraph{Sentiment Analysis}  
The sentiment analysis task is based on the Stanford Sentiment Treebank (SST-2) dataset \citep{socher}, which consists of movie review sentences labeled as either ``positive'' or ``negative.'' For example, an entry from this dataset might be: ``An extremely unpleasant film. \(\rightarrow\) negative.'' We use the subset of SST-2 curated by \citet{honovich}, which excludes incomplete sentences and those with more than 10 words, resulting in 1,167 entries. For further details, see \citeposs{honovich} study.

\paragraph{Translation}  
The language translation dataset was constructed using data from \citet{connenau}, which pairs English words with their translations in French and Spanish. The train and test splits were merged into a single dataset and the cognates were filtered out. This process results in 4,705 pairs for English-French and 5,200 pairs for English-Spanish. The original datasets included multiple translations for some input words. These duplicates were filtered using GPT-4, following the method used for Antonym and Synonym.

The tasks described so far are \textit{Abstractive} NLP tasks, where the information to be generated is not present in the prompt. We also evaluate the opposite case: \textit{Extractive} tasks, where the answer exists within the prompt, and the goal is to retrieve it.

\paragraph{CoNLL-2003}  
Our extractive tasks are based on a subset of the CoNLL-2003 English named entity recognition (NER) dataset \citep{ner}, a widely used NLP benchmark for evaluating NER models. The NER task involves extracting the correct entity from a given sentence based on a specific property. For our study, we use three tasks derived by \citet{fv} from CoNLL-2003: NER-person, NER-location, and NER-organization, where the labels correspond to the name of a person, location, or organization, respectively. Each dataset is constructed by merging the CoNLL-2003 ``train'' and ``validation'' splits into a single dataset and filtering sentences to include only those with a single instance of the specified class. This reduces task ambiguity, as sentences with multiple instances of the same class could have more than one correct answer.

\subsubsection{Prompting}
The default template for prompting the GPT-J model with \(T\) demonstrations is structured as follows:
\begin{equation*}
    \texttt{Q:}\{x_1\}\texttt{\textbackslash nA:}\{y_1\}\texttt{\textbackslash n} \texttt{\textbackslash n} \ldots \texttt{Q:}\{x_\text{query}\}\texttt{\textbackslash nA:}
\end{equation*}
In our experiments with shuffled prompts, we randomly shuffle the labels \(\{y_i\}_{i=1}^T\) among each other. For zero-shot prompts, which contain no demonstrations, prompts consists solely of the query: ``\(\texttt{Q:}\{x_\text{query}\}\texttt{\textbackslash nA:}\)''. When testing Task Vector \citep{icl_creates_task_vectors} and State Vector \citep{dongfang}, which require a separator token between inputs and outputs, we treat ``\texttt{\textbackslash n}'' as the separator token.

\subsubsection{Evaluation}
We uniformly sample in-context examples (for few-shot) and a query input \(N = 256\) times, assessing performance as the accuracy if the first token generated by the model matches the first token of the query label (if the label consists of more than one token). To compute the margin of error (denoted after \(\pm\) in Tables \ref{tab:language_results} and \ref{tab:language_results_full}), we first calculate the standard error of the mean (SEM) as follows:
\begin{equation*}
    \operatorname{SEM} = \frac{\text{standard deviation}}{\sqrt{N}}.
\end{equation*}
Next, the t-critical value is determined using the t-distribution with a 95\% confidence level and degrees of freedom (\(N - 1\)). The margin of error is then calculated as:
\begin{equation*}
    \text{Margin of error} = t_\text{critical} \times \text{SEM}.
\end{equation*}
For example, if the mean accuracy is 39.1 and the margin of error is 5.2, the true mean lies between 33.9 and 44.3 with a 95\% confidence. Thus, if the confidence intervals of different methods overlap with the maximum mean accuracy, it indicates that the differences in performance are not statistically significant despite the highest accuracy.

\subsection{Training Learnable Task Vector}
The LTV parameters are initialized by sampling from the standard Gaussian distribution, totaling \(L \times J\) learnable parameters. We did not employ additional techniques such as dropout, activation functions, or gradient clipping in learning the weights, which are neither clipped nor bounded. The Adam optimizer, with a learning rate of \(5 \times 10^{-5}\), was used in all experiments. Training is terminated if the validation loss does not decrease for 50 consecutive gradient steps. The transformer parameters always remain frozen.

\subsubsection{Learning from Regression Data}
\label{app:ltv_training}
We optimize the LTV weights using mini-batch gradient descent on a dataset we compiled, consisting of \(100 \times 256 = 25,600\) function samples (100 times the batch size). Prompts of length \(T_\mathbf{v} > T_\text{train}\) are constructed for these functions. We reserve the 20\% of the dataset as a validation set for monitoring loss. For each mini-batch of prompts \(\{p_i^{f_i}\}_{i = 1}^N\) sampled from the dataset, the gradient descent step is defined as
\begin{gather*}
    \mathcal{L}(\Phi) = \frac{1}{N} \sum_{i = 1}^{N} \Big(\Tilde{M}_\theta \big(p_i^{f_i} \mid \mathbf{v}^{f_i}_\Phi \big) - f_i(x_{i, \text{query}})\Big)^2,  \\
    \Phi \leftarrow \Phi - \eta \cdot \nabla \mathcal{L}(\Phi),
\end{gather*}
where \(p_i^{f_i}\) represents a prompt corresponding to a unique function \(f_i\) and \(\eta\) is the learning rate. This method and dataset compilation are applied uniformly across all three function classes.

We recognize that generating distinct functions and prompts at each gradient step could potentially provide an infinite variety of data and functions. This raises concerns about whether LTV overfits to the function classes. Although it was argued that the likelihood of the model encountering a training-similar prompt is extremely low \citep{garg_icl}, we opted for a static dataset approach. Our experiments were conducted using this dataset, with evaluations performed on the prompts constructed from different weight vectors during inference.

\subsubsection{Learning from Language Data}
Given the dataset \(P^t\) for task \(t\), each gradient step involves sampling 100 prompts with replacement, each containing 5 demonstrations. These prompts are processed by the transformer to collect attention activations, and the mean of these activations across the sampled prompts is computed. This mean is then passed through the LTV layer to compute the corresponding LTV, as illustrated in \eqref{fig:ltv_inference}. 

For training, we sample a batch of 32 prompts, each with 5 demonstrations, but with all labels shuffled, rendering the input-output pairs non-informative. The LTV weights are updated to maximize the probability of the correct label for the query input:
\begin{equation*}
    \mathcal{L}(\Phi) = - \frac{1}{N} \sum_{i = 1}^N \log \Big( \Tilde{M}_\theta\big(y_\text{query} \mid \Tilde{p}_i^t; \mathbf{v}_\Phi^t\big) \Big).
\end{equation*}

\subsection{Training LoRA}
{We use LoRA (Low-Rank Adaptation) with a rank of 8, applied to the key (\(K\)), query (\(Q\)), value (\(V\)), and output projection matrices in the self-attention mechanism of the GPT-2 model (12 layers, 8 attention heads per layer, and 256 hidden dimensions). The GPT-2 model is fine-tuned with LoRA on the same static dataset and over the same number of iterations (2000) used for training LTV.

Each of these matrices originally has dimensions \(256 \times 256\). LoRA introduces two low-rank matrices, \(A \in \mathbb{R}^{256 \times 8}\) and \(B \in \mathbb{R}^{8 \times 256}\), replacing the trainable weight matrix with their product. This setup results in 4096 trainable parameters per matrix, calculated as \(2 \times 256 \times 8\).

Since our task is relatively simple and the model is extremely lightweight, we apply LoRA to all layers for comprehensive coverage. This adds 16,384 trainable parameters per layer (\(4 \times 4096\)), as LoRA operates on four matrices (\(K\), \(Q\), \(V\), and output) in each transformer layer. With 12 transformer layers in total, the model fine-tunes 196,608 parameters (\(12 \times 16384\)). This parameter-efficient approach significantly reduces the training cost compared to updating all model parameters.

\subsection{Detailed Ablation Studies}
\label{app:ablation_details}
Our argument is based on the premise that while earlier layers build foundational representations, the most refined and actionable insights for predictions are concentrated in the outputs of the last layer. The motivation for these studies is to examine the model’s resilience to variations in prompt length.

To this end, we freeze all transformer and LTV parameters to analyze the stability of last hidden state distributions across varying prompt lengths, rendering stationary distribution. We generate a dataset of 25,600 prompts (100 times the batch size) with a maximum length \(T_\text{max}\). Prompts were trimmed to \(T_{\text{train}}\) for the vanilla transformer and extended beyond \(T_{\text{train}}\) for the FV- and LTV-integrated transformers. The last hidden states from these configurations were compiled into two datasets, \(X_1\) and \(X_2\), corresponding to the vanilla transformer and the FV- and LTV-integrated transformers, respectively. The samples within each dataset are independent and identically distributed, allowing us to estimate their probability distributions.

We estimate the probability distributions using Kernel Density Estimation (KDE), employing the standard KDE implementation from \texttt{scipy} \citep{scipy} with default settings. However, directly estimating a probability distribution in a high-dimensional space often leads to the \textit{curse of dimensionality}, where the volume of data required to effectively estimate the distribution grows exponentially with the number of dimensions. A practical solution to this challenge is to employ SVD for dimensionality reduction. This involves decomposing the data matrices \(X_i \in \mathbb{R}^{M \times d}\) as:
\begin{align*}
    X_1 &= U_1 \Sigma_1 V_1^\top, \\
    X_2 &= U_2 \Sigma_2 V_2^\top,
\end{align*}
where \(M\) is the number of collected samples and \(U_i\) contains the principal components \(u_{i,1}, u_{i,2}, \ldots, u_{i,n}\) as column vectors. These principal components (PCs) form the column space of \(X_i\):
\begin{equation*}
    \operatorname{span}(u_{i,1}, u_{i,2}, \ldots, u_{i,n}) = \operatorname{colspace}(X_i),
\end{equation*}
where each \(u_{i, k}\) is orthogonal to \(u_{i, k^\prime \neq k}\) and ordered by decreasing variance that they explain. Specifically, the first \(n\) principal components represent the directions along which the data varies the most, capturing the most significant patterns in the data. These components are likely more informative and relevant for distinguishing different behaviors or properties of the data. 

Rather than estimating the distribution underlying the entire datasets \(X_i\) as multivariate distributions, we employ Gaussian KDE\footnote{\url{https://docs.scipy.org/doc/scipy/reference/generated/scipy.stats.gaussian_kde.html}} to estimate each PC as a unimodal distribution. This approach is advantageous since KDE performs better with univariate data. However, transitioning from multivariate to univariate requires the assumption that the PCs are uncorrelated. We validate this assumption by observing that the nondiagonal entries of the correlation matrices of \(U_i\) are on the order of \(10^{-2}\), with diagonal entries being approximately 1, effectively an identity function, confirming that the PCs are indeed uncorrelated.

\begin{figure}[htbp]
  \centering
    \includegraphics[width=0.99\linewidth]{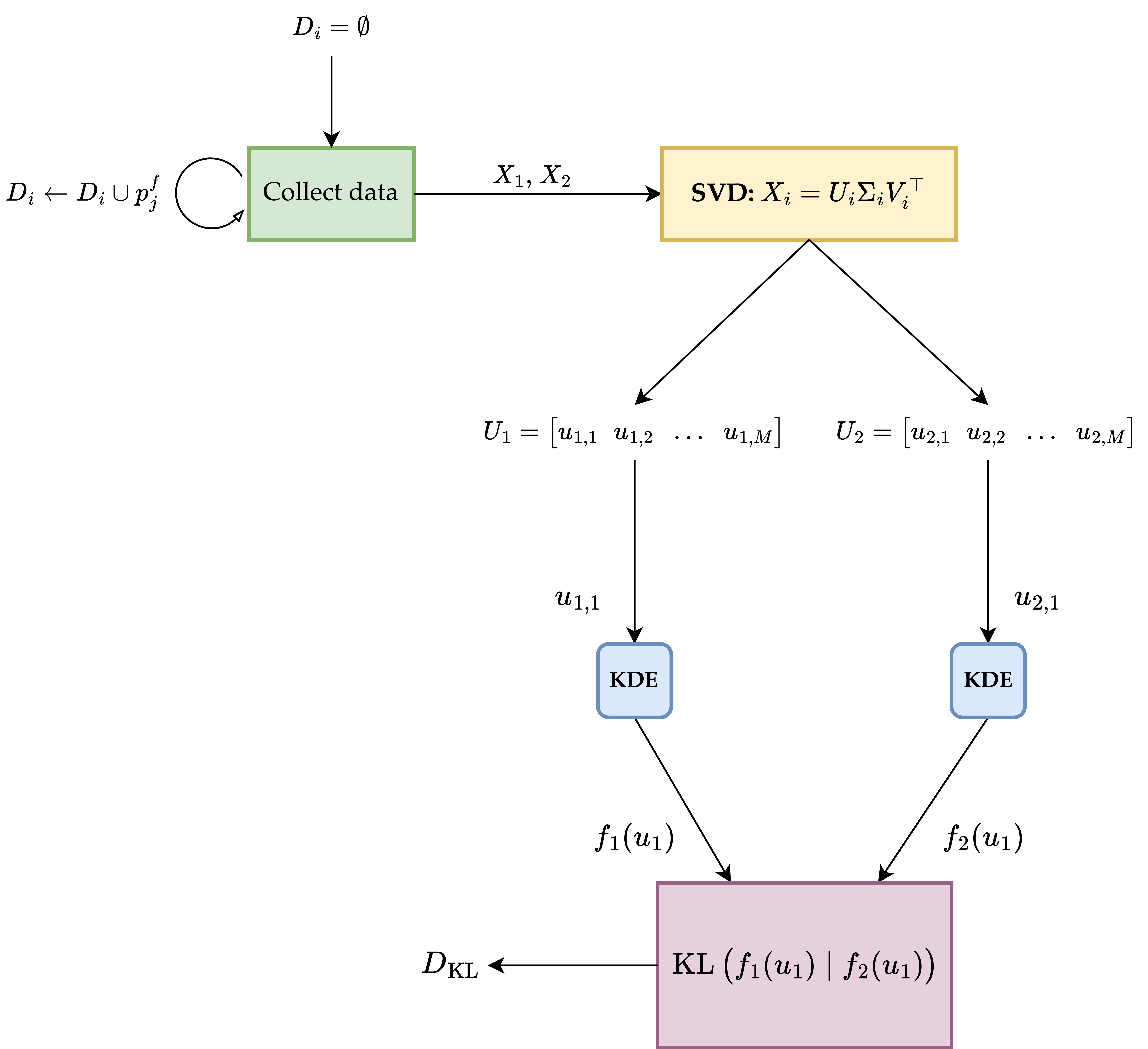}
    \caption{The diagram illustrates our pipeline for ablation studies. We start by collecting \(M = 25,600\) prompts corresponding to a selected task \(f\). Subsequently, the first principal component of the column space of these datasets is extracted using SVD. Finally, we report the KL divergence between the KDE-estimated distributions of these components.}
\label{fig:ablations}
\end{figure}

We use KL divergence between the KDE-estimated distributions of the column pairs to quantitatively assess distributional similarities. However, we find that these divergence values are negligibly small, except for the first one, which accounts for the most variance within the dataset. The negligible divergence scores for higher-order components suggest that these vectors contribute minimally to differentiating the datasets. Thus, focusing on the first component, which shows substantial divergence, is statistically justified and highlights the critical variations relevant to model generalization.

This experimentation process is depicted in Figure \ref{fig:ablations}. Additionally, histograms illustrating the KDE-estimated distributions of the first principal components are provided in Appendix \ref{app:ablation_results} to offer a clear view of their similarity.

\subsection{Computational Resources}
\label{sec:computational_resources}
The computational experiments were conducted on a high-performance system with an AMD Ryzen Threadripper PRO 3995WX processor and 515 GB of RAM. For GPU acceleration, two NVIDIA RTX A6000 GPUs were used, each with 49 GB of memory. This setup efficiently supported the transformer models used in our research.

\begin{figure*}[!hpbt]
    \centering
    \begin{align*}
        &\text{\small {\blue} Transformer} \quad &&\text{\small {\red} Transformer + LoRA} \quad &&\text{\small {\green} Transformer + FV (optimized)} \\
        &\text{\small {\purple} Transformer + ICV (tuned)} \quad &&\text{\small {\orange} Transformer + LTV} \quad &&
    \end{align*}
	\subfigure[\phantom{(a)} \(T_\mathbf{v} = 41\)]{
                \includegraphics[width=0.25\linewidth]{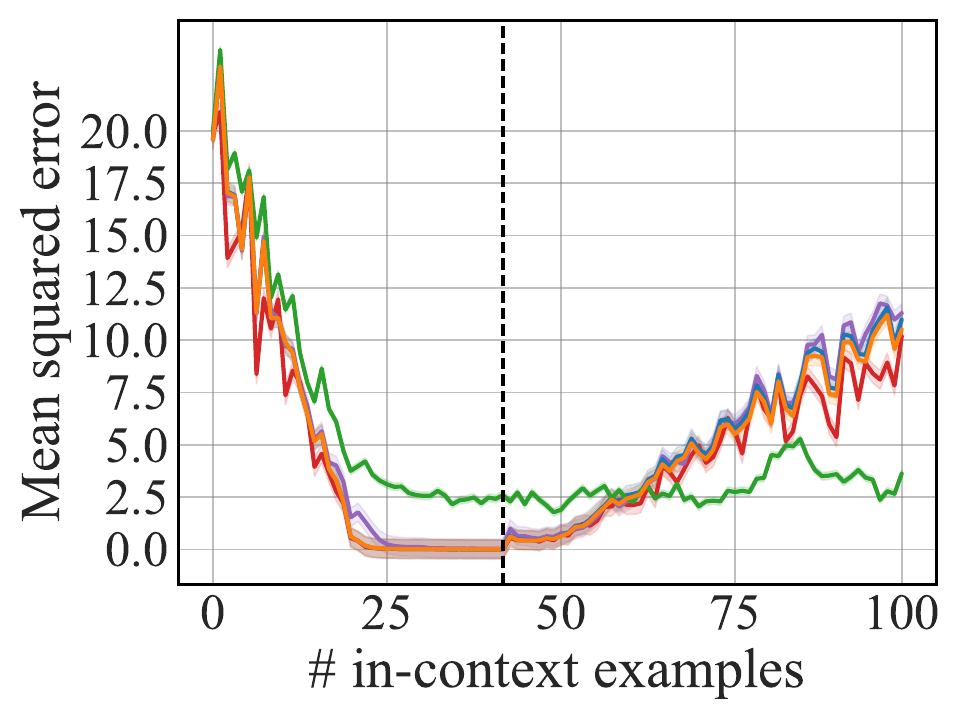}%
        }
        \subfigure[\phantom{(a)} \(T_\mathbf{v} = 42\)]{
                \includegraphics[width=0.25\linewidth]{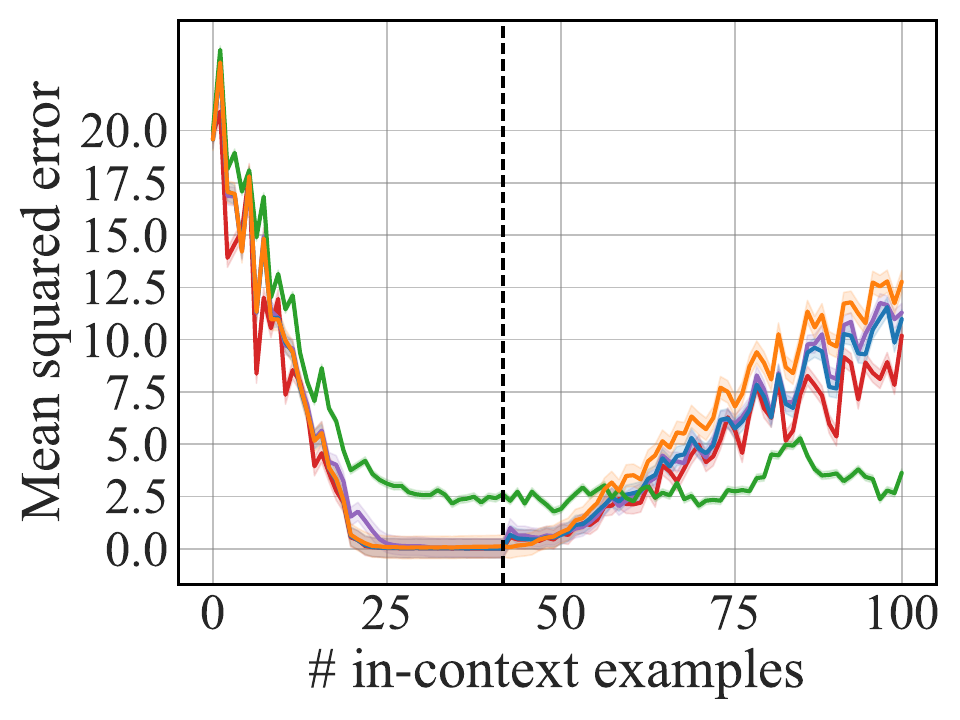}%
        }
        \subfigure[\phantom{(a)} \(T_\mathbf{v} = 56\)]{
                \includegraphics[width=0.25\linewidth]{figures/inference/linear_regression/seq_len_56.pdf}%
        }%
        \hfill
        \subfigure[\phantom{(a)} \(T_\mathbf{v} = 71\)]{
                \includegraphics[width=0.25\linewidth]{figures/inference/linear_regression/seq_len_71.pdf}%
        }
        \subfigure[\phantom{(a)} \(T_\mathbf{v} = 86\)]{
                \includegraphics[width=0.25\linewidth]{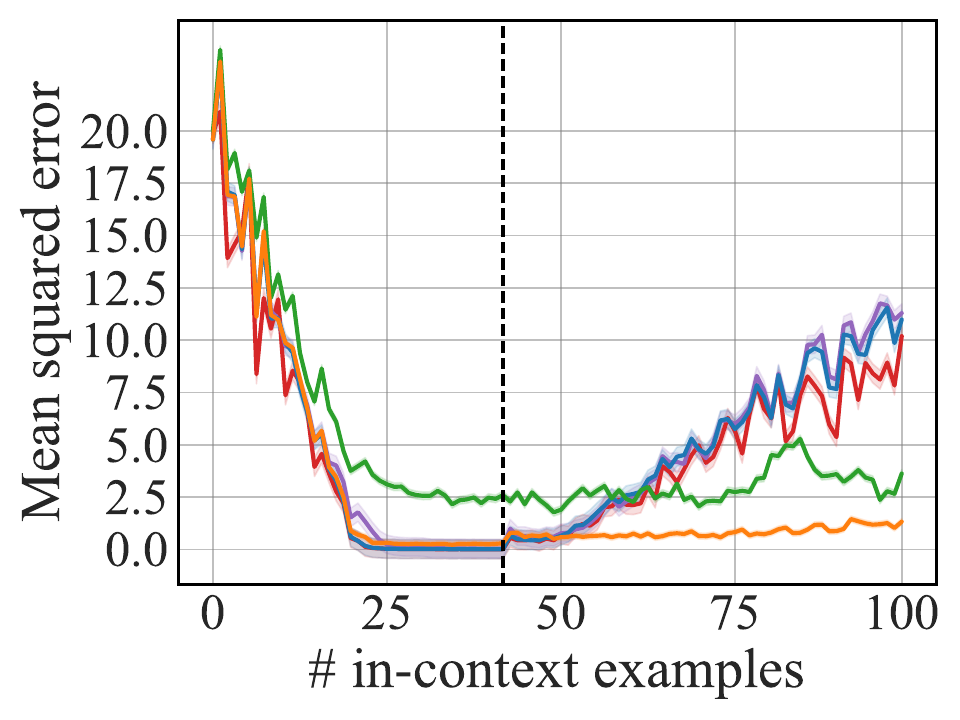}%
        }
        \subfigure[\phantom{(a)} \(T_\mathbf{v} = 101\)]{
                \includegraphics[width=0.25\linewidth]{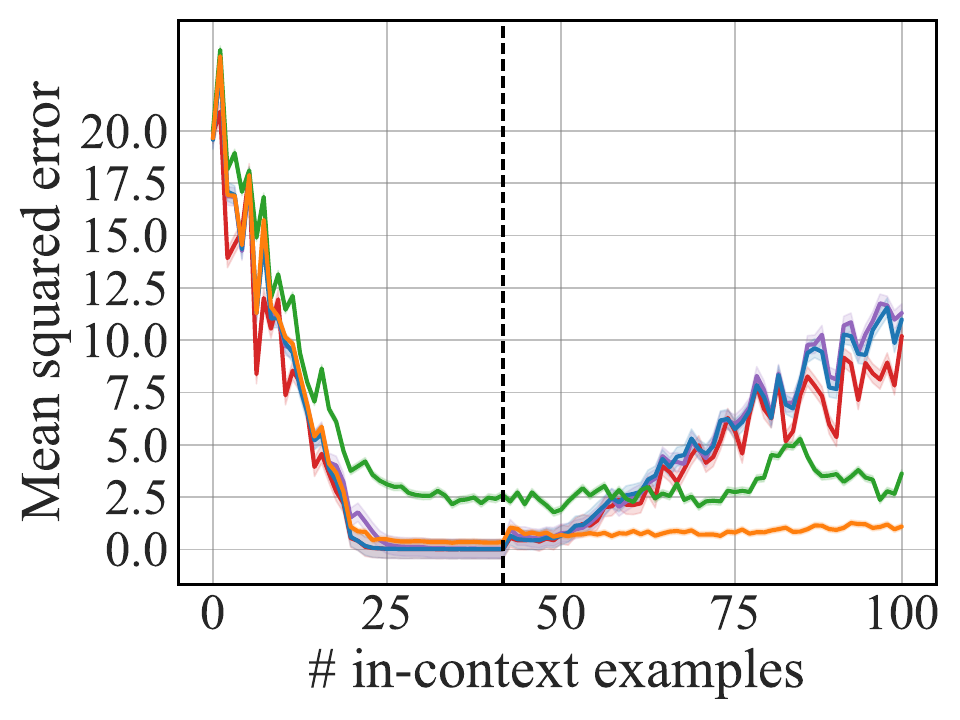}%
        }
    \caption{Evaluation on the class of linear functions, with the transformer pre-trained on up to \(T_\text{train} = 41\) examples per prompt. Results are averaged over a batch of 256 randomly selected tasks. The shaded area represents the 95\% confidence interval over the sampled prompts. \(T_\mathbf{v}\) denotes the prompt length used during LTV training.}
    \label{fig:complete_results_lin}
\end{figure*}

\section{Complete Set of Results}
\label{app:complete_results}
We present the complete set of results that could not be included in the main body due to page limitation. Specifically, this section includes:
\begin{enumerate}
    \item \textbf{Regression Tasks:} Evaluations with different LTV training lengths $T_\mathbf{v}$
    \item \textbf{Regression Tasks:} Evaluations with chain-of-thought prompting based on the findings of \citet{cot_relu_nets}
    \item \textbf{Regression Tasks:} Evaluations under distributional shifts
    \item \textbf{NLP Tasks:} Accuracy scores (with zero- and few-shot prompting) across 15 NLP benchmarks
    \item \textbf{Ablation Studies:} Complete ablation results, including KL divergence scores across different $T_\mathbf{v}$ and the corresponding histograms illustrating the distributions of the optimally in-context-learned and LTV-integrated models
\end{enumerate}

\subsection{Evaluations on Regression Tasks}  
\label{app:complete_regression_evals}
The loss curves are shown in Figures \ref{fig:complete_results_lin}, \ref{fig:complete_results_sparse_lin}, and \ref{fig:complete_results_relu}. As expected, no performance improvement is observed for \(T_\mathbf{v} = T_\text{train}\), as the transformer already achieves optimal performance. For visual comparison, the LTV trained with the maximum number of examples is included, though it does not provide insights into generalizability, as it is exposed to all \(T_\text{max}\) examples.

\begin{figure*}[!tbh]
    \centering
    \begin{align*}
        &\text{\small {\blue} Transformer} \quad &&\text{\small {\red} Transformer + LoRA} \quad &&\text{\small {\green} Transformer + FV (optimized)} \\
        &\text{\small {\purple} Transformer + ICV (tuned)} \quad &&\text{\small {\orange} Transformer + LTV} \quad &&
    \end{align*}
	\subfigure[\phantom{(a)} \(T_\mathbf{v} = 41\)]{
                \includegraphics[width=0.25\linewidth]{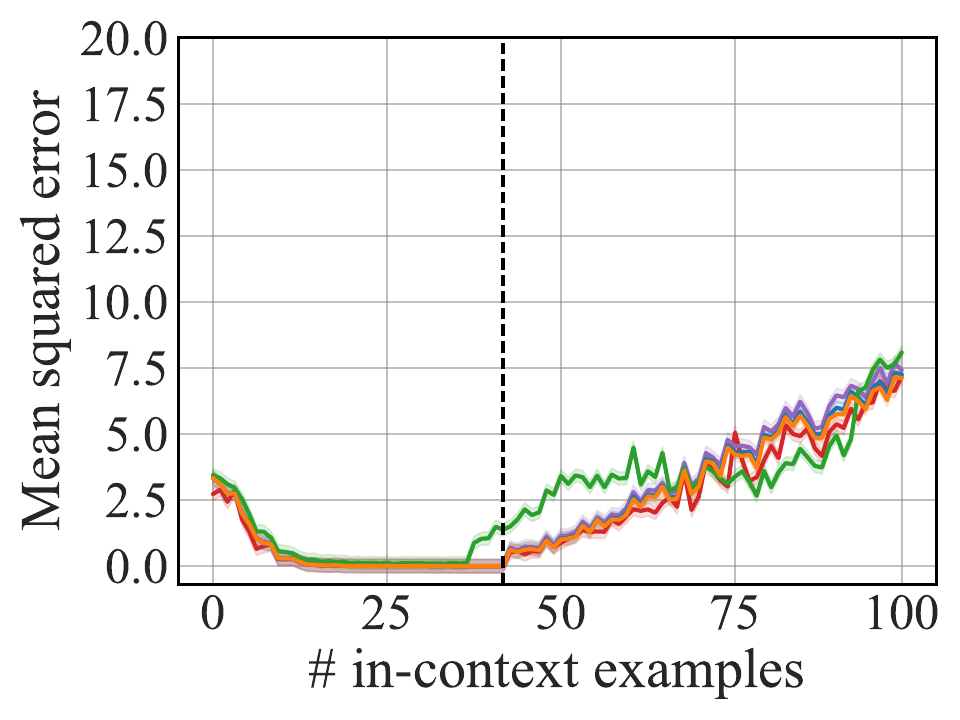}%
        }
        \subfigure[\phantom{(a)} \(T_\mathbf{v} = 42\)]{
                \includegraphics[width=0.25\linewidth]{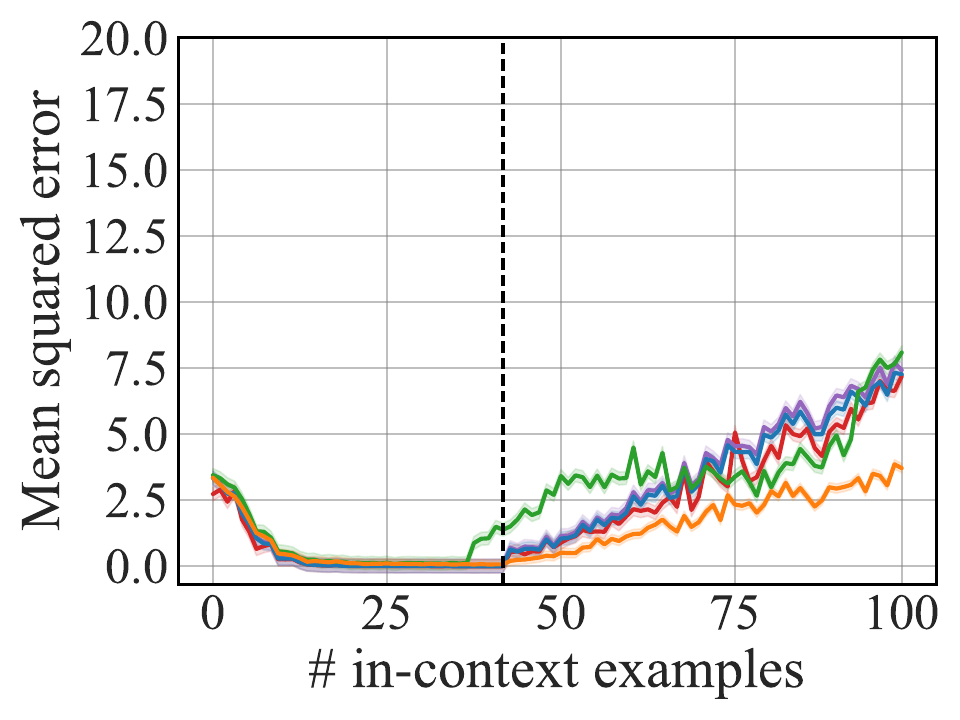}%
        }
        \subfigure[\phantom{(a)} \(T_\mathbf{v} = 56\)]{
                \includegraphics[width=0.25\linewidth]{figures/inference/sparse_linear_regression/seq_len_56.pdf}%
        }%
        \hfill
        \subfigure[\phantom{(a)} \(T_\mathbf{v} = 71\)]{
                \includegraphics[width=0.25\linewidth]{figures/inference/sparse_linear_regression/seq_len_71.pdf}%
        }
        \subfigure[\phantom{(a)} \(T_\mathbf{v} = 86\)]{
                \includegraphics[width=0.25\linewidth]{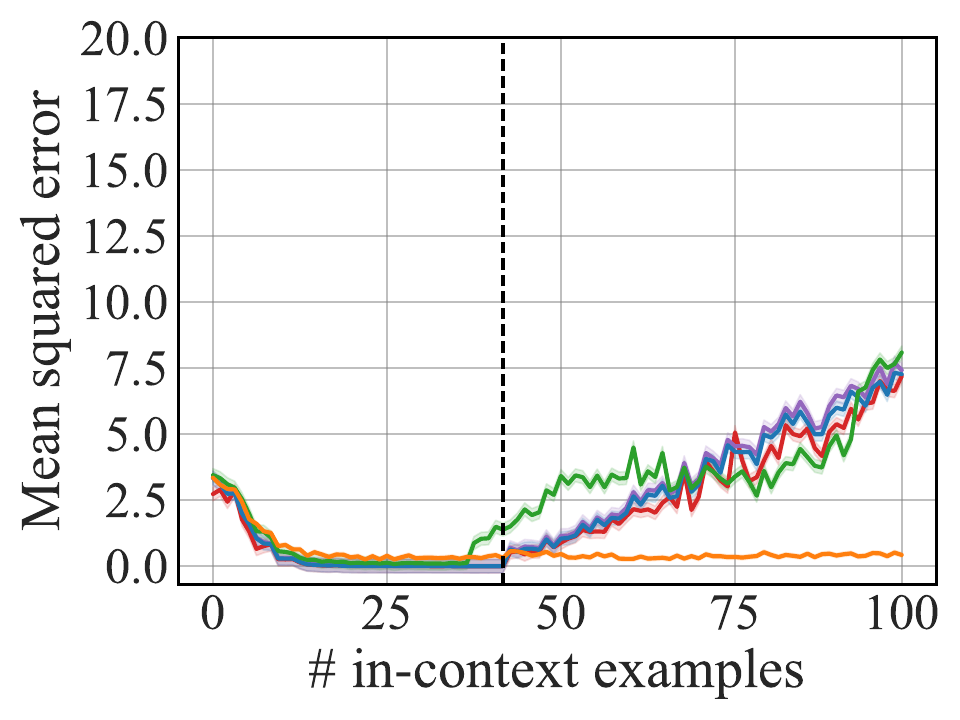}%
        }
        \subfigure[\phantom{(a)} \(T_\mathbf{v} = 101\)]{
                \includegraphics[width=0.25\linewidth]{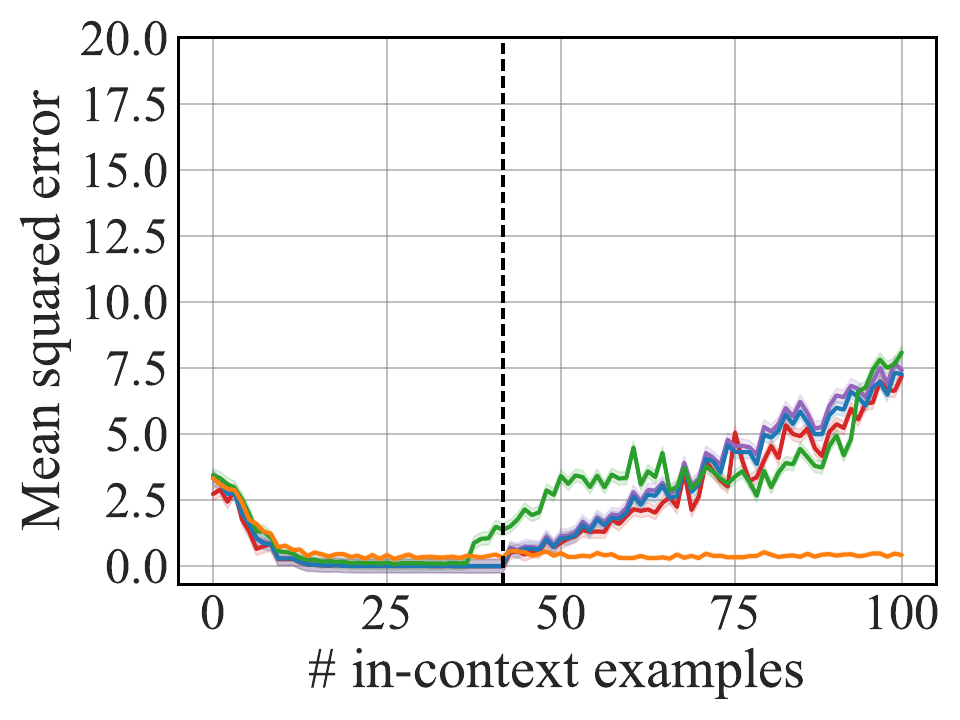}%
        }
    \caption{Evaluation on the class of sparse linear functions, with the transformer pre-trained on up to \(T_\text{train} = 41\) examples per prompt. Results are averaged over a batch of 256 randomly selected tasks. The shaded area represents the 95\% confidence interval over the sampled prompts. \(T_\mathbf{v}\) denotes the prompt length used during LTV training.}
    \label{fig:complete_results_sparse_lin}
\end{figure*}

\begin{figure*}[!tbh]
    \centering
    \begin{align*}
        &\text{\small {\blue} Transformer} \quad &&\text{\small {\red} Transformer + LoRA} \quad &&\text{\small {\green} Transformer + FV (optimized)} \\
        &\text{\small {\purple} Transformer + ICV (tuned)} \quad &&\text{\small {\orange} Transformer + LTV} \quad &&
    \end{align*}
	\subfigure[\phantom{(a)} \(T_\mathbf{v} = 101\)]{
                \includegraphics[width=0.25\linewidth]{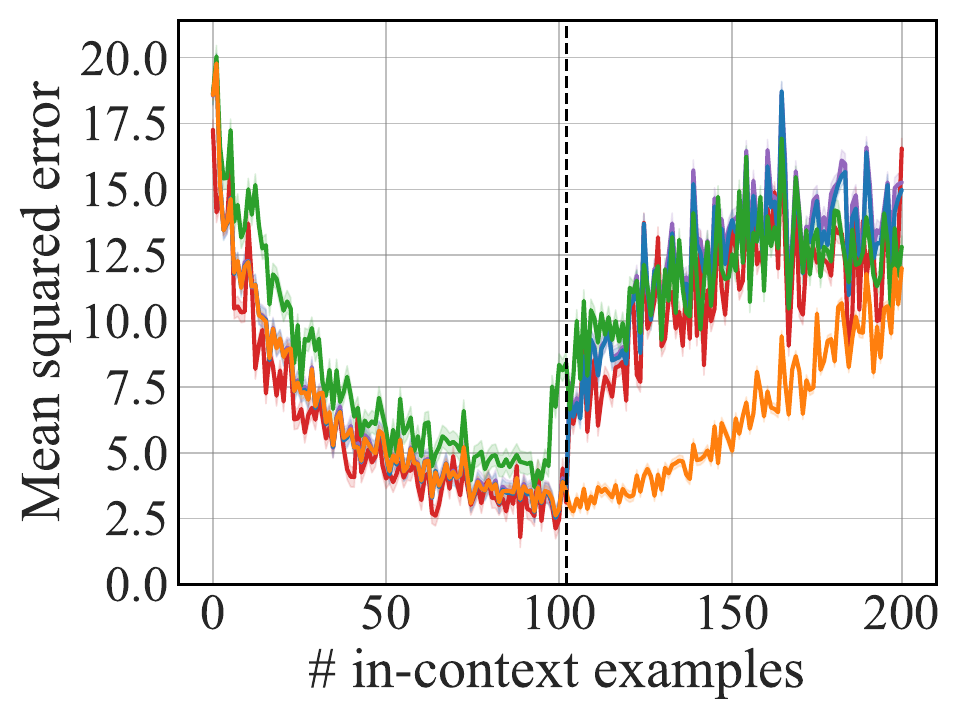}%
        }
        \subfigure[\phantom{(a)} \(T_\mathbf{v} = 102\)]{
                \includegraphics[width=0.25\linewidth]{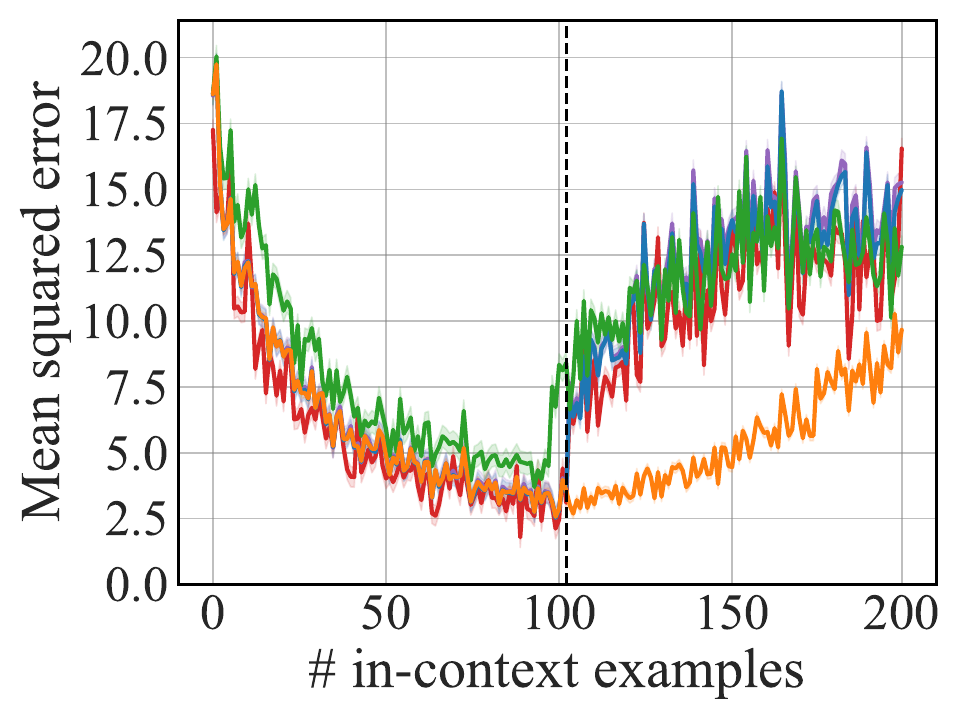}%
        }
        \subfigure[\phantom{(a)} \(T_\mathbf{v} = 126\)]{
                \includegraphics[width=0.25\linewidth]{figures/inference/relu_2nn_regression/seq_len_126.pdf}%
        }%
        \hfill
        \subfigure[\phantom{(a)} \(T_\mathbf{v} = 151\)]{
                \includegraphics[width=0.25\linewidth]{figures/inference/relu_2nn_regression/seq_len_151.pdf}%
        }
        \subfigure[\phantom{(a)} \(T_\mathbf{v} = 176\)]{
                \includegraphics[width=0.25\linewidth]{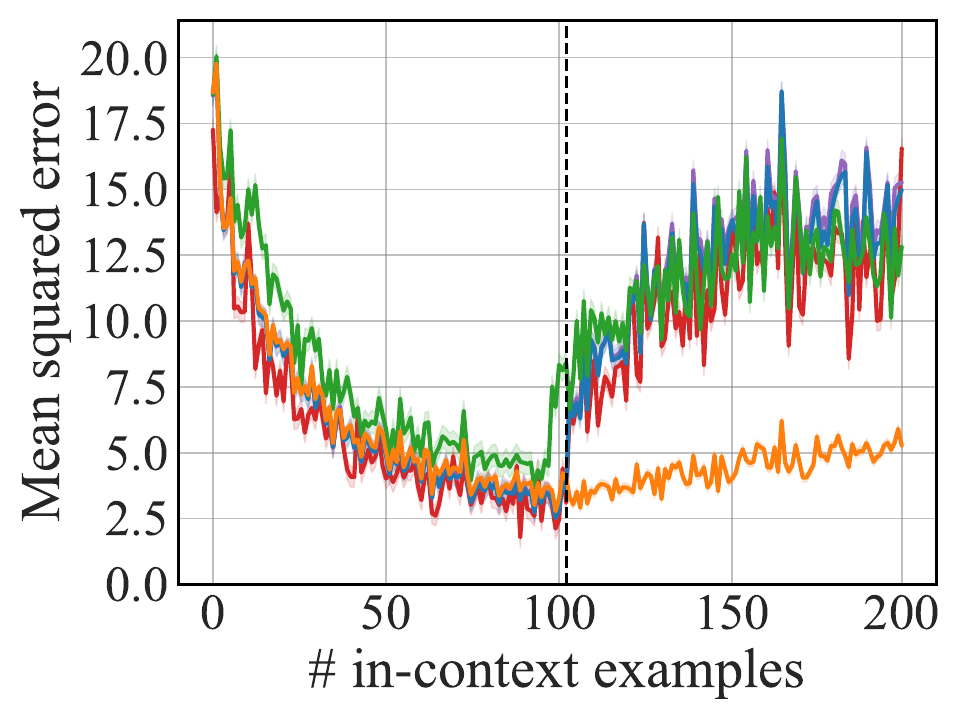}%
        }
        \subfigure[\phantom{(a)} \(T_\mathbf{v} = 201\)]{
                \includegraphics[width=0.25\linewidth]{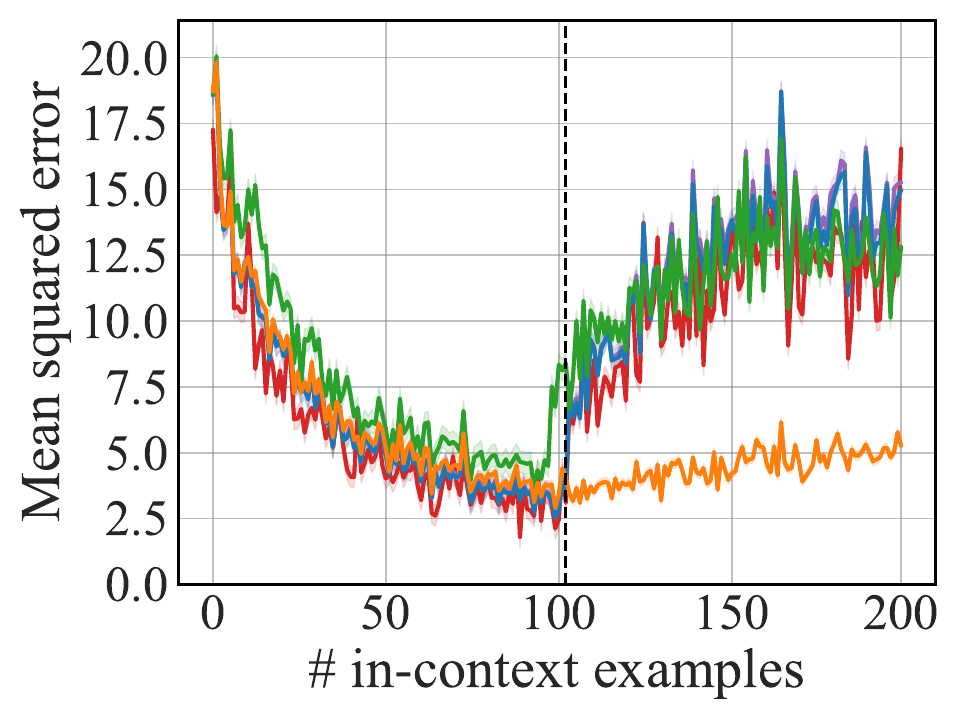}%
        }
    \caption{Evaluation on the class of 2-layer ReLU neural networks, with the transformer pre-trained on up to \(T_\text{train} = 101\) examples per prompt. Results are averaged over a batch of 256 randomly selected tasks. The shaded area represents the 95\% confidence interval over the sampled prompts. \(T_\mathbf{v}\) denotes the prompt length used during LTV training.}
    \label{fig:complete_results_relu}
\end{figure*}

\subsection{Chain-of-Thought Prompting}

\citet{cot_relu_nets} investigate how CoT enhances ICL for compositional functions, formalizing it as a two-phase process: filtering relevant data and applying ICL on the filtered input. During filtering, the model focuses on the relevant tokens in the prompt based on a given instruction. It then processes the filtered prompt to generate an intermediate output before proceeding to the next step in the sequence.

Multi-layer networks can be viewed as compositional functions since the output depends on the hidden layers, which are functions of the input. Accordingly, \citet{cot_relu_nets} specifically study 2-layer ReLU networks, as in our work. In this context, the relevant tokens at the $k$-th step correspond to the output of the $k$-th hidden layer. For instance, let $z_k$ represent the $k$-th intermediate step. During inference, the relevant tokens for $z_k$ are $\{z_{k-1}^i, z_{k}^i\}_{i=1}^T$, where $T$ is the number of examples in the prompt.

Building on this, we demonstrate that LTV can generalize to capture latent representations at intermediate steps, enabling task fidelity in our length generalization benchmark using \citeposs{cot_relu_nets}'s proposed CoT method. This provides a more flexible and scalable framework for compositional reasoning.

\begin{figure*}[!hptb]
    \centering
    \begin{align*}
        &\text{\small {\blue} Transformer} \quad &&\text{\small {\orange} Transformer + LTV} \quad &&\text{\small {\green} Transformer + CoT-I} \\ &\text{\small {\red} Transformer + CoT-I/O} \quad &&\text{\small {\purple} Transformer + CoT-I + \underline{LTV}} \quad &&\text{\small {\brown} Transformer + CoT-I/O + \underline{LTV}}
    \end{align*}
	\subfigure[\phantom{(a)} \(T_\mathbf{v} = 126\)]{
                \includegraphics[width=0.32\linewidth]{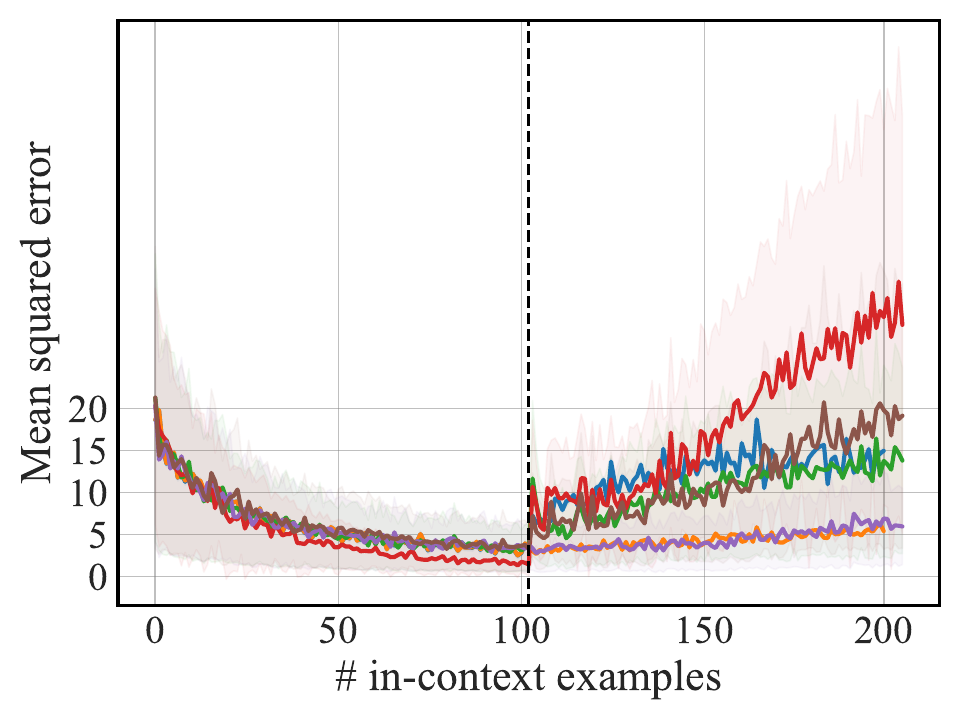}%
        }
        \subfigure[\phantom{(a)} \(T_\mathbf{v} = 151\)]{
                \includegraphics[width=0.32\linewidth]{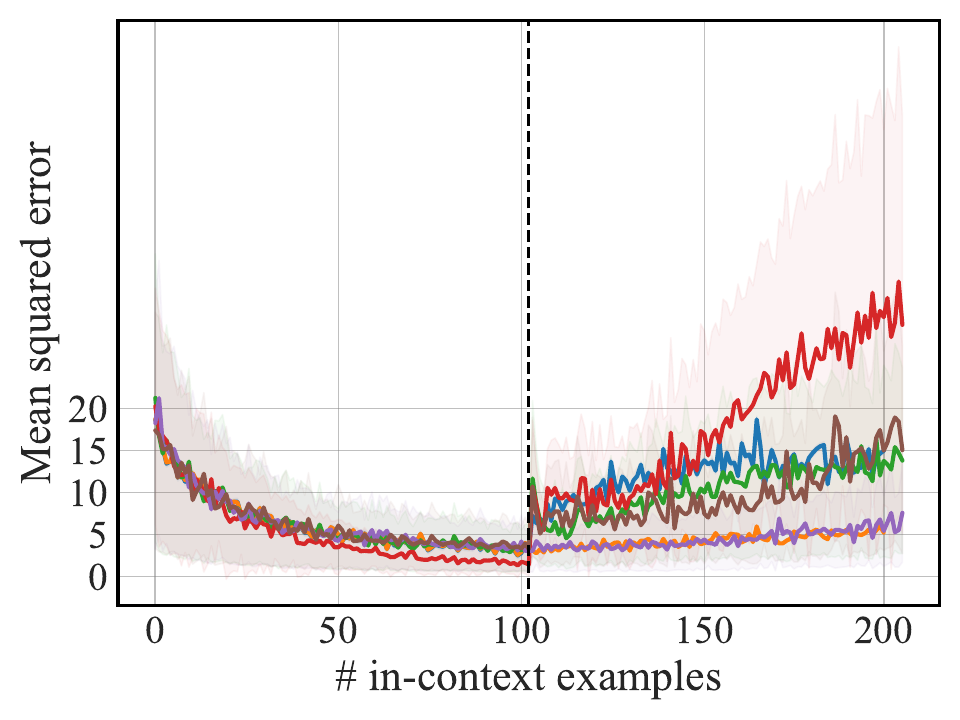}%
        }
        \subfigure[\phantom{(a)} \(T_\mathbf{v} = 176\)]{
                \includegraphics[width=0.32\linewidth]{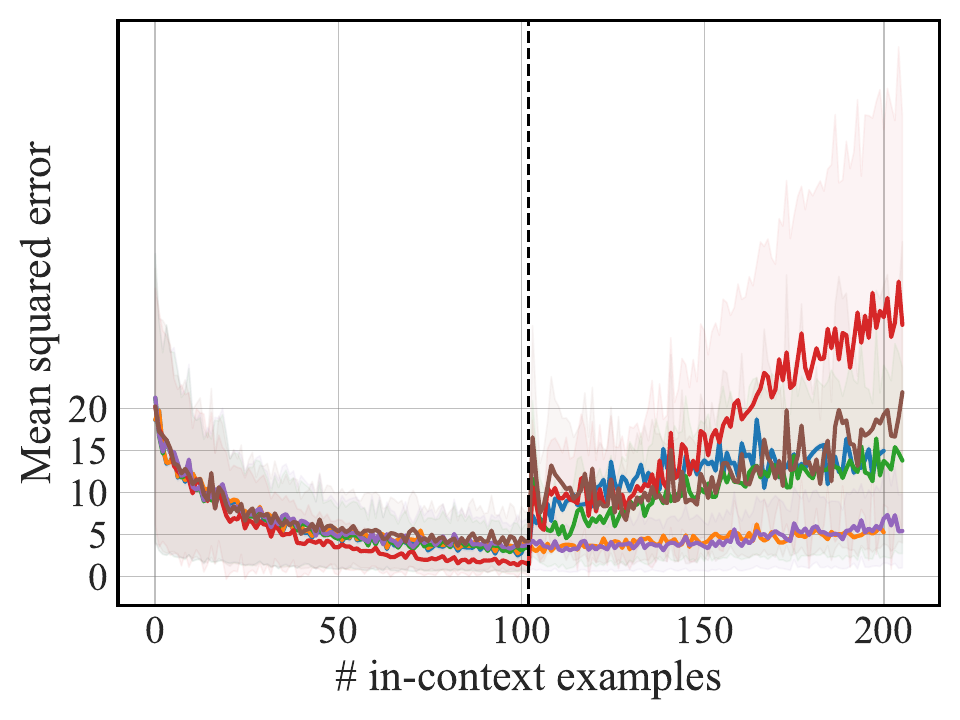}%
        }
    \caption{Evaluation on the class of 2-layer ReLU neural networks, with the transformer pre-trained on up to \(T_\text{train} = 101\) chain-of-thought examples per prompt. Results are averaged over a batch of 256 randomly selected tasks. The shaded area represents the 95\% confidence interval over the sampled prompts. \(T_\mathbf{v}\) denotes the prompt length used during LTV training.}
    \label{fig:cot_results}
\end{figure*}

\subsubsection{Experimental Details}

\paragraph{Model Training} As in Appendix \ref{app:model_training_synth}, we pre-train a custom GPT-2 model (with the same configuration) separately for the two variants of \citet{cot_relu_nets}'s method: CoT-I and CoT-I/O. The task involves 2-layer ReLU network regression, \emph{i.e.}, \( f(x) = W_2 \text{ReLU}(W_1 x) \). The prompt is structured as \( p^f = \{x_1, g(x_1), f(x_1), \dots, x_T, g(x_T), f(x_T)\} \), where \( g(x) = \text{ReLU}(W_1 x) \) represents an intermediate computation.

For each \( i \leq T \), CoT-I/O performs iterative two-step predictions:  
\begin{gather}
 \hat{s}^{(1)}_i = M_\theta\big(p_\mathrm{CoT}(i)\big), \nonumber \\
 \hat{y}_i = \hat{s}^{(2)}_i = M_\theta\big( \{p_\mathrm{CoT}(i), \hat{s}^{(1)}_i\} \big). \label{eq:cot_io_final_pred}
\end{gather}
In contrast, CoT-I directly predicts the final output in a single step:  
\[
\hat{y}_i = M_\theta\big( p_\mathrm{CoT}(i) \big).
\]
The objective functions for pre-training the GPT-2 model are expressed for CoT-I/O and CoT-I respectively as: 
\begin{multline*}
\text{\small $
    \min_\theta \mathbb{E}_{f \sim \mathcal{D}_\mathcal{F}, x \sim \mathcal{D}_\mathcal{X}} \bigg[ \frac{1}{T+1} \sum_{i=0}^{T} \big(\hat{s}^{(1)}_{i+1} - g(x_{i+1})\big)^2$} \\ \text{\small $+ \big(\hat{s}^{(2)}_{i+1} - f(x_{i+1})\big)^2 \bigg],$ \normalsize} \\
\text{\small $
    \min_\theta \mathbb{E}_{f \sim \mathcal{D}_\mathcal{F}, x \sim \mathcal{D}_\mathcal{X}} \bigg[ \frac{1}{T+1} \sum_{i=0}^{T} \big(\hat{y}_{i+1} - f(x_{i+1})\big)^2 \bigg].$ \normalsize}  
\end{multline*}

\begin{figure*}[!tbh]
    \centering
    \begin{align*}
        &\text{\small {\blue} Transformer} \qquad &&\text{\small {\green} Transformer + FV (optimized)} \\ &\text{\small {\purple} Transformer + ICV (tuned)} \qquad &&\text{\small {\orange} Transformer + LTV}
    \end{align*}
	\subfigure[\phantom{(a)} \(T_\mathbf{v} = 41\)]{
                \includegraphics[width=0.25\linewidth]{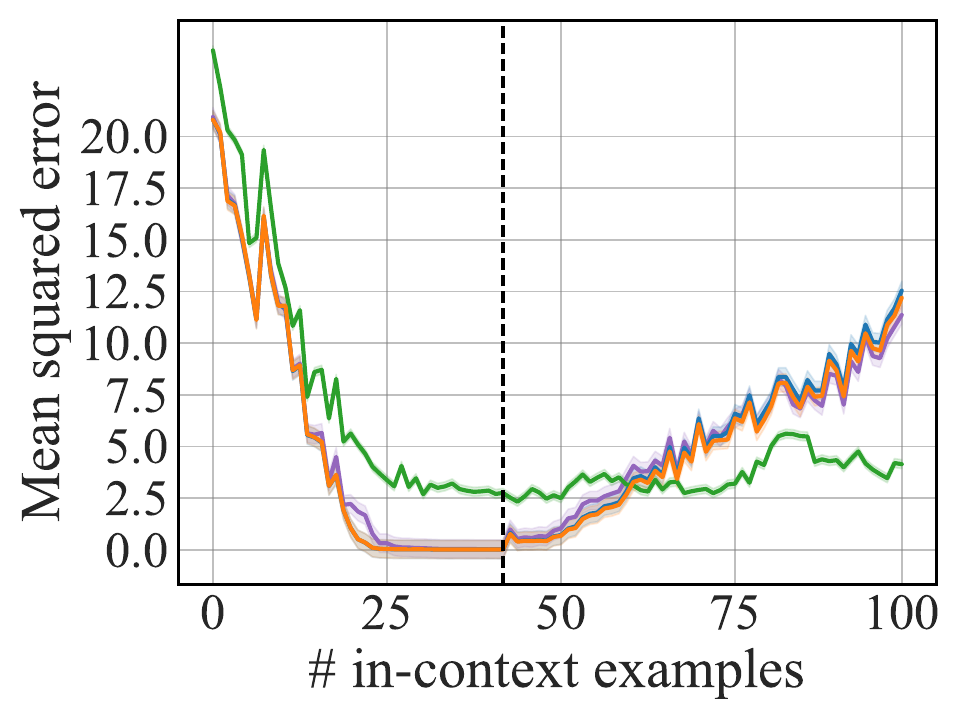}%
        }
        \subfigure[\phantom{(a)} \(T_\mathbf{v} = 42\)]{
                \includegraphics[width=0.25\linewidth]{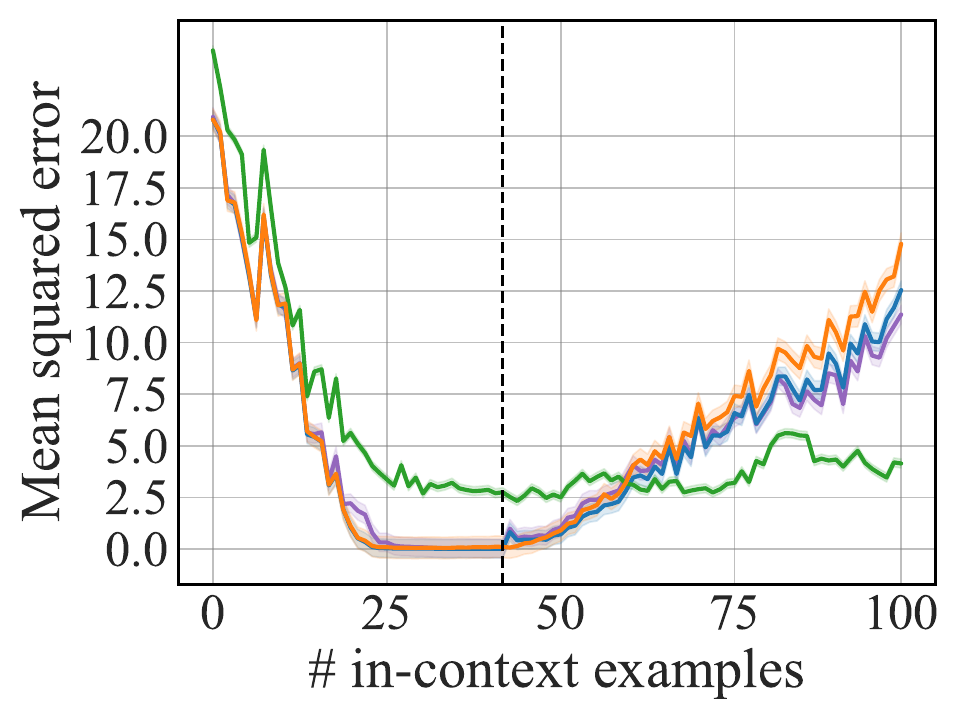}%
        }
        \subfigure[\phantom{(a)} \(T_\mathbf{v} = 56\)]{
                \includegraphics[width=0.25\linewidth]{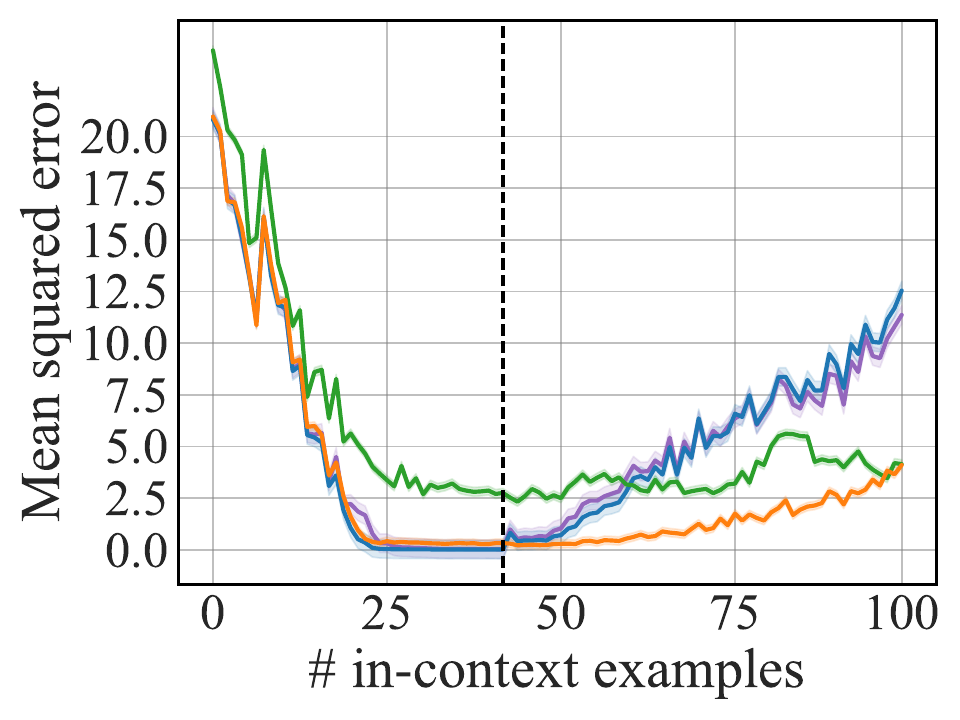}%
        }%
        \hfill
        \subfigure[\phantom{(a)} \(T_\mathbf{v} = 71\)]{
                \includegraphics[width=0.25\linewidth]{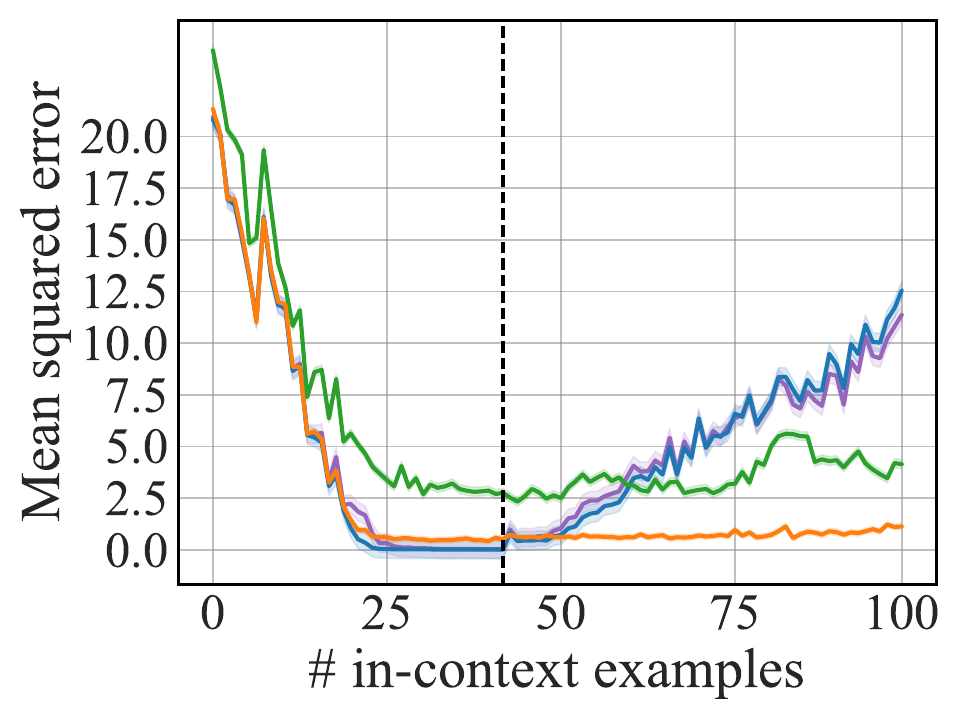}%
        }
        \subfigure[\phantom{(a)} \(T_\mathbf{v} = 86\)]{
                \includegraphics[width=0.25\linewidth]{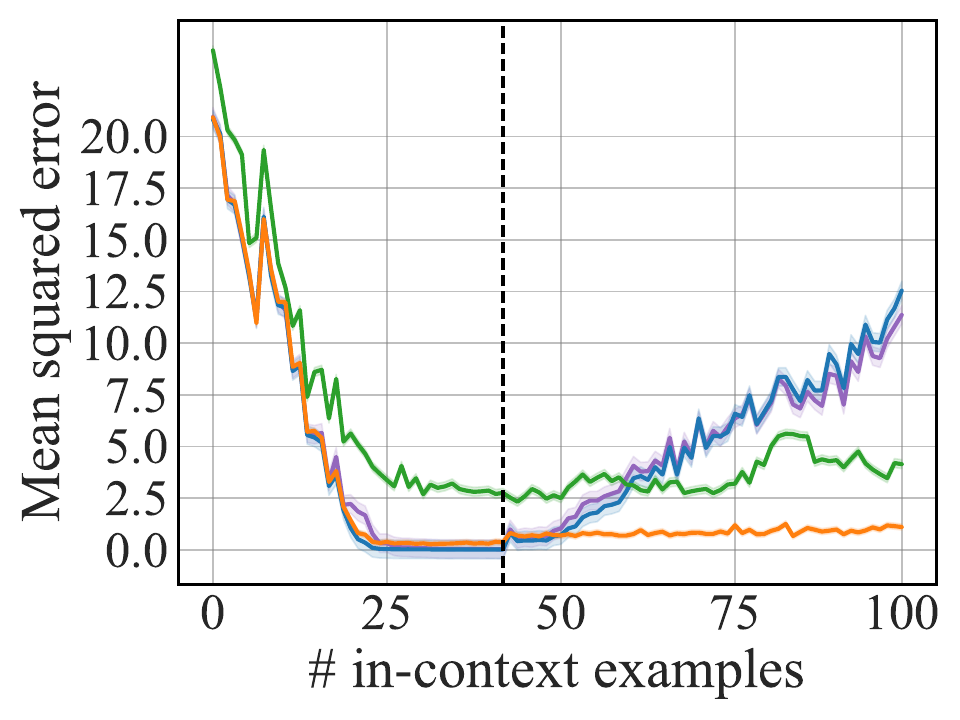}%
        }
        \subfigure[\phantom{(a)} \(T_\mathbf{v} = 101\)]{
                \includegraphics[width=0.25\linewidth]{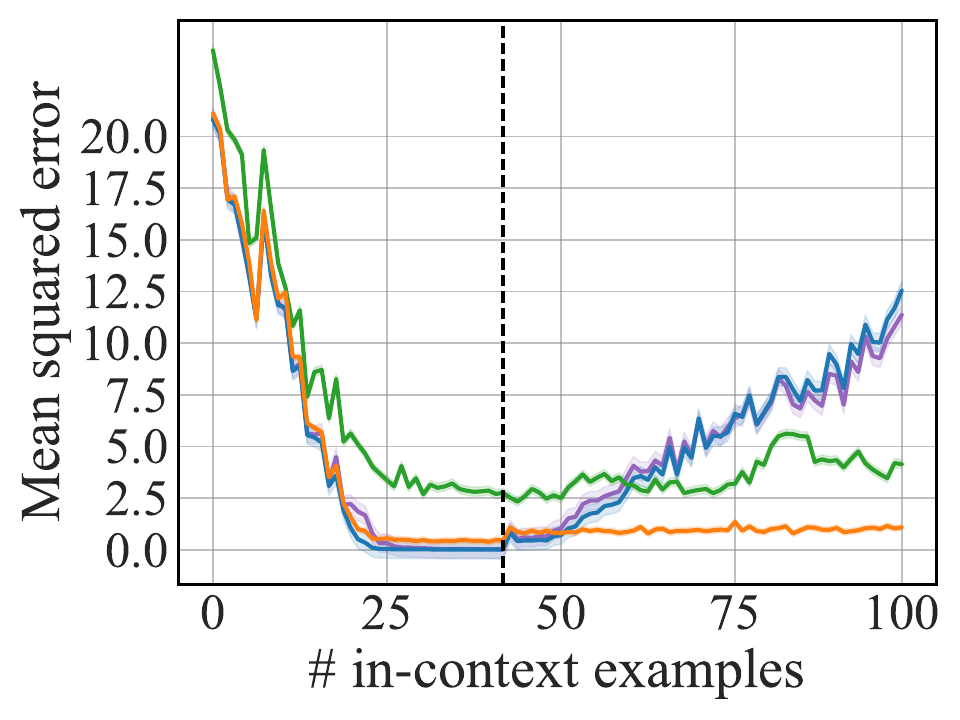}%
        }
    \caption{Evaluation on the class of linear functions on prompts with noisy labels, with the transformer pre-trained on up to \(T_\text{train} = 41\) examples per prompt. Results are averaged over a batch of 256 randomly selected tasks. The shaded area represents the 95\% confidence interval over the sampled prompts. \(T_\mathbf{v}\) denotes the prompt length used in LTV training.}
    \label{fig:dist_shift_noisy}
\end{figure*}

\paragraph{Optimizing the Learnable Task Vector}
In CoT-I/O, we train the LTV for each intermediate step. The objective functions for training LTVs for \( g(x) \) and \( f(x) \) are, respectively,  
\begin{multline*}
\text{\small $
    \min_{\Phi} \mathbb{E}_{f \sim \mathcal{D}_\mathcal{F}, x \sim \mathcal{D}_\mathcal{X}} \bigg[ \bigg( \tilde{M}_\theta \big(p_\text{CoT}(T_v+1) \mid v_\Phi^g \big)$} \\ \text{\small $ - g(x_\text{query}) \bigg)^2 \bigg],$ \normalsize} \\
\text{\small $
    \min_{\Phi} \mathbb{E}_{f \sim \mathcal{D}_\mathcal{F}, x \sim \mathcal{D}_\mathcal{X}} \bigg[ \bigg( \tilde{M}_\theta \big(\{p_\text{CoT}(T_v+1), g(x_\mathrm{query})\}$ \normalsize} \\ \text{\small $\mid v_\Phi^f \big) - f(x_\text{query}) \bigg)^2 \bigg].$ \normalsize}
\end{multline*}
Here, \( p_\text{CoT}(T_v+1) \) consists of \(T_v\) support examples and a new query:
\begin{equation*}
\{x_1, g(x_1), f(x_1), \dots, x_{T_v}, g(x_{T_v}), f(x_{T_v}), x_\text{query}\}.
\end{equation*}
The LTVs, \( v_\Phi^g = \{v^g_i\}_{i=1}^L \) and \( v_\Phi^f = \{v^f_i\}_{i=1}^L \), are applied at corresponding layers and steps, modifying transformer outputs as \( \tilde{M}_\theta(\cdot \mid v_\Phi^g) \) and \( \tilde{M}_\theta(\cdot \mid v_\Phi^f) \).  

In CoT-I, a single LTV is trained directly for the final step: 
\begin{multline*}
\text{\small $
    \min_{\Phi} \mathbb{E}_{f \sim \mathcal{D}_\mathcal{F}, x \sim \mathcal{D}_\mathcal{X}} \bigg[ \bigg( \tilde{M}_\theta \big(p_\text{CoT}(T_v+1) \mid v_\Phi^f \big)$ \normalsize} \\ \text{\small $- f(x_\text{query}) \bigg)^2 \bigg].$ \normalsize}
\end{multline*}
This formulation allows CoT-I to bypass intermediate step modeling, focusing solely on final output prediction.  

We construct the pre-training and LTV datasets following the same setting as in Appendix \ref{app:ltv_training}, sampling \(N\) i.i.d. tasks from an isotropic Gaussian with \(T_\text{train} = 101\) and \(N = 100 \times 256 = 25,600\).

\subsubsection{Results}
Results are provided in Figure \ref{fig:cot_results}. For simplicity, we do not include the baselines (\emph{i.e.}, FV and ICV). First, we observe that CoT-I/O (both with and without LTV) performs poorly. This is expected because the final prediction depends on the intermediate step prediction, \emph{i.e.}, $\hat{s}^{(1)}_i$ in \eqref{eq:cot_io_final_pred}. If the intermediate step introduces noise, it propagates into the final prediction, leading to degraded performance. Consequently, a more direct approach, \emph{i.e.}, CoT-I, achieves better and more effective prediction performance.

The latter case is particularly notable, as the model's in-context solving abilities improve up to \(T = T_\text{train}\), showing stable and lower error levels. However, CoT-I still struggles to maintain task fidelity when \(T > T_\text{train}\) on our benchmark.

For LTV, we observe that whether CoT is used or not, LTV alone can sustain task behavior. This indicates that CoT, when combined with LTV, performs well primarily due to LTV, with minimal contribution to preserving task fidelity. This suggests that our method does not require additional ICL enhancements to maintain task fidelity.

\subsection{Distributional Shift}
\label{app:dist_shift_results}
We identify two scenarios from \citet{garg_icl} where the transformer model's performance notably degrades during ICL inference: noisy linear regression and skewed covariance matrix.

\paragraph{Noisy linear regression} Noise is added to the output of each example in the form of a standard Gaussian distribution. Specifically, the \(i\)-th output is defined as \(w^\top x_i + \epsilon_i\), where \(\epsilon_i \sim \mathcal{N}(0, 1)\). While the transformer and LTV are trained on standard linear regression, the data during the ICL inference phase is modified by this additive noise. In Figure \ref{fig:dist_shift_noisy}, we observe that while the performance of FV degrades significantly, LTV is only slightly affected. Specifically, LTV requires training on more examples to maintain the same performance as in the noise-free setting. For instance, the performance of LTV at \(T_\mathbf{v} = 71\) is comparable to its performance in a noise-free environment at around \(T_\mathbf{v} = 56\).

\begin{figure*}[!tbh]
    \centering
    \begin{align*}
        &\text{\small {\blue} Transformer} \qquad &&\text{\small {\green} Transformer + FV (optimized)} \\ &\text{\small {\purple} Transformer + ICV (tuned)} \qquad &&\text{\small {\orange} Transformer + LTV}
    \end{align*}
	\subfigure[\phantom{(a)} \(T_\mathbf{v} = 41\)]{
                \includegraphics[width=0.25\linewidth]{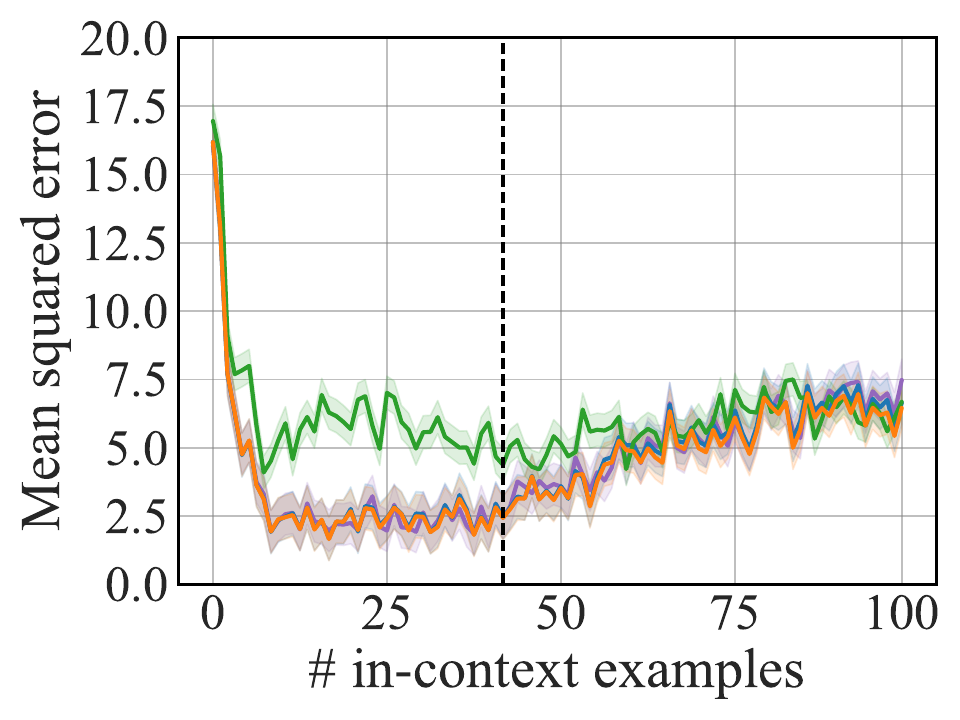}%
        }
        \subfigure[\phantom{(a)} \(T_\mathbf{v} = 42\)]{
                \includegraphics[width=0.25\linewidth]{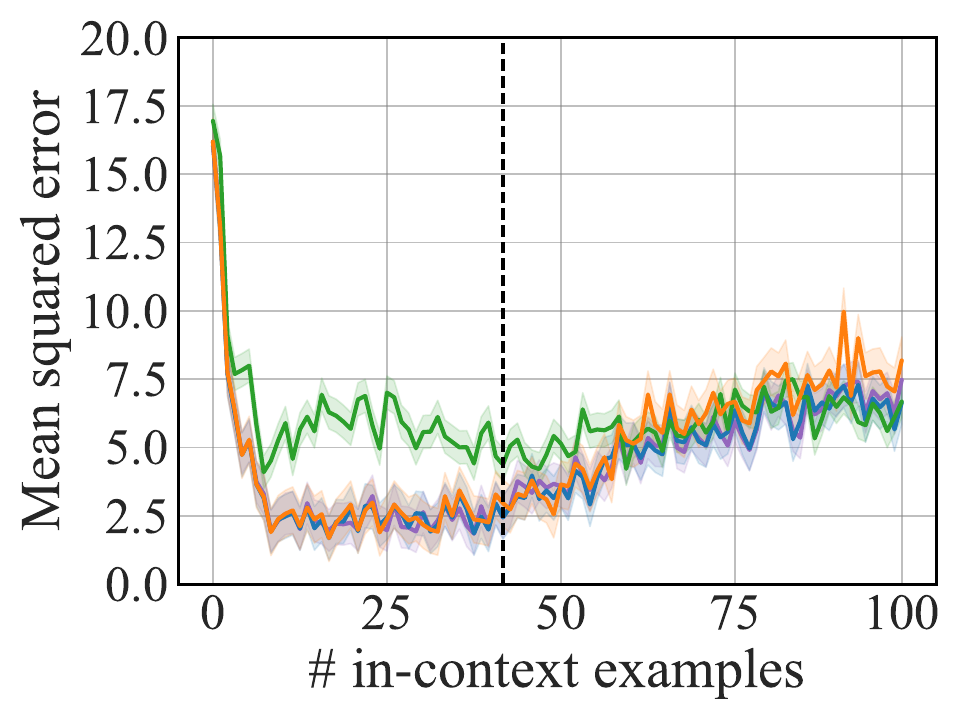}%
        }
        \subfigure[\phantom{(a)} \(T_\mathbf{v} = 56\)]{
                \includegraphics[width=0.25\linewidth]{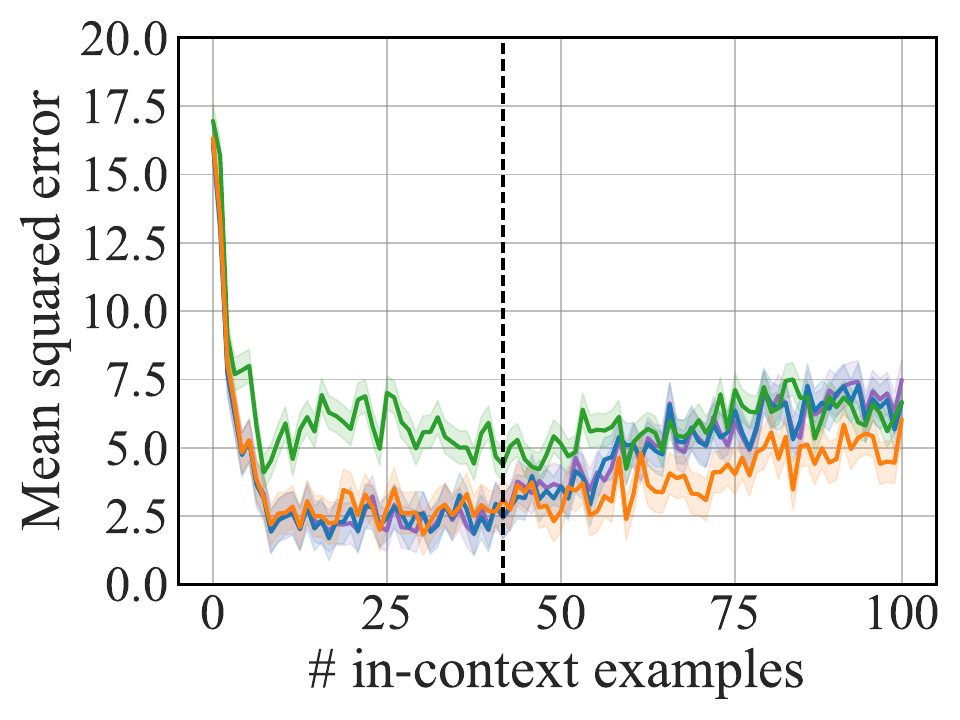}%
        }%
        \hfill
        \subfigure[\phantom{(a)} \(T_\mathbf{v} = 71\)]{
                \includegraphics[width=0.25\linewidth]{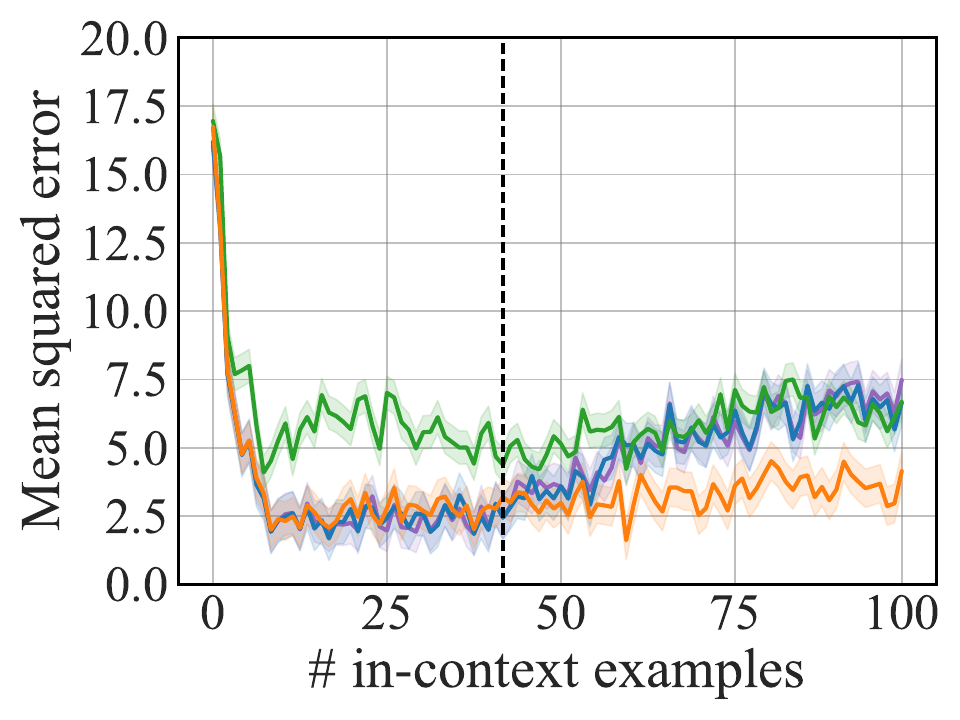}%
        }
        \subfigure[\phantom{(a)} \(T_\mathbf{v} = 86\)]{
                \includegraphics[width=0.25\linewidth]{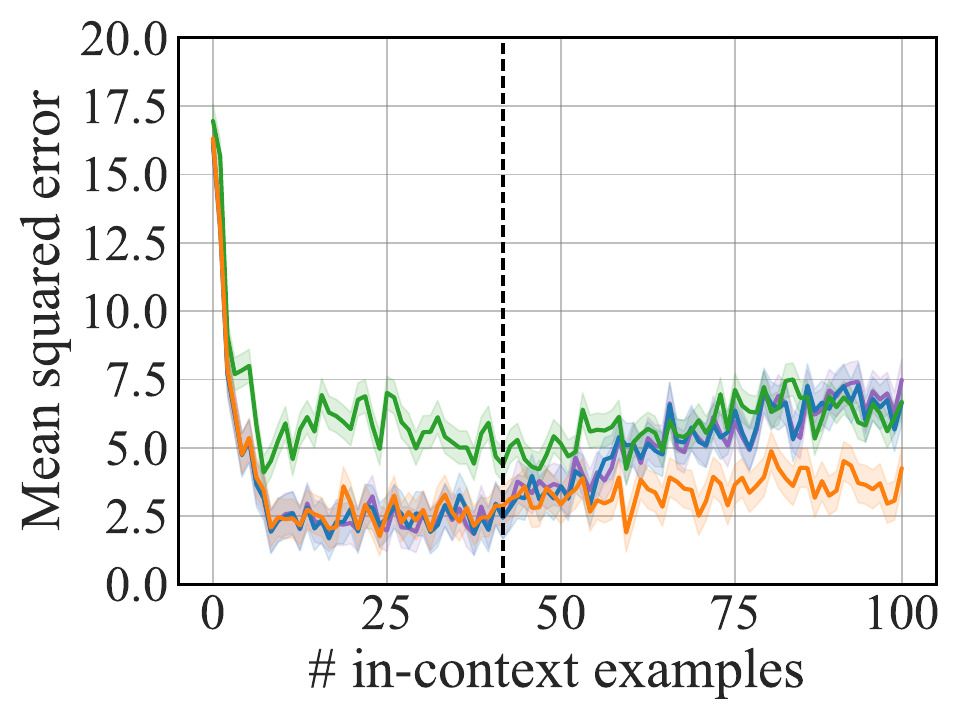}%
        }
        \subfigure[\phantom{(a)} \(T_\mathbf{v} = 101\)]{
                \includegraphics[width=0.25\linewidth]{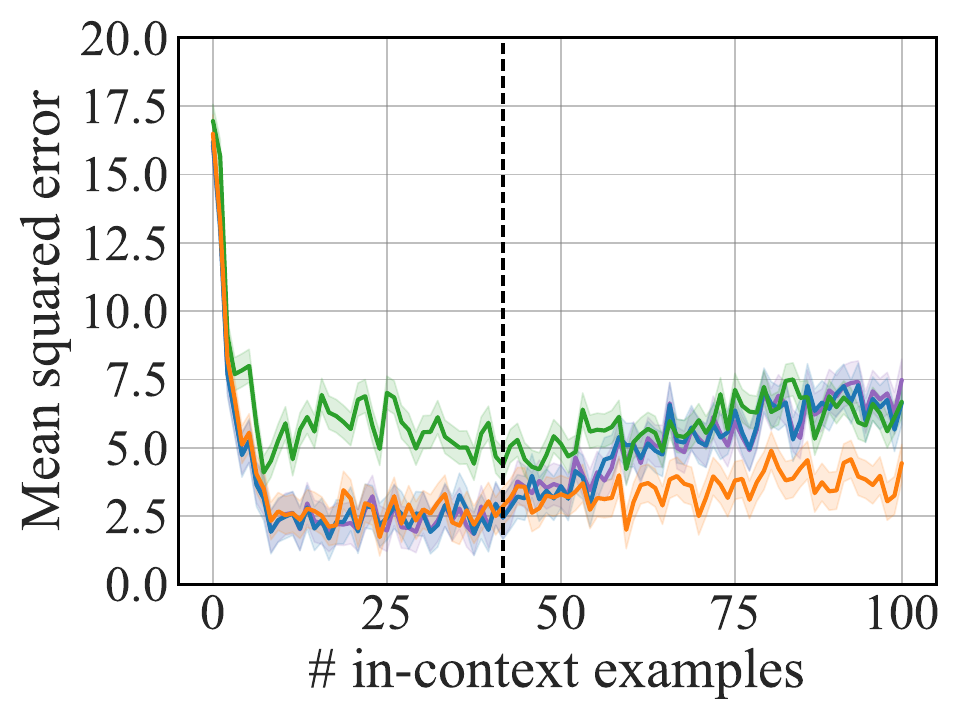}%
        }
    \caption{Evaluation on the class of linear functions under skewed covariance, with the transformer pre-trained on up to \(T_\text{train} = 41\) examples per prompt. Results are averaged over a batch of 256 randomly selected tasks. The shaded area represents the 95\% confidence interval over the sampled prompts. \(T_\mathbf{v}\) denotes the prompt length used in LTV training.}
    \label{fig:dist_shift_skewed_lin_reg}
\end{figure*}

\paragraph{Skewed covariance} The inputs for the prompts are sampled from a zero-mean skewed Gaussian distribution: \(x \sim \mathcal{N}(0, \Tilde{\Sigma})\), where the eigenbasis of the skewed covariance matrix \(\Tilde{\Sigma}\) is chosen uniformly at random. Each \(i\)-th eigenvalue of this matrix is proportional to \(1/i^2\). The results, shown in Figures \ref{fig:dist_shift_skewed_lin_reg}, \ref{fig:dist_shift_skewed_sparse_lin_reg}, and \ref{fig:dist_shift_relu_nets}, align with the findings of \citet{garg_icl}, with error curves being more unstable and oscillatory compared to the isotropic Gaussian case. While LTV exhibits some sensitivity to this instability, it maintains a low mean error and preserves performance to a certain extent.

\begin{figure*}[!tbh]
    \centering
    \begin{align*}
        &\text{\small {\blue} Transformer} \qquad &&\text{\small {\green} Transformer + FV (optimized)} \\ &\text{\small {\purple} Transformer + ICV (tuned)} \qquad &&\text{\small {\orange} Transformer + LTV}
    \end{align*}
	\subfigure[\phantom{(a)} \(T_\mathbf{v} = 41\)]{
                \includegraphics[width=0.25\linewidth]{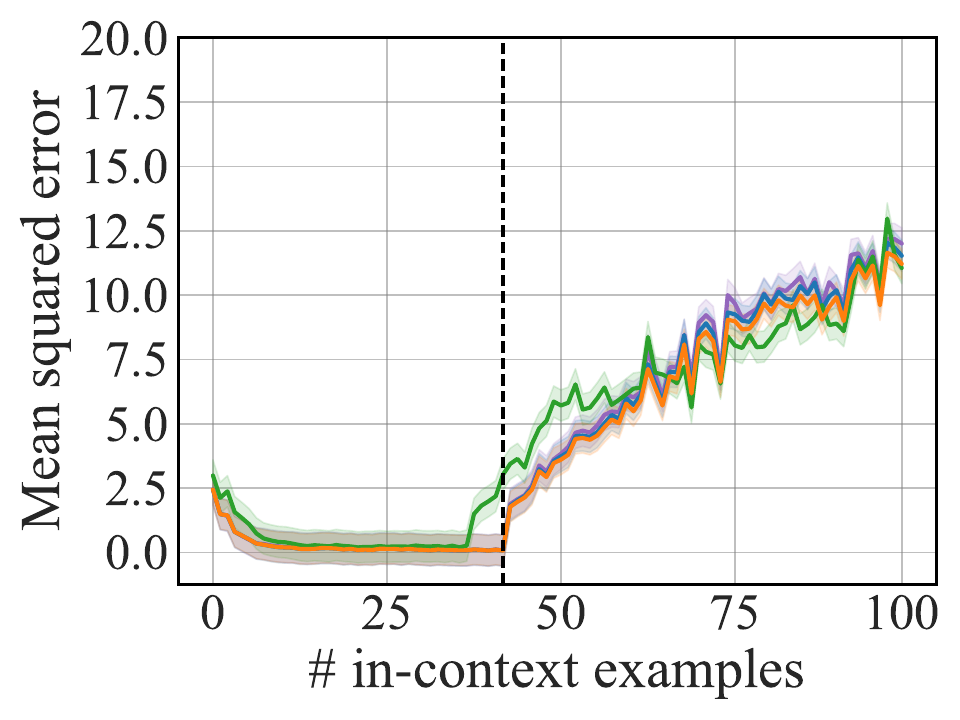}%
        }
        \subfigure[\phantom{(a)} \(T_\mathbf{v} = 42\)]{
                \includegraphics[width=0.25\linewidth]{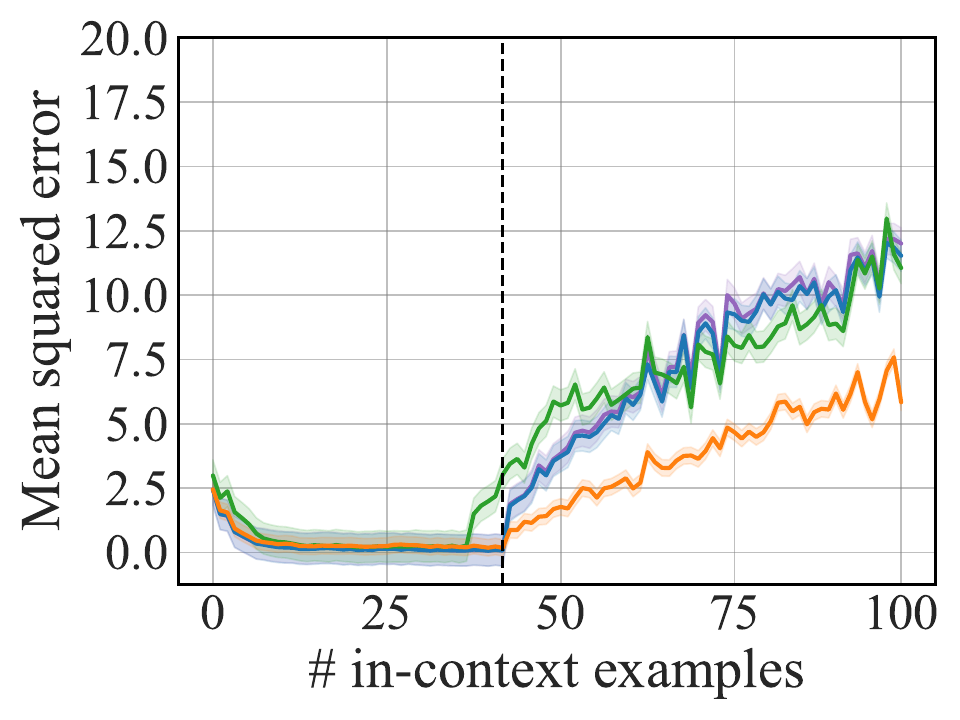}%
        }
        \subfigure[\phantom{(a)} \(T_\mathbf{v} = 56\)]{
                \includegraphics[width=0.25\linewidth]{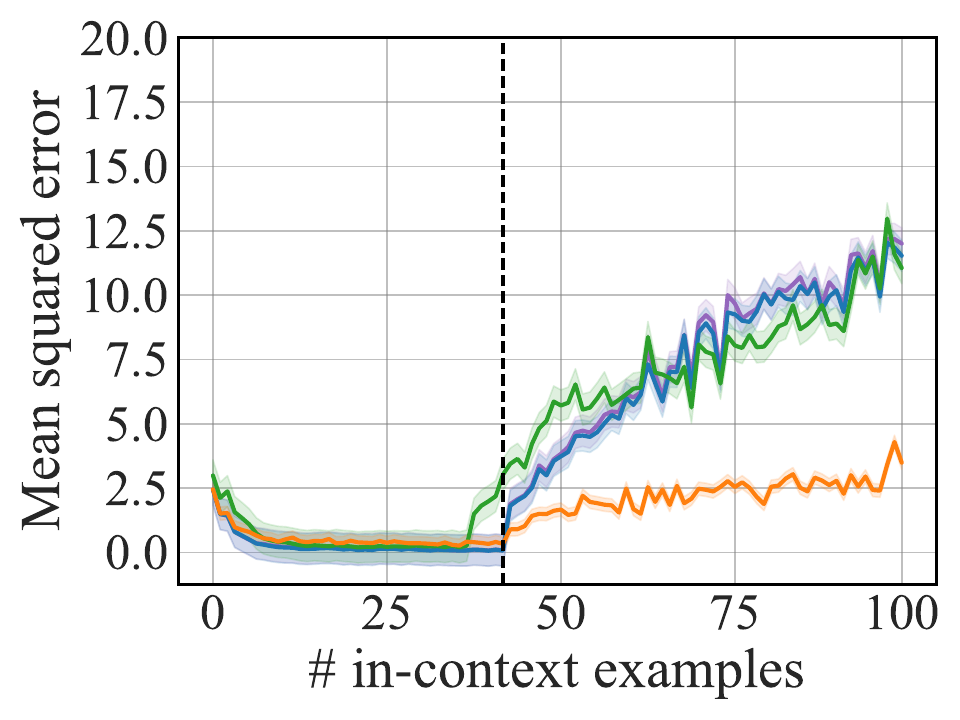}%
        }%
        \hfill
        \subfigure[\phantom{(a)} \(T_\mathbf{v} = 71\)]{
                \includegraphics[width=0.25\linewidth]{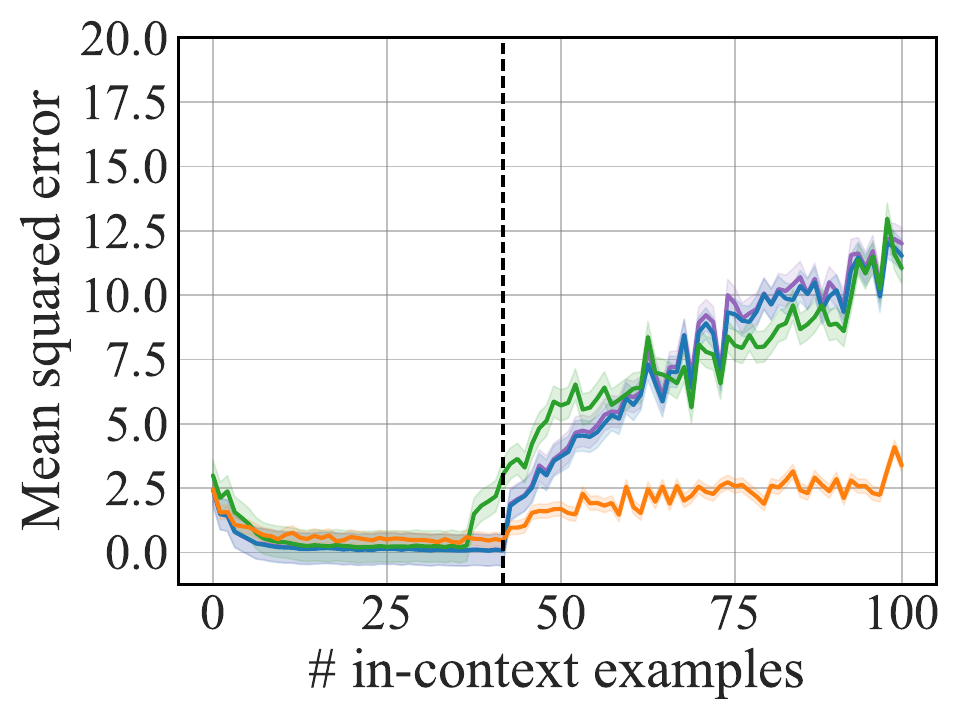}%
        }
        \subfigure[\phantom{(a)} \(T_\mathbf{v} = 86\)]{
                \includegraphics[width=0.25\linewidth]{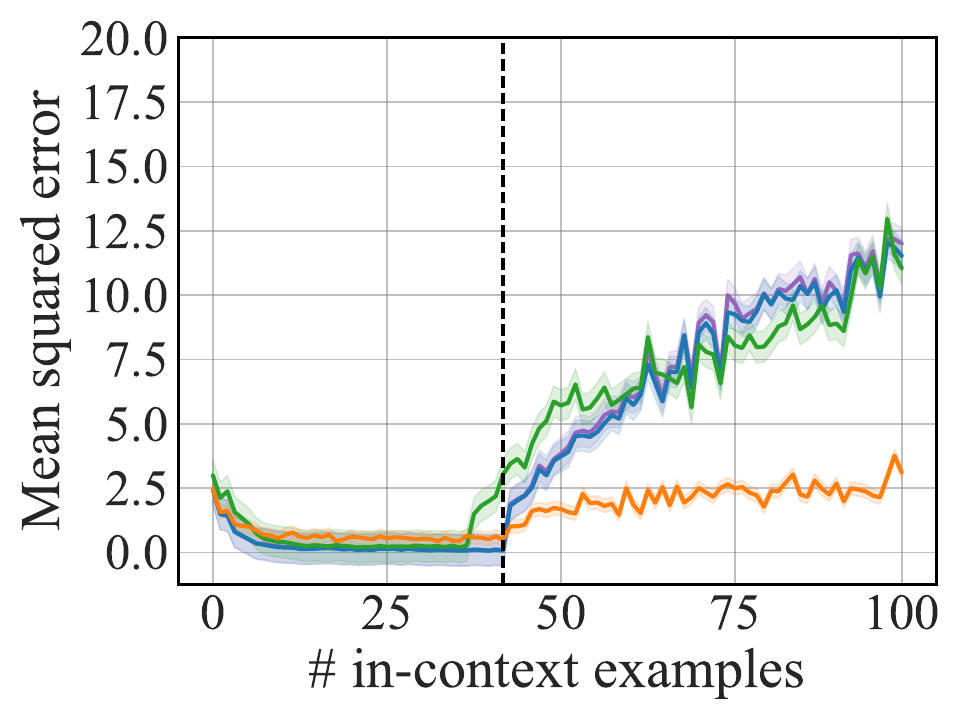}%
        }
        \subfigure[\phantom{(a)} \(T_\mathbf{v} = 101\)]{
                \includegraphics[width=0.25\linewidth]{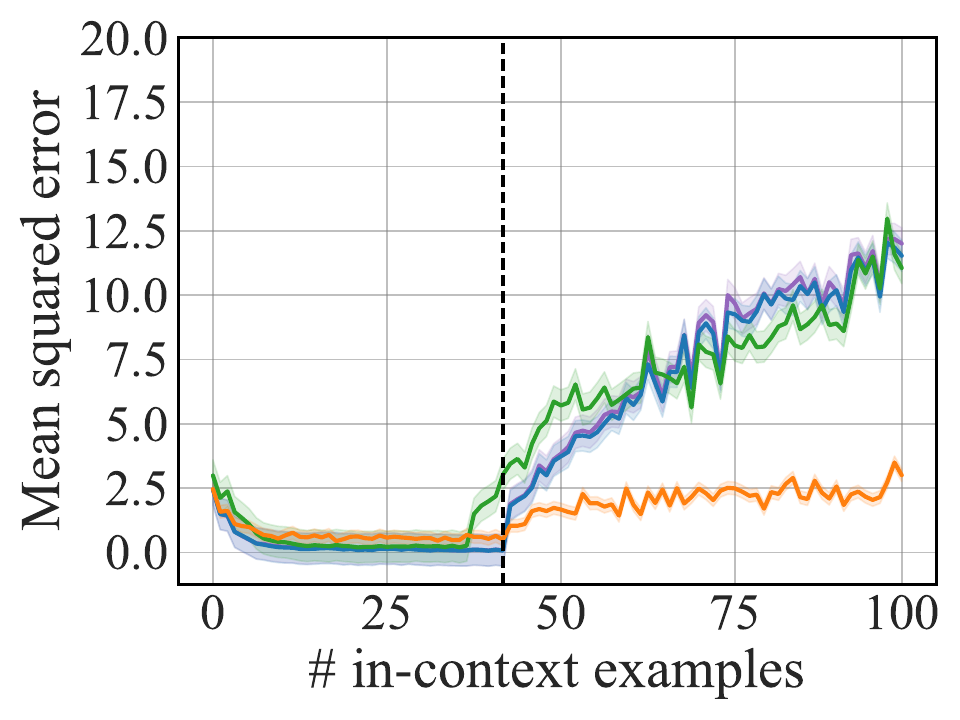}%
        }
    \caption{Evaluation on the class of sparse linear functions under skewed covariance, with the transformer pre-trained on up to \(T_\text{train} = 41\) examples per prompt. Results are averaged over a batch of 256 randomly selected tasks. The shaded area represents the 95\% confidence interval over the sampled prompts. \(T_\mathbf{v}\) denotes the prompt length used in LTV training.}
    \label{fig:dist_shift_skewed_sparse_lin_reg}
\end{figure*}

\begin{figure*}[!tbh]
    \centering
    \begin{align*}
        &\text{\small {\blue} Transformer} \qquad &&\text{\small {\green} Transformer + FV (optimized)} \\ &\text{\small {\purple} Transformer + ICV (tuned)} \qquad &&\text{\small {\orange} Transformer + LTV}
    \end{align*}
	\subfigure[\phantom{(a)} \(T_\mathbf{v} = 101\)]{
                \includegraphics[width=0.25\linewidth]{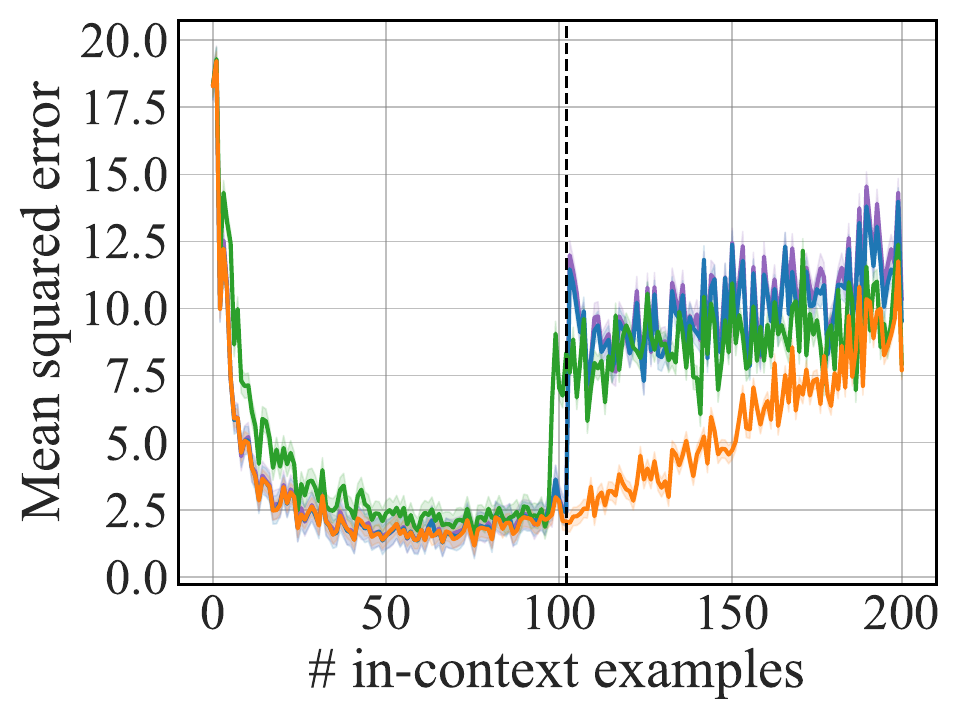}%
        }
        \subfigure[\phantom{(a)} \(T_\mathbf{v} = 102\)]{
                \includegraphics[width=0.25\linewidth]{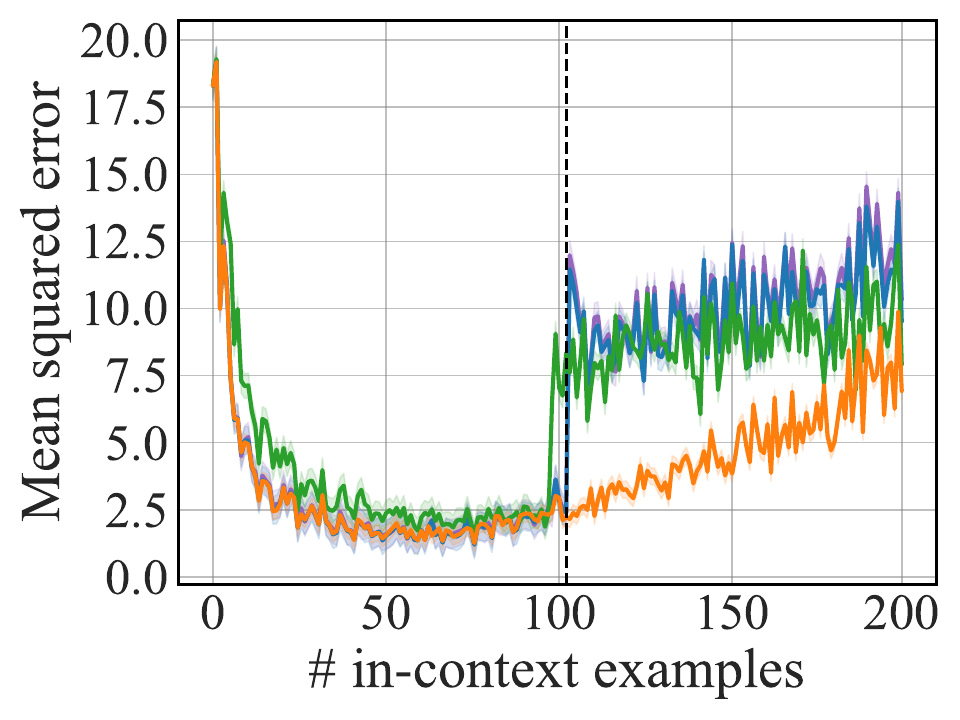}%
        }
        \subfigure[\phantom{(a)} \(T_\mathbf{v} = 126\)]{
                \includegraphics[width=0.25\linewidth]{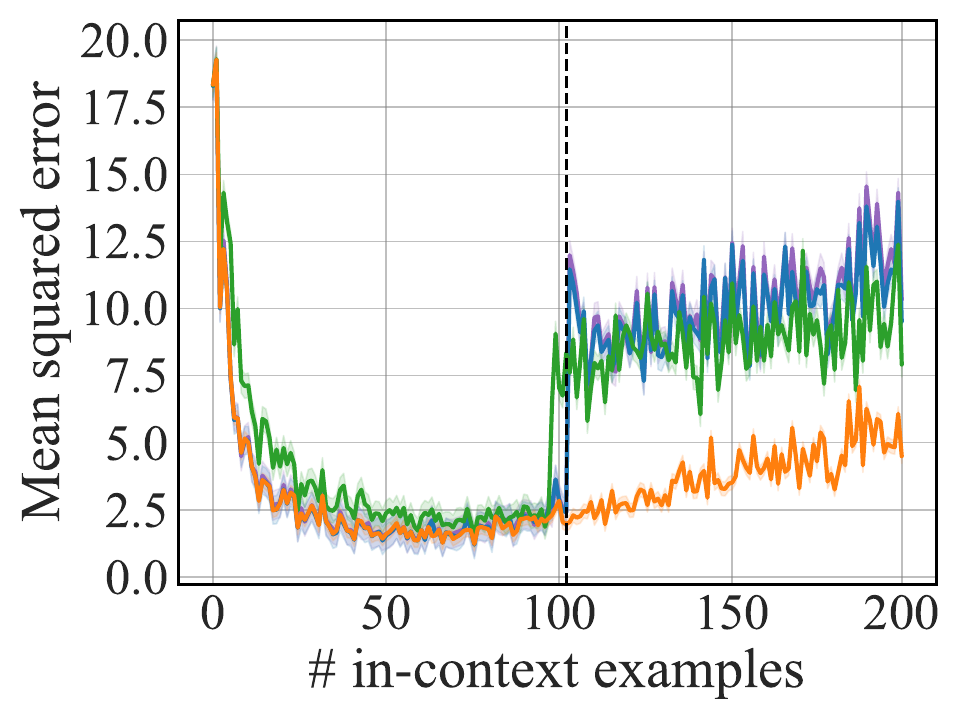}%
        }%
        \hfill
        \subfigure[\phantom{(a)} \(T_\mathbf{v} = 151\)]{
                \includegraphics[width=0.25\linewidth]{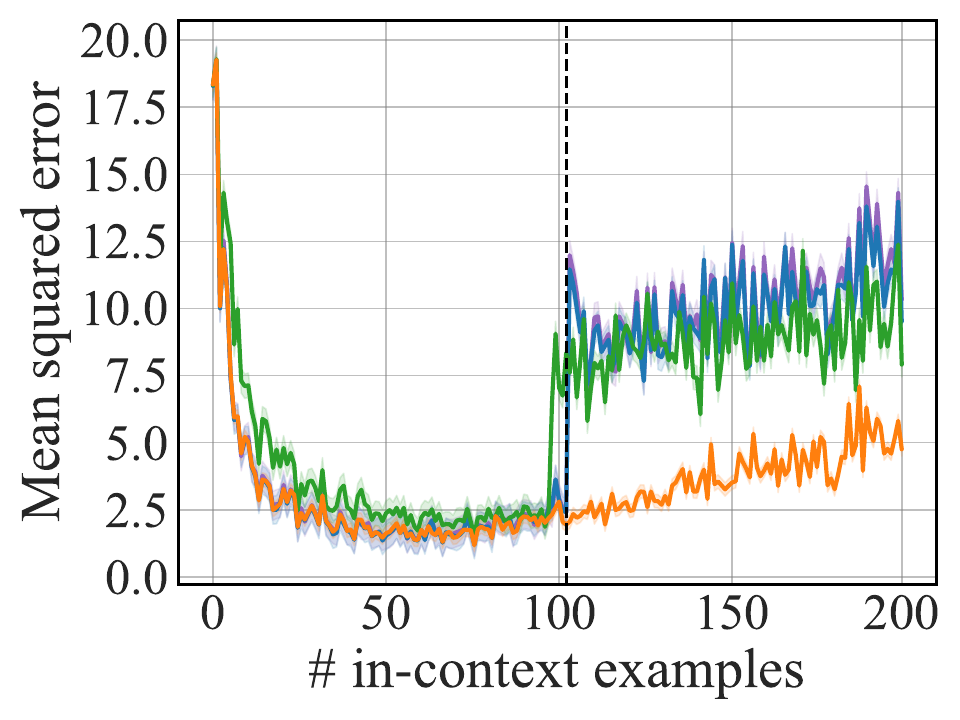}%
        }
        \subfigure[\phantom{(a)} \(T_\mathbf{v} = 176\)]{
                \includegraphics[width=0.25\linewidth]{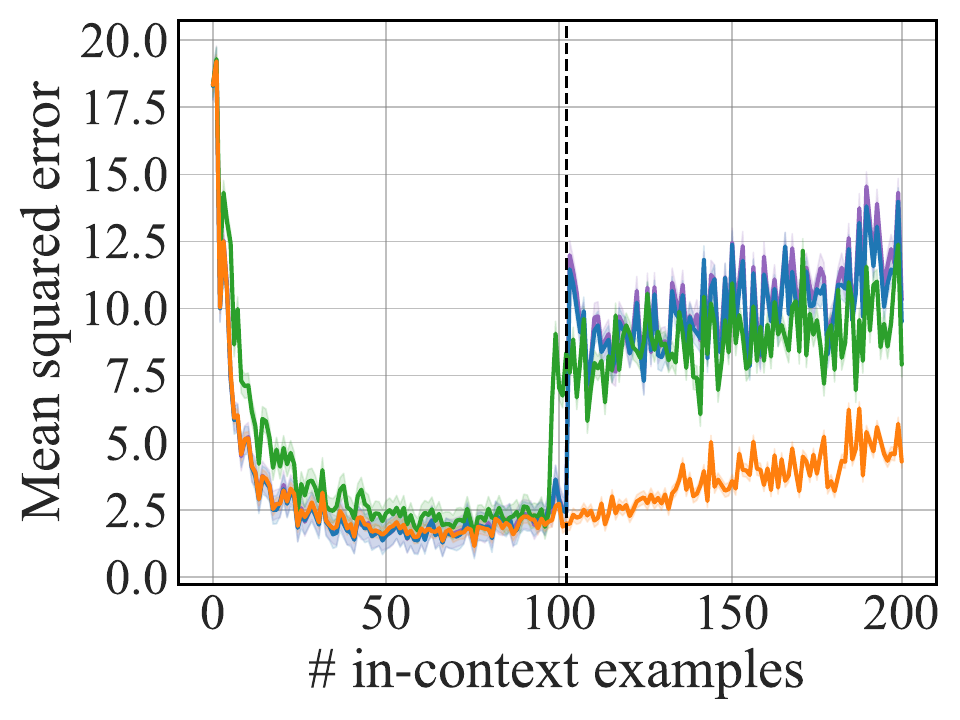}%
        }
        \subfigure[\phantom{(a)} \(T_\mathbf{v} = 201\)]{
                \includegraphics[width=0.25\linewidth]{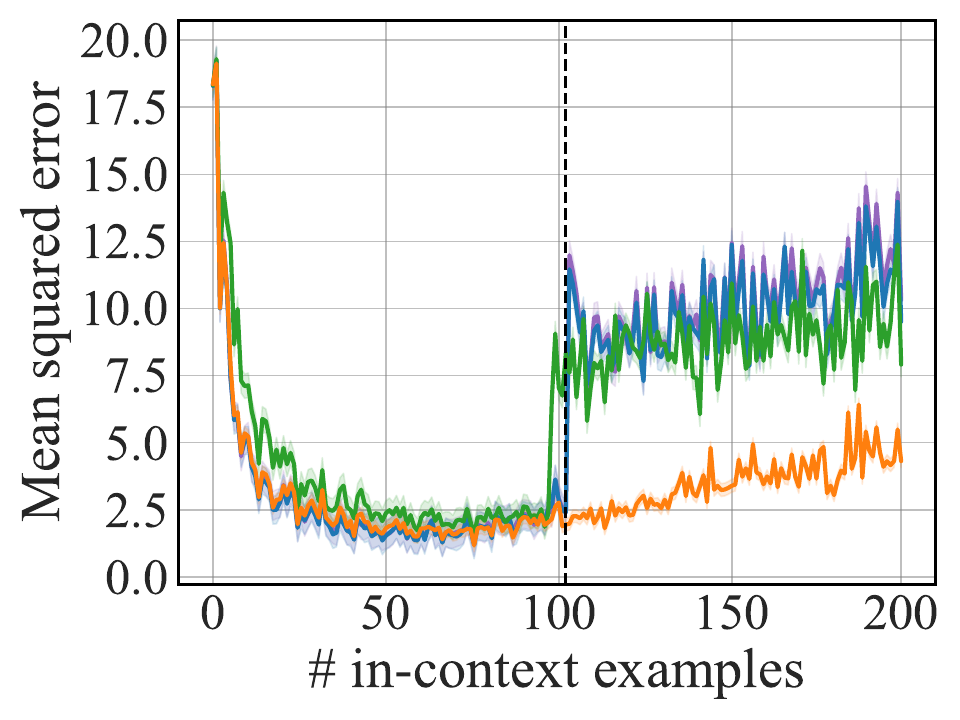}%
        }
    \caption{Evaluation on the class of 2-layer ReLU neural networks under skewed covariance, with the transformer pre-trained on up to \(T_\text{train} = 101\) examples per prompt. Results are averaged over a batch of 256 randomly selected tasks. The shaded area represents the 95\% confidence interval over the sampled prompts. \(T_\mathbf{v}\) denotes the prompt length used in LTV training.}
    \label{fig:dist_shift_relu_nets}
\end{figure*}

\begin{table*}[!htbp]
\centering
\resizebox{\textwidth}{!}{%
\begin{tabular}{@{}c lccccccc@{}}
\toprule
\multicolumn{1}{l}{} & Dataset       & Prompt        & Transformer & + FV   & + ICV  & + SV   & + TV   & + LTV (ours) \\ 
\midrule
\multirow{24}{*}{\begin{sideways}Abstractive\end{sideways}} 

& \multirow{2}{*}{AG News \citep{{zhang}}} & Zero-shot & 0.0 $\pm$ 0.0 & 30.1 $\pm$ 5.7 & 33.6 $\pm$ 5.6 & 31.7 $\pm$ 5.7 & 34.5 $\pm$ 5.0 & \highlight[customorange]{\textbf{46.9 $\pm$ 6.2}} \\
 & & Few-shot & \highlight[customorange]{67.2 $\pm$ 5.8} & \highlight[customorange]{66.0 $\pm$ 5.8} & 64.7 $\pm$ 5.4 & 66.1 $\pm$ 4.5 & 57.9 $\pm$ 5.8 & \highlight[customorange]{\textbf{71.1 $\pm$ 5.6}} \\
 & \multirow{2}{*}{Antonym \citep{{antonym_synonym_dataset}}} & Zero-shot & 2.3 $\pm$ 1.9 & 35.2 $\pm$ 5.9 & 46.4 $\pm$ 5.6 & 35.2 $\pm$ 6.1 & 29.6 $\pm$ 6.0 & \highlight[customorange]{\textbf{53.5 $\pm$ 6.2}} \\
 & & Few-shot & 37.5 $\pm$ 6.0 & \highlight[customorange]{64.5 $\pm$ 5.9} & \highlight[customorange]{61.8 $\pm$ 5.7} & \highlight[customorange]{64.8 $\pm$ 5.9} & \highlight[customorange]{64.9 $\pm$ 5.2} & \highlight[customorange]{\textbf{65.6 $\pm$ 5.9}} \\
 & \multirow{2}{*}{CommonsenseQA \citep{{talmor}}} & Zero-shot & 10.2 $\pm$ 3.7 & 23.0 $\pm$ 5.2 & 28.9 $\pm$ 5.2 & 21.7 $\pm$ 1.1 & 22.0 $\pm$ 4.0 & \highlight[customorange]{\textbf{41.0 $\pm$ 6.1}} \\
 & & Few-shot & \highlight[customorange]{18.0 $\pm$ 4.7} & \highlight[customorange]{20.3 $\pm$ 5.0} & \highlight[customorange]{20.0 $\pm$ 4.6} & \highlight[customorange]{20.3 $\pm$ 2.1} & \highlight[customorange]{19.5 $\pm$ 3.4} & \highlight[customorange]{\textbf{21.1 $\pm$ 5.0}} \\
 & \multirow{2}{*}{English-French \citep{{connenau}}} & Zero-shot & 0.0 $\pm$ 0.0 & 23.4 $\pm$ 5.2 & 40.2 $\pm$ 5.3 & 24.4 $\pm$ 2.7 & 25.4 $\pm$ 5.1 & \highlight[customorange]{\textbf{53.5 $\pm$ 6.2}} \\
 & & Few-shot & 52.0 $\pm$ 6.2 & \highlight[customorange]{71.9 $\pm$ 5.5} & 68.0 $\pm$ 5.0 & \highlight[customorange]{74.4 $\pm$ 4.2} & \highlight[customorange]{75.0 $\pm$ 5.3} & \highlight[customorange]{\textbf{75.4 $\pm$ 5.3}} \\
 & \multirow{2}{*}{English-Spanish \citep{{connenau}}} & Zero-shot & 0.0 $\pm$ 0.0 & 9.0 $\pm$ 3.5 & \highlight[customorange]{16.5 $\pm$ 4.3} & 9.7 $\pm$ 3.5 & 9.4 $\pm$ 2.9 & \highlight[customorange]{\textbf{20.7 $\pm$ 5.0}} \\
 & & Few-shot & 36.7 $\pm$ 5.9 & \highlight[customorange]{50.4 $\pm$ 6.2} & \highlight[customorange]{49.4 $\pm$ 5.9} & \highlight[customorange]{50.8 $\pm$ 4.2} & \highlight[customorange]{50.6 $\pm$ 4.9} & \highlight[customorange]{\textbf{53.9 $\pm$ 6.1}} \\
 & \multirow{2}{*}{Landmark-Country \citep{{hernandez}}} & Zero-shot & 0.0 $\pm$ 0.0 & 60.9 $\pm$ 6.0 & 62.7 $\pm$ 5.1 & 62.9 $\pm$ 5.6 & \highlight[customorange]{69.9 $\pm$ 6.0} & \highlight[customorange]{\textbf{71.1 $\pm$ 5.6}} \\
 & & Few-shot & 66.8 $\pm$ 5.8 & 68.0 $\pm$ 5.8 & 68.4 $\pm$ 5.0 & 65.6 $\pm$ 5.3 & 66.3 $\pm$ 5.6 & \highlight[customorange]{\textbf{76.2 $\pm$ 5.3}} \\
 & \multirow{2}{*}{Person-Instrument \citep{{hernandez}}} & Zero-shot & 0.0 $\pm$ 0.0 & 11.3 $\pm$ 3.9 & 17.1 $\pm$ 4.9 & 13.1 $\pm$ 2.2 & \highlight[customorange]{30.1 $\pm$ 3.5} & \highlight[customorange]{\textbf{33.2 $\pm$ 5.8}} \\
 & & Few-shot & 52.7 $\pm$ 6.2 & 48.4 $\pm$ 6.2 & 53.2 $\pm$ 5.9 & 47.9 $\pm$ 6.1 & 43.6 $\pm$ 6.2 & \highlight[customorange]{\textbf{60.5 $\pm$ 6.0}} \\
 & \multirow{2}{*}{Person-Occupation \citep{{hernandez}}} & Zero-shot & 0.0 $\pm$ 0.0 & 4.3 $\pm$ 2.5 & 17.2 $\pm$ 3.9 & 9.0 $\pm$ 2.9 & 8.1 $\pm$ 5.8 & \highlight[customorange]{\textbf{36.3 $\pm$ 5.9}} \\
 & & Few-shot & 30.9 $\pm$ 5.7 & 30.1 $\pm$ 5.7 & 37.6 $\pm$ 5.6 & 29.2 $\pm$ 5.2 & 28.8 $\pm$ 5.7 & \highlight[customorange]{\textbf{49.6 $\pm$ 6.2}} \\
 & \multirow{2}{*}{Person-Sport \citep{{hernandez}}} & Zero-shot & 0.0 $\pm$ 0.0 & 2.7 $\pm$ 2.0 & \highlight[customorange]{12.9 $\pm$ 3.5} & 12.5 $\pm$ 3.1 & 12.6 $\pm$ 3.0 & \highlight[customorange]{\textbf{16.0 $\pm$ 4.5}} \\
 & & Few-shot & \highlight[customorange]{85.5 $\pm$ 4.3} & \highlight[customorange]{86.3 $\pm$ 4.2} & 83.3 $\pm$ 3.5 & \highlight[customorange]{88.3 $\pm$ 4.1} & \highlight[customorange]{86.5 $\pm$ 4.0} & \highlight[customorange]{\textbf{89.5 $\pm$ 3.8}} \\
 & \multirow{2}{*}{Product-Company \citep{{hernandez}}} & Zero-shot & 0.0 $\pm$ 0.0 & 29.3 $\pm$ 5.6 & 43.0 $\pm$ 5.3 & 28.2 $\pm$ 4.9 & 28.7 $\pm$ 5.6 & \highlight[customorange]{\textbf{66.4 $\pm$ 5.8}} \\
 & & Few-shot & 57.4 $\pm$ 6.1 & 62.5 $\pm$ 6.0 & 66.6 $\pm$ 5.8 & 68.0 $\pm$ 5.6 & 65.0 $\pm$ 4.7 & \highlight[customorange]{\textbf{74.2 $\pm$ 5.4}} \\
 & \multirow{2}{*}{Sentiment Analysis \citep{{socher}}} & Zero-shot & 0.0 $\pm$ 0.0 & 0.0 $\pm$ 0.0 & 4.8 $\pm$ 1.9 & 0.0 $\pm$ 0.0 & 0.0 $\pm$ 0.0 & \highlight[customorange]{\textbf{10.9 $\pm$ 3.8}} \\
 & & Few-shot & 74.6 $\pm$ 5.4 & 69.5 $\pm$ 5.7 & 72.5 $\pm$ 3.2 & 74.2 $\pm$ 2.5 & 77.1 $\pm$ 3.4 & \highlight[customorange]{\textbf{94.5 $\pm$ 2.8}} \\
 & \multirow{2}{*}{Synonym \citep{{antonym_synonym_dataset}}} & Zero-shot & 1.2 $\pm$ 1.3 & 2.7 $\pm$ 2.0 & 14.2 $\pm$ 3.6 & 2.6 $\pm$ 4.6 & 2.5 $\pm$ 3.3 & \highlight[customorange]{\textbf{18.8 $\pm$ 4.8}} \\
 & & Few-shot & 6.2 $\pm$ 3.0 & 10.5 $\pm$ 3.8 & 14.6 $\pm$ 4.5 & 10.5 $\pm$ 4.2 & 10.1 $\pm$ 3.0 & \highlight[customorange]{\textbf{40.6 $\pm$ 6.1}} \\
 \midrule

\multirow{6}{*}{\begin{sideways}Extractive\end{sideways}} 
& \multirow{2}{*}{NER-person \citep{{ner}}} & Zero-shot & 5.5 $\pm$ 2.8 & 48.8 $\pm$ 6.2 & 49.9 $\pm$ 5.8 & 46.3 $\pm$ 1.3 & 47.8 $\pm$ 5.3 & \highlight[customorange]{\textbf{58.2 $\pm$ 6.1}} \\
 & & Few-shot & 11.7 $\pm$ 4.0 & 56.2 $\pm$ 6.1 & 65.9 $\pm$ 5.2 & 53.7 $\pm$ 3.3 & 62.6 $\pm$ 3.9 & \highlight[customorange]{\textbf{79.3 $\pm$ 5.0}} \\
 & \multirow{2}{*}{NER-location \citep{{ner}}} & Zero-shot & 6.6 $\pm$ 3.1 & \highlight[customorange]{\textbf{37.9 $\pm$ 6.0}} & 26.4 $\pm$ 5.1 & \highlight[customorange]{34.6 $\pm$ 3.9} & 27.2 $\pm$ 4.7 & 27.3 $\pm$ 5.5 \\
 & & Few-shot & 23.4 $\pm$ 5.2 & 43.4 $\pm$ 6.1 & 47.4 $\pm$ 5.9 & 43.5 $\pm$ 4.4 & 41.9 $\pm$ 5.6 & \highlight[customorange]{\textbf{57.4 $\pm$ 6.1}} \\
 & \multirow{2}{*}{NER-organization \citep{{ner}}} & Zero-shot & 21.9 $\pm$ 5.1 & 48.8 $\pm$ 6.2 & 52.5 $\pm$ 6.0 & 46.1 $\pm$ 1.2 & 46.6 $\pm$ 4.3 & \highlight[customorange]{\textbf{64.1 $\pm$ 5.9}} \\
 & & Few-shot & 16.8 $\pm$ 4.6 & 50.8 $\pm$ 6.2 & 54.1 $\pm$ 5.0 & 53.5 $\pm$ 2.6 & 51.1 $\pm$ 4.1 & \highlight[customorange]{\textbf{75.0 $\pm$ 5.3}} \\
\bottomrule
\end{tabular}}
\caption{Complete set of accuracy scores (\%) for zero-shot and few-shot (5-shot) predictions, averaged across 256 random seeds. \(\pm\) represents the margin of error at a 95\% confidence level. The highest accuracy is marked in \textbf{boldface}, and the statistically best-performing method is \highlight[customorange]{highlighted}.}
\label{tab:language_results_full}
\end{table*}

\begin{table*}[!bpht]
\centering
    \begin{tabular}{@{}lccc@{}}
        \toprule
        Configuration at \(T_\text{max}\) & Linear regression & Sparse linear regression & 2-layer ReLU NN \\ \midrule
        Transformer & 1.098 & 1.110 & 0.103 \\
        + Function Vector & 0.646 & 0.572 & 0.223 \\
        + LTV (\(T_\mathbf{v} = \{41, 41, 101\}\)) & 1.000 & 1.056 & 0.095 \\
        + LTV (\(T_\mathbf{v} = \{42, 42, 102\}\)) & 0.952 & 0.984 & 0.037 \\
        + LTV (\(T_\mathbf{v} = \{56, 56, 126\}\)) & 0.532 & 0.260 & 0.016 \\
        + LTV (\(T_\mathbf{v} = \{71, 71, 151\}\)) & 0.460 & 0.203 & 0.017 \\
        + LTV (\(T_\mathbf{v} = \{86, 86, 176\}\)) & 0.344 & 0.199 & \textbf{0.013} \\
        + LTV (\(T_\mathbf{v} = \{101, 101, 201\}\)) & \textbf{0.300} & \textbf{0.096} & 0.038 \\
        \bottomrule
    \end{tabular}
    \caption{KL divergence values are computed between the distributions of the last hidden states of the vanilla transformer at \(T = T_{\text{train}}\) and the listed configurations at \(T = T_{\text{max}}\), where \(T_{\text{max}} = 101\) for linear functions and \(T_{\text{max}} = 201\) for neural networks. Kernel density estimation (KDE) is used to estimate the probability densities over a dataset of 25,600 samples. The lowest KL divergence score (\emph{i.e.}, the most aligned configuration) is marked in \textbf{boldface}.}
    \label{tab:ablations}
\end{table*}

\subsection{Natural Language Processing}
\label{app:nlp_results}
The complete set of accuracy scores on NLP benchmarks is reported in Table \ref{tab:language_results_full}. Detailed descriptions of these benchmarks can be found in Appendix \ref{app:language_tasks}.

\subsection{Ablation Studies}
\label{app:ablation_results}
The KL divergence scores are reported in Table \ref{tab:ablations}. Additionally, the histograms in Figures \ref{fig:ablation_histograms_lin_reg}, \ref{fig:ablation_histograms_sparse_lin_reg}, and \ref{fig:ablation_histograms_relu_nets} illustrate the (unnormalized) probability density functions of the last hidden states, based on 25,600 collected samples.

The plots correspond well with the computed KL divergence scores. As the KL divergence values decrease, the histograms show greater alignment. Specifically, as the training prompt length for the LTV configurations increases, their density profiles become narrower, more closely resembling the shape of the vanilla transformer's distribution at \(T_{\text{train}}\). From a different perspective, this visual alignment supports our hypothesis once again: An optimized LTV with sufficiently long prompts performs near-optimally, as it effectively maintains the last hidden state distribution close to that of the model performing under \(T = T_{\text{train}}\).

\begin{figure*}[tph]
    \centering
    \begin{equation*}
        \text{\small {\coloredsquare{customdarkblue}} Vanilla transformer at \(T_\text{train}\)}  \qquad \text{\small {\coloredsquare{customdarkred}} Configuration at \(T_\text{max}\)}
    \end{equation*}
	\subfigure[]{
            \subfigure[\shortstack{\phantom{(a)} Transformer \\ \phantom{(a)} $D_\text{KL} = 1.098$}]{
                \includegraphics[width=0.19\linewidth]{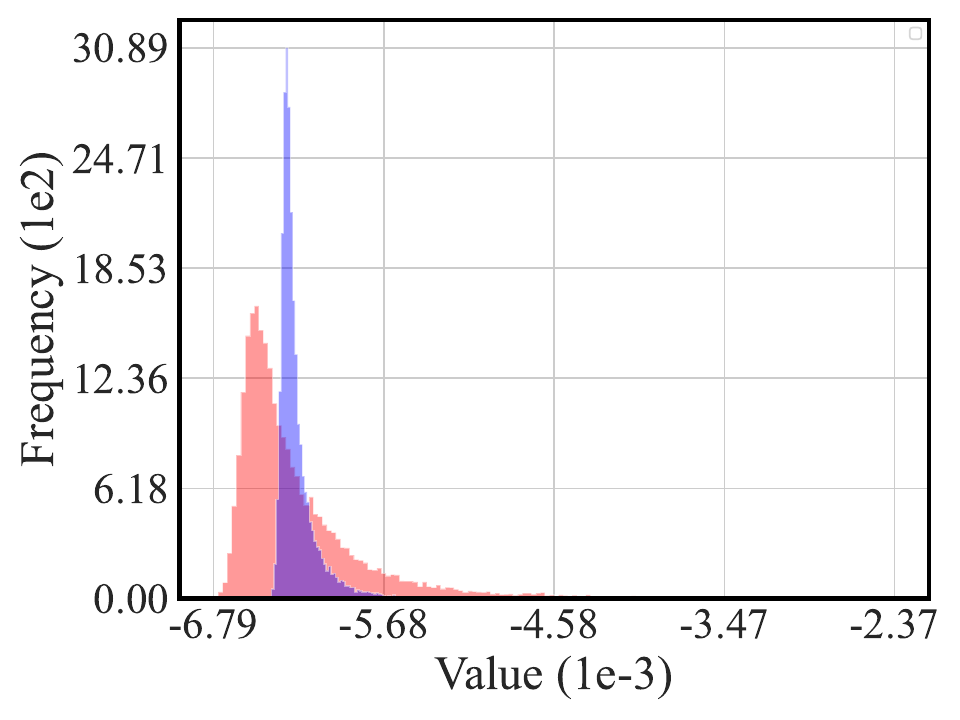}%
            }
            \subfigure[\shortstack{\phantom{(a)} Function Vector \\ \phantom{(a)} $D_\text{KL} = 0.646$}]{
                \includegraphics[width=0.19\linewidth]{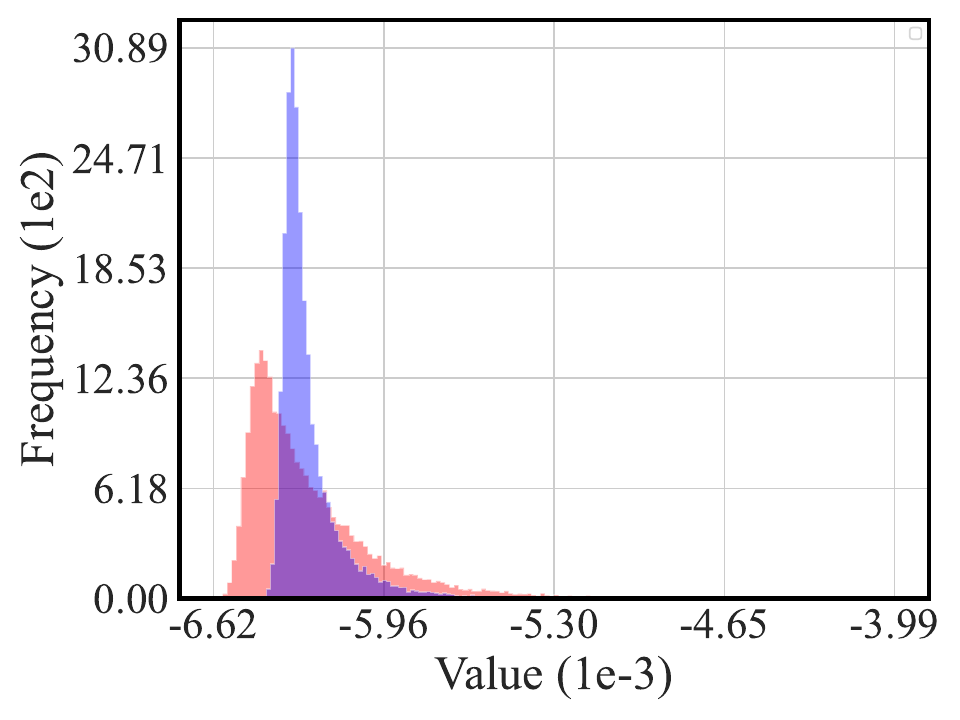}%
            }
            \subfigure[\shortstack{\phantom{(a)} LTV (\(T_\mathbf{v} = 41\)) \\ \phantom{(a)} $D_\text{KL} = 1.000$}]{
                \includegraphics[width=0.19\linewidth]{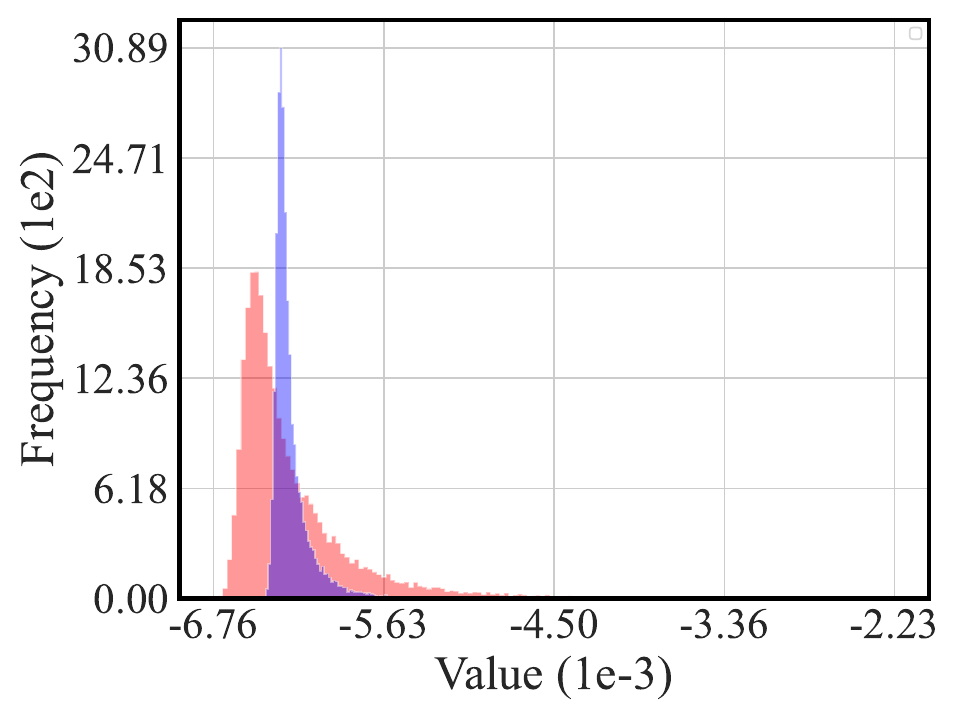}%
            }
            \subfigure[\shortstack{\phantom{(a)} LTV (\(T_\mathbf{v} = 42\)) \\ \phantom{(a)} $D_\text{KL} = 0.952$}]{
                \includegraphics[width=0.19\linewidth]{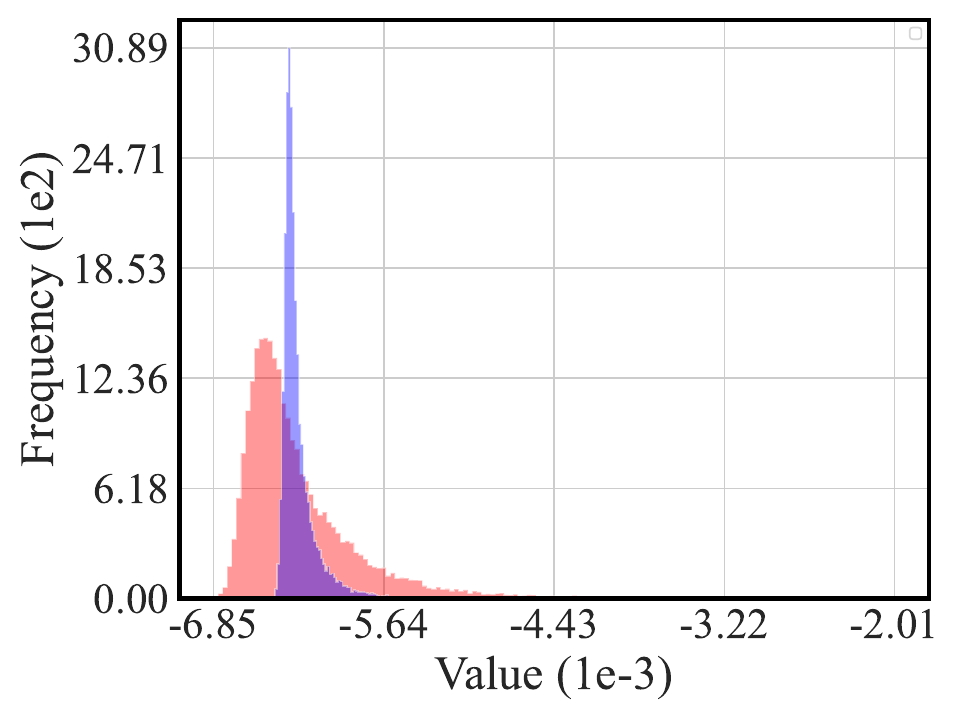}%
            }
        }
	\subfigure[]{
            \subfigure[\shortstack{\phantom{(a)} LTV (\(T_\mathbf{v} = 56\)) \\ \phantom{(a)} $D_\text{KL} = 0.532$}]{
                \includegraphics[width=0.19\linewidth]{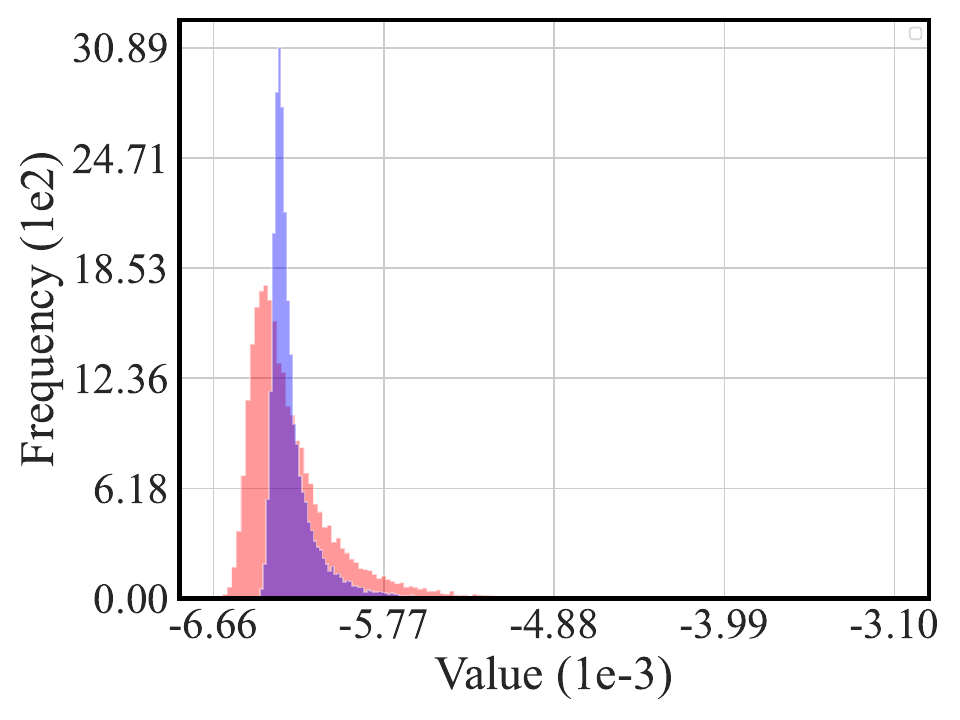}%
            }
            \subfigure[\shortstack{\phantom{(a)} LTV (\(T_\mathbf{v} = 71\)) \\ \phantom{(a)} $D_\text{KL} = 0.460$}]{
                \includegraphics[width=0.19\linewidth]{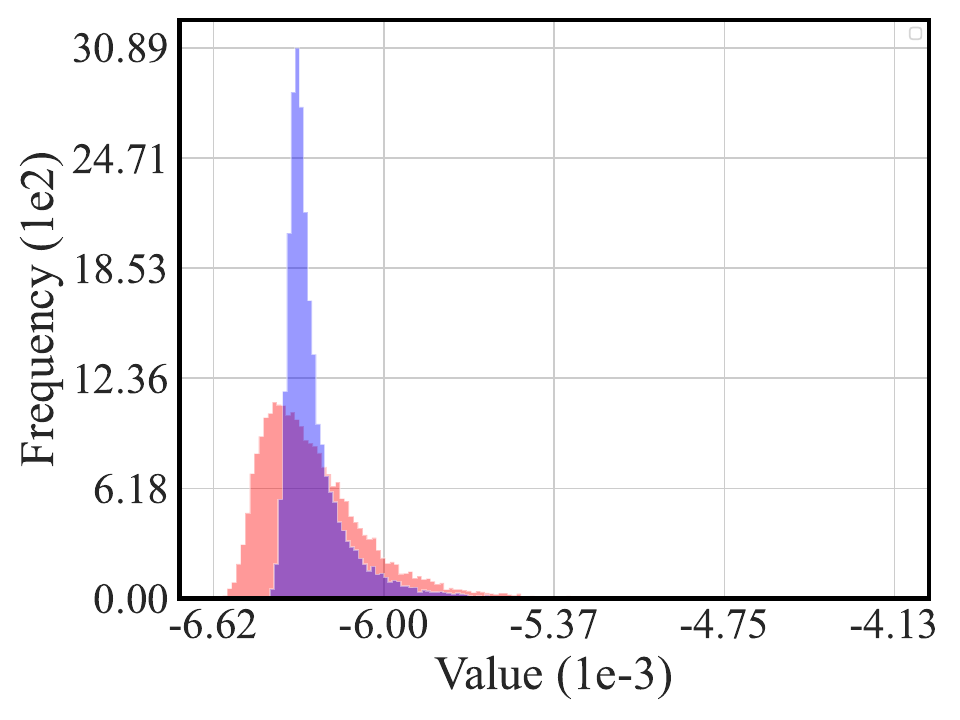}%
            }
            \subfigure[\shortstack{\phantom{(a)} LTV (\(T_\mathbf{v} = 86\)) \\ \phantom{(a)} $D_\text{KL} = 0.344$}]{
                \includegraphics[width=0.19\linewidth]{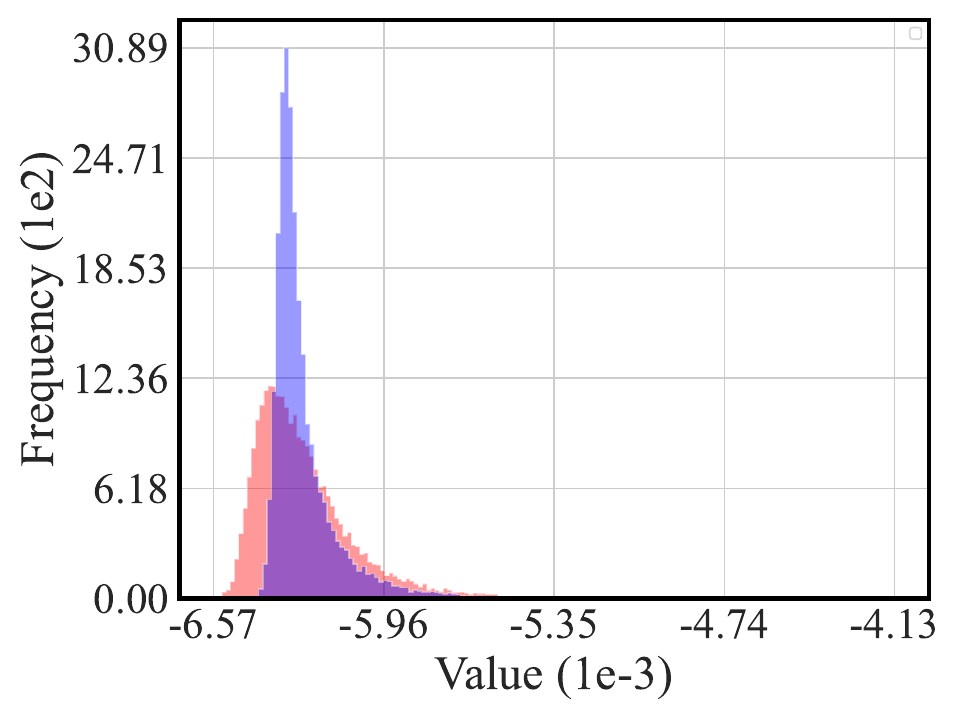}%
            }
            \subfigure[\shortstack{\phantom{(a)} LTV (\(T_\mathbf{v} = 101\)) \\ \phantom{(a)} $D_\text{KL} = 0.300$}]{
                \includegraphics[width=0.19\linewidth]{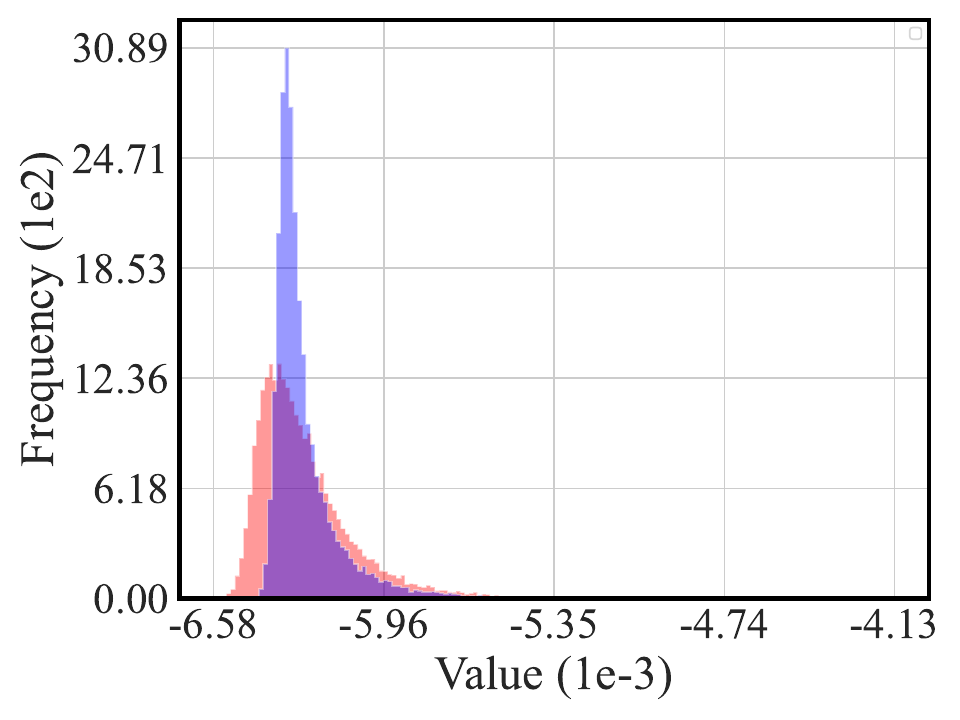}%
            }
        }
    \caption{Histograms of the empirical distribution of the last hidden states for the vanilla transformer collected at \(T_{\text{train}}\) and the tested configuration at the maximum length \(T_{\text{max}} = 101\) under linear functions. These histograms are generated using a dataset of 25,600 samples and correspond to the KL divergence scores reported in Table \ref{tab:ablations}.}
    \label{fig:ablation_histograms_lin_reg}
\end{figure*}

\begin{figure*}[tph]
    \centering
    \begin{equation*}
        \text{\small {\coloredsquare{customdarkblue}} Vanilla transformer at \(T_\text{train}\)}  \qquad \text{\small {\coloredsquare{customdarkred}} Configuration at \(T_\text{max}\)}
    \end{equation*}
	\subfigure[]{
            \subfigure[\shortstack{\phantom{(a)} Transformer \\ \phantom{(a)} $D_\text{KL} = 1.110$}]{
                \includegraphics[width=0.19\linewidth]{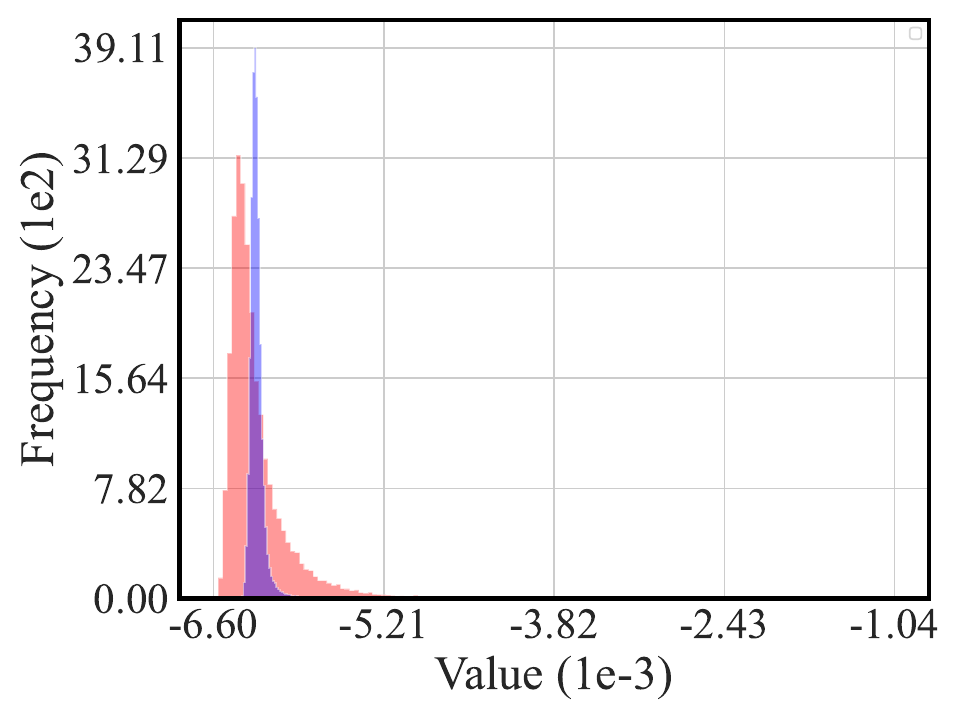}%
            }
            \subfigure[\shortstack{\phantom{(a)} Function Vector \\ \phantom{(a)} $D_\text{KL} = 0.572$}]{
                \includegraphics[width=0.19\linewidth]{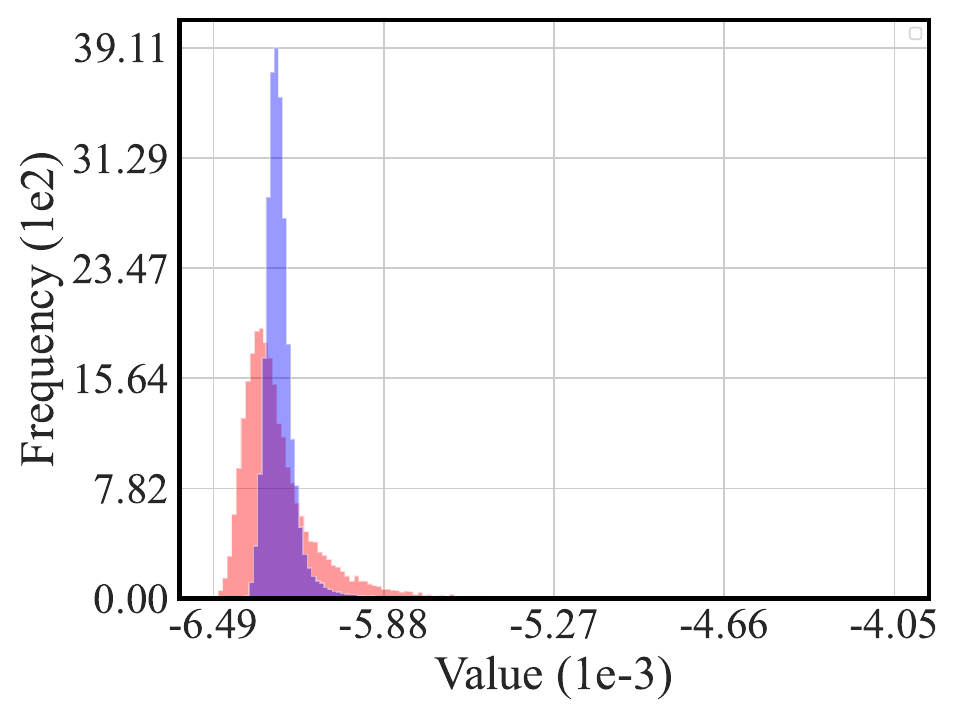}%
            }
            \subfigure[\shortstack{\phantom{(a)} LTV (\(T_\mathbf{v} = 41\)) \\ \phantom{(a)} $D_\text{KL} = 1.056$}]{
                \includegraphics[width=0.19\linewidth]{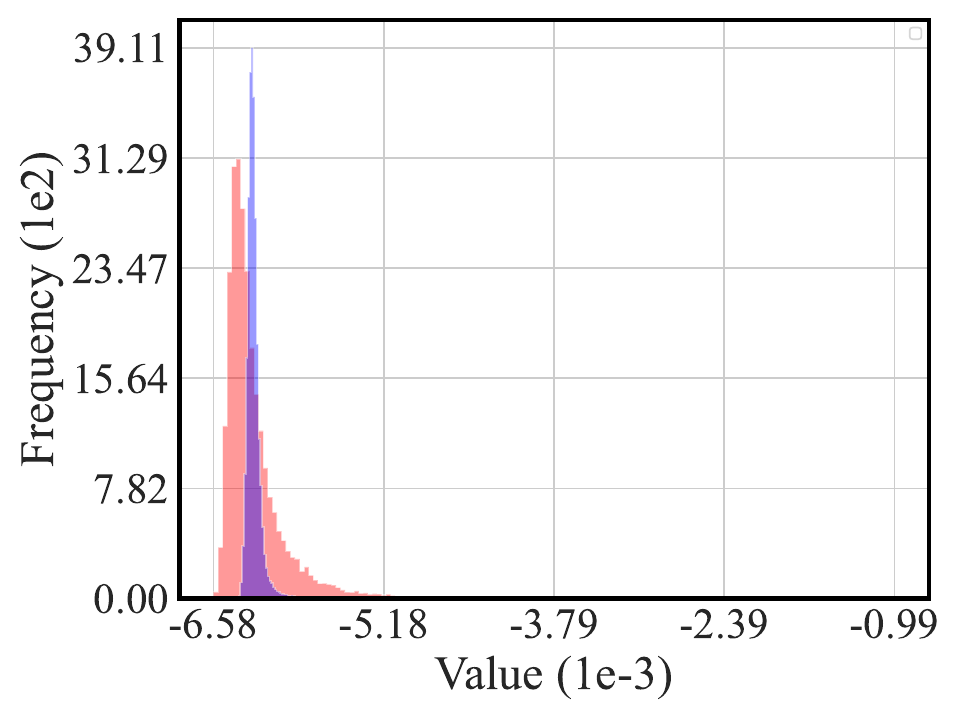}%
            }
            \subfigure[\shortstack{\phantom{(a)} LTV (\(T_\mathbf{v} = 42\)) \\ \phantom{(a)} $D_\text{KL} = 0.984$}]{
                \includegraphics[width=0.19\linewidth]{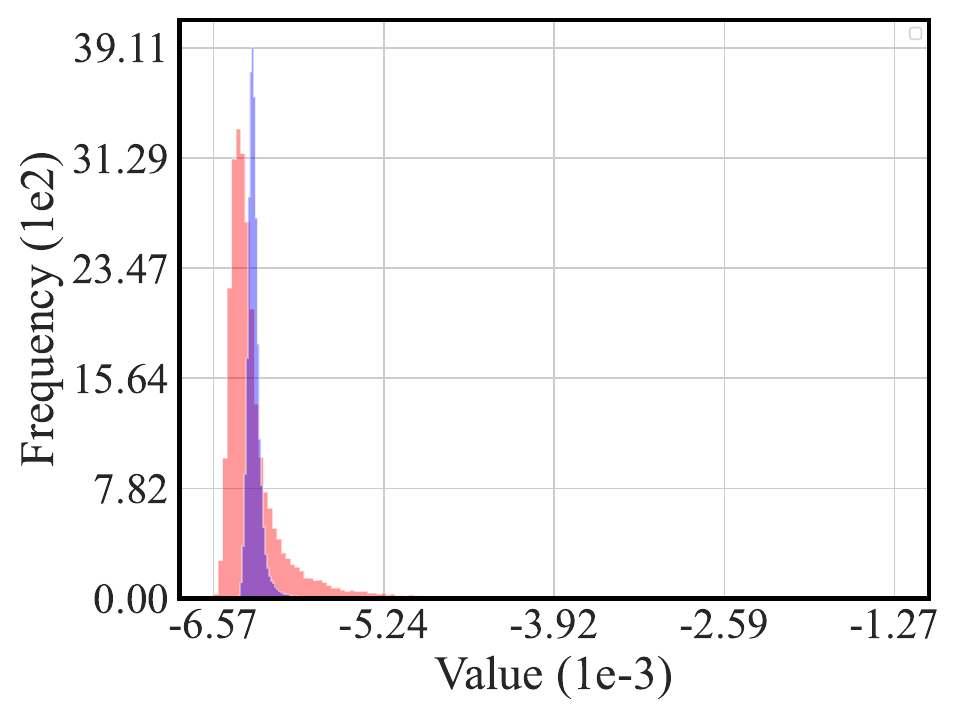}%
            }
        }
	\subfigure[]{
            \subfigure[\shortstack{\phantom{(a)} LTV (\(T_\mathbf{v} = 56\)) \\ \phantom{(a)} $D_\text{KL} = 0.260$}]{
                \includegraphics[width=0.19\linewidth]{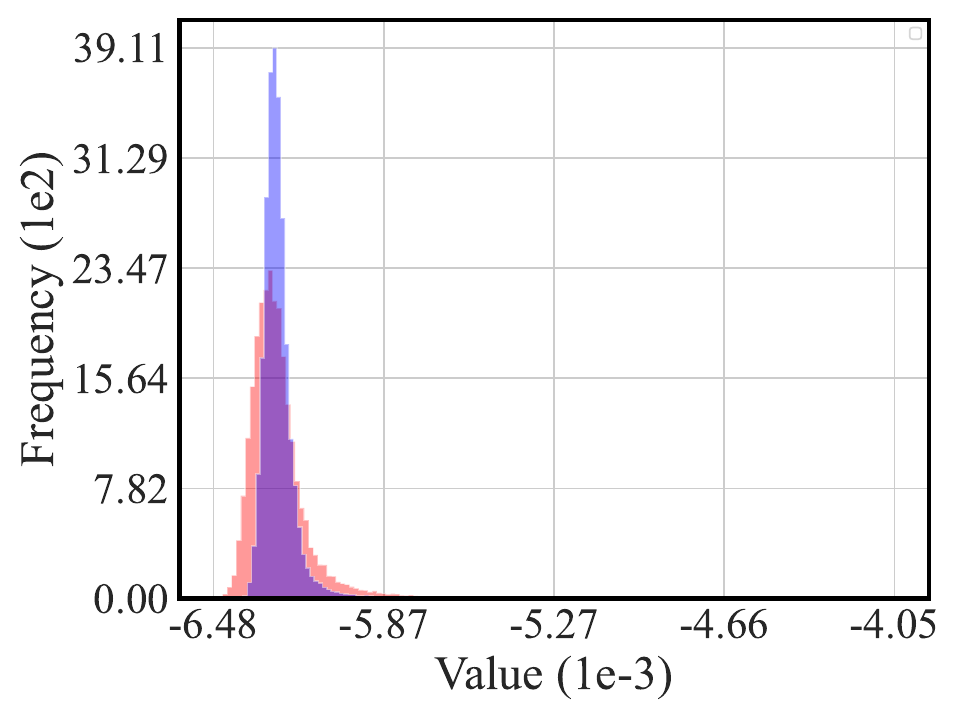}%
            }
            \subfigure[\shortstack{\phantom{(a)} LTV (\(T_\mathbf{v} = 71\)) \\ \phantom{(a)} $D_\text{KL} = 0.203$}]{
                \includegraphics[width=0.19\linewidth]{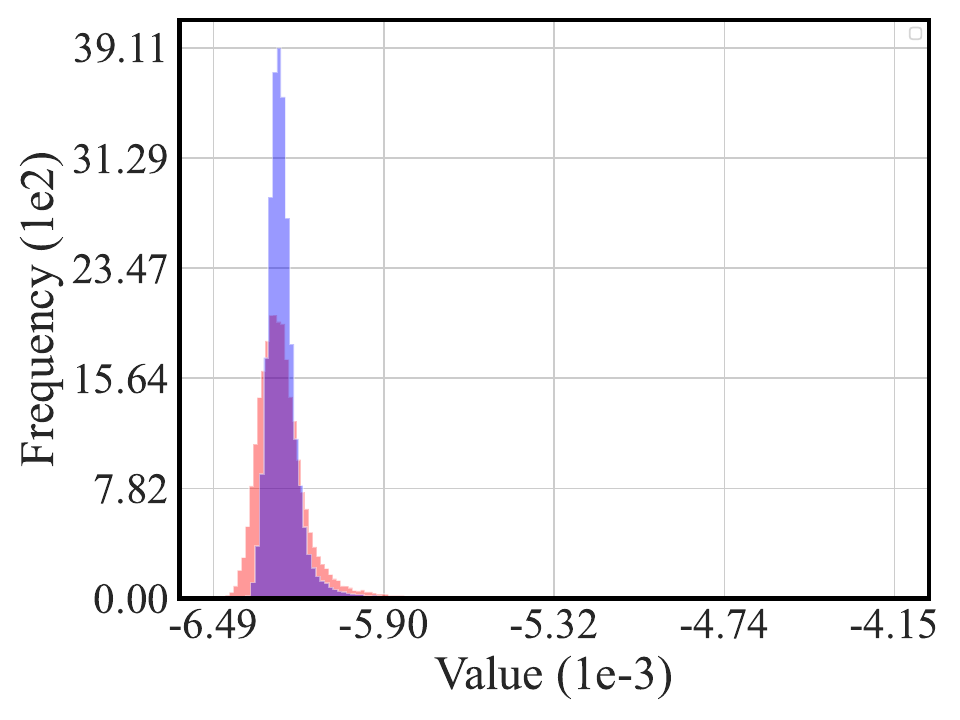}%
            }
            \subfigure[\shortstack{\phantom{(a)} LTV (\(T_\mathbf{v} = 86\)) \\ \phantom{(a)} $D_\text{KL} = 0.199$}]{
                \includegraphics[width=0.19\linewidth]{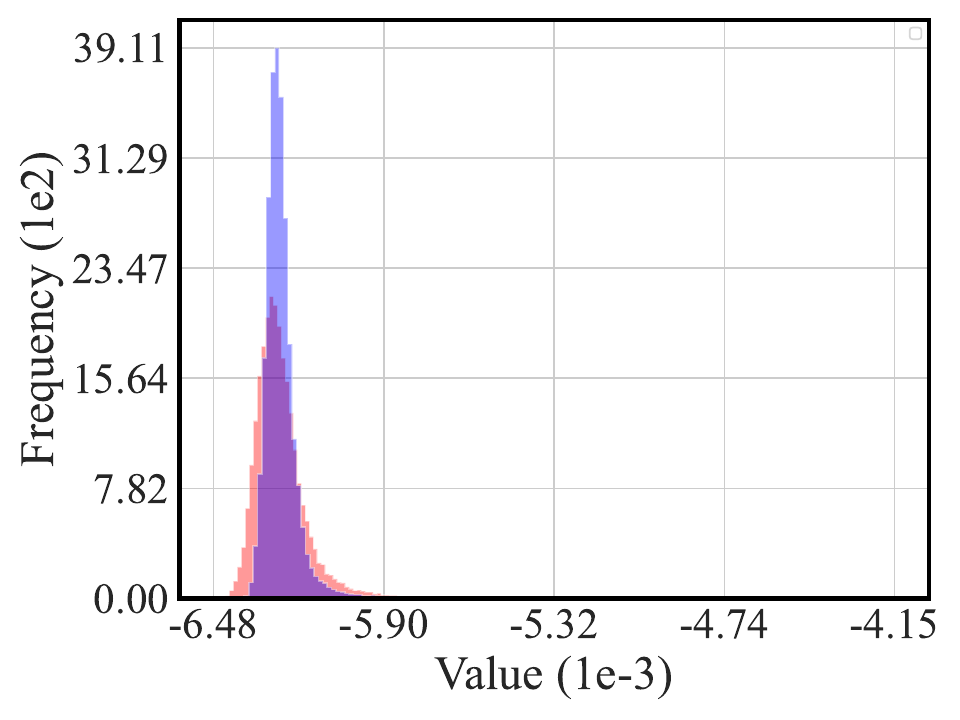}%
            }
            \subfigure[\shortstack{\phantom{(a)} LTV (\(T_\mathbf{v} = 101\)) \\ \phantom{(a)} $D_\text{KL} = 0.096$}]{
                \includegraphics[width=0.19\linewidth]{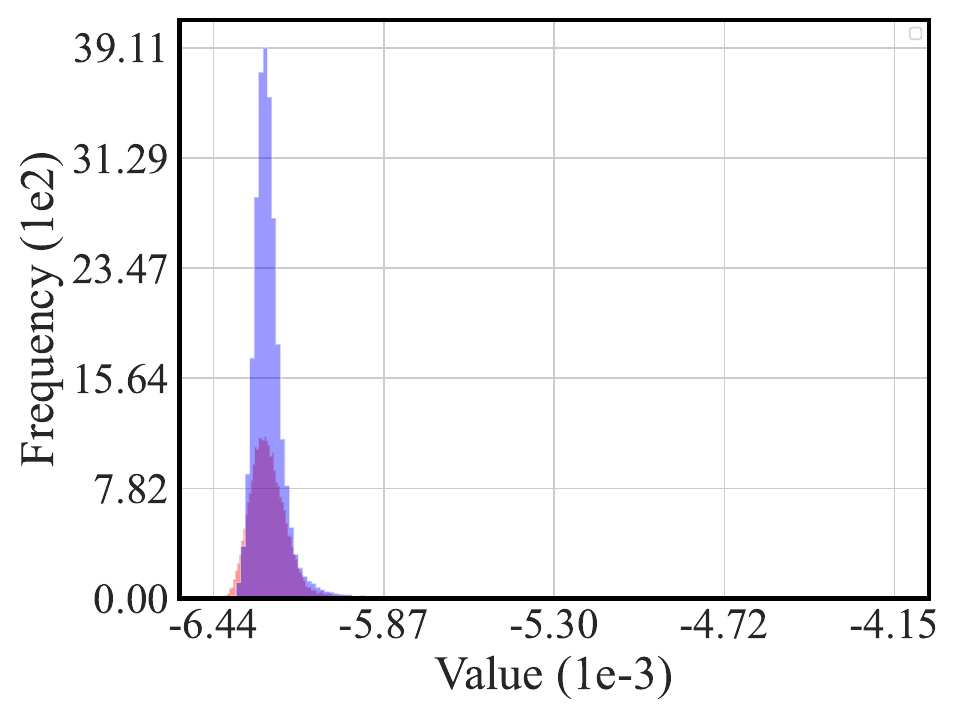}%
            }
        }
    \caption{Histograms of the empirical distribution of the last hidden states for the vanilla transformer collected at \(T_{\text{train}}\) and the tested configuration at the maximum length \(T_{\text{max}} = 101\) under sparse linear functions. These histograms are generated using a dataset of 25,600 samples and correspond to the KL divergence scores reported in Table \ref{tab:ablations}.}
    \label{fig:ablation_histograms_sparse_lin_reg}
\end{figure*}

\begin{figure*}[tph]
    \centering
    \begin{equation*}
        \text{\small {\coloredsquare{customdarkblue}} Vanilla transformer at \(T_\text{train}\)}  \qquad \text{\small {\coloredsquare{customdarkred}} Configuration at \(T_\text{max}\)}
    \end{equation*}
	\subfigure[]{
            \subfigure[\shortstack{\phantom{(a)} Transformer \\ \phantom{(a)} $D_\text{KL} = 0.103$}]{
                \includegraphics[width=0.19\linewidth]{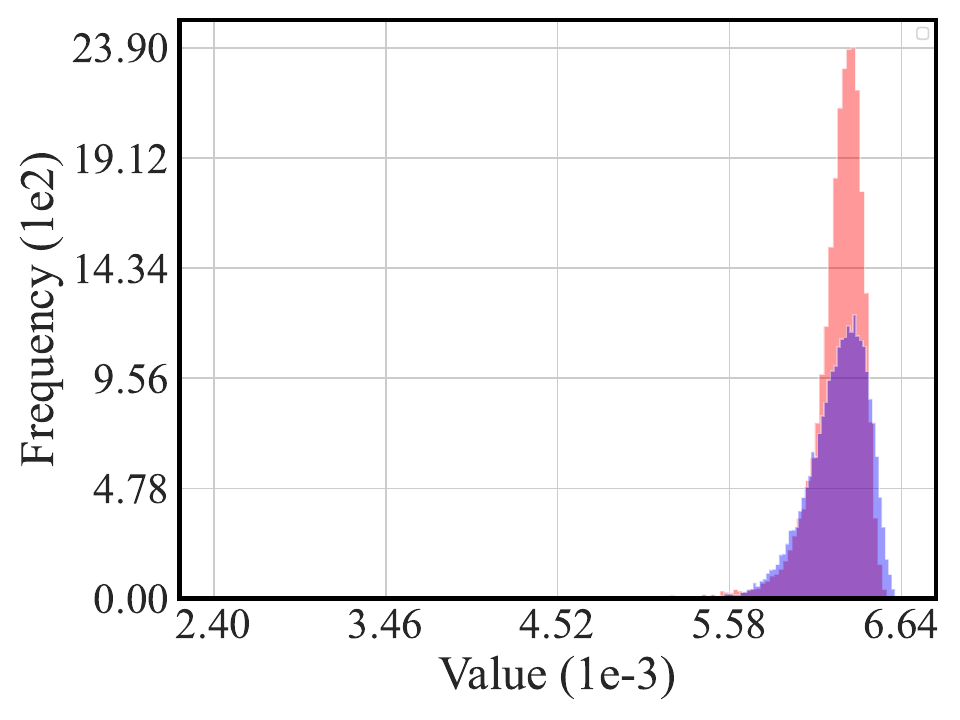}%
            }
            \subfigure[\shortstack{\phantom{(a)} Function Vector \\ \phantom{(a)} $D_\text{KL} = 0.223$}]{
                \includegraphics[width=0.19\linewidth]{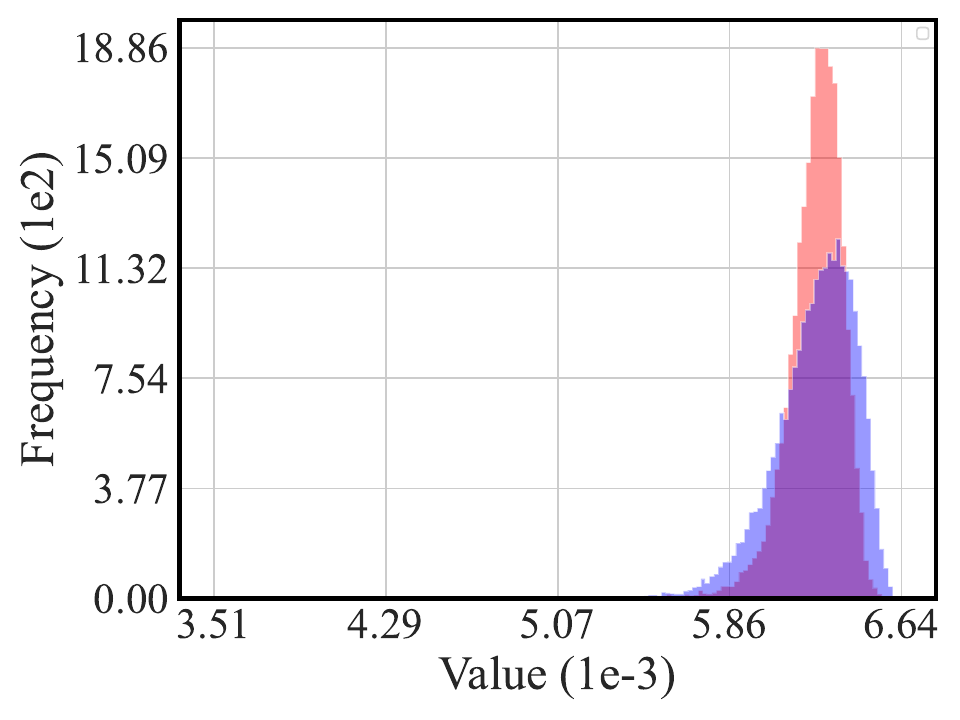}%
            }
            \subfigure[\shortstack{\phantom{(a)} LTV (\(T_\mathbf{v} = 101\)) \\ \phantom{(a)} $D_\text{KL} = 0.095$}]{
                \includegraphics[width=0.19\linewidth]{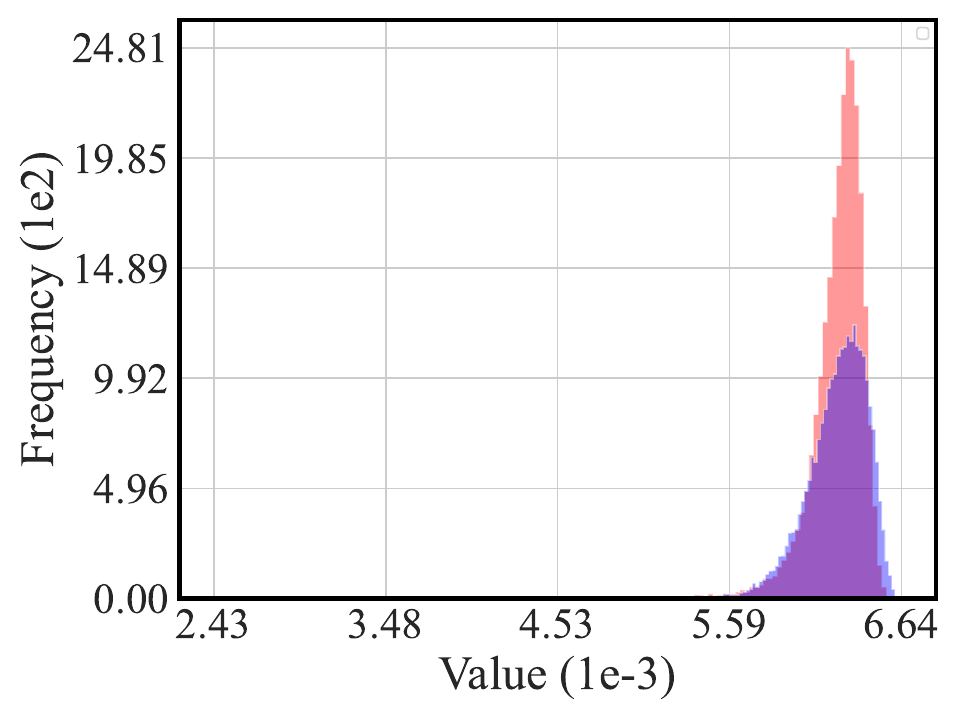}%
            }
            \subfigure[\shortstack{\phantom{(a)} LTV (\(T_\mathbf{v} = 102\)) \\ \phantom{(a)} $D_\text{KL} = 0.037$}]{
                \includegraphics[width=0.19\linewidth]{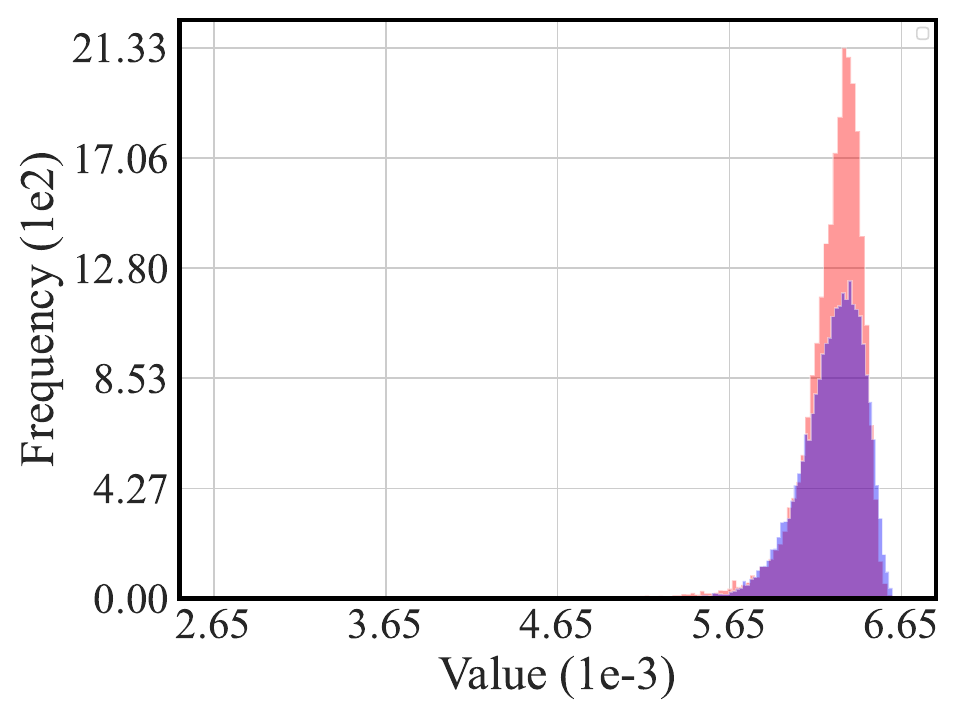}%
            }
        }
	\subfigure[]{
            \subfigure[\shortstack{\phantom{(a)} LTV (\(T_\mathbf{v} = 126\)) \\ \phantom{(a)} $D_\text{KL} = 0.016$}]{
                \includegraphics[width=0.19\linewidth]{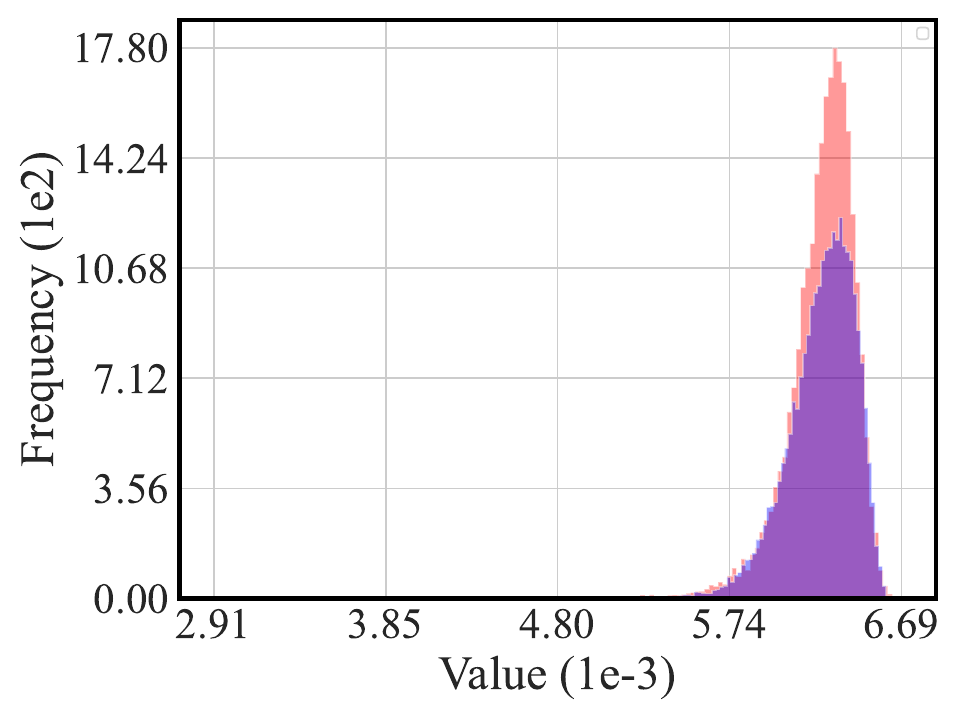}%
            }
            \subfigure[\shortstack{\phantom{(a)} LTV (\(T_\mathbf{v} = 151\)) \\ \phantom{(a)} $D_\text{KL} = 0.017$}]{
                \includegraphics[width=0.19\linewidth]{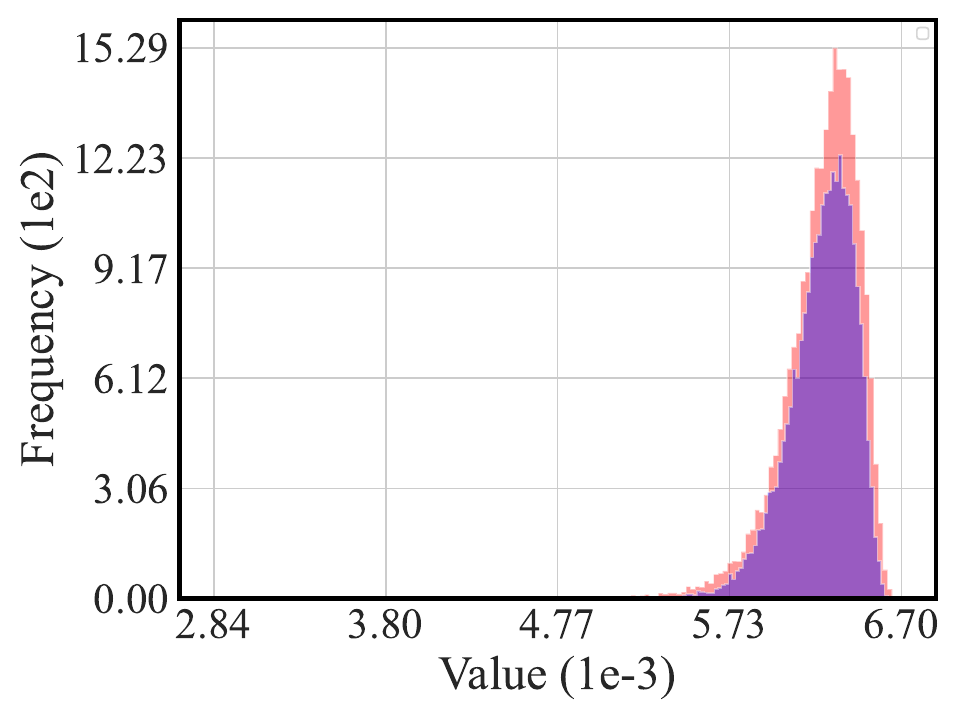}%
            }
            \subfigure[\shortstack{\phantom{(a)} LTV (\(T_\mathbf{v} = 176\)) \\ \phantom{(a)} $D_\text{KL} = 0.013$}]{
                \includegraphics[width=0.19\linewidth]{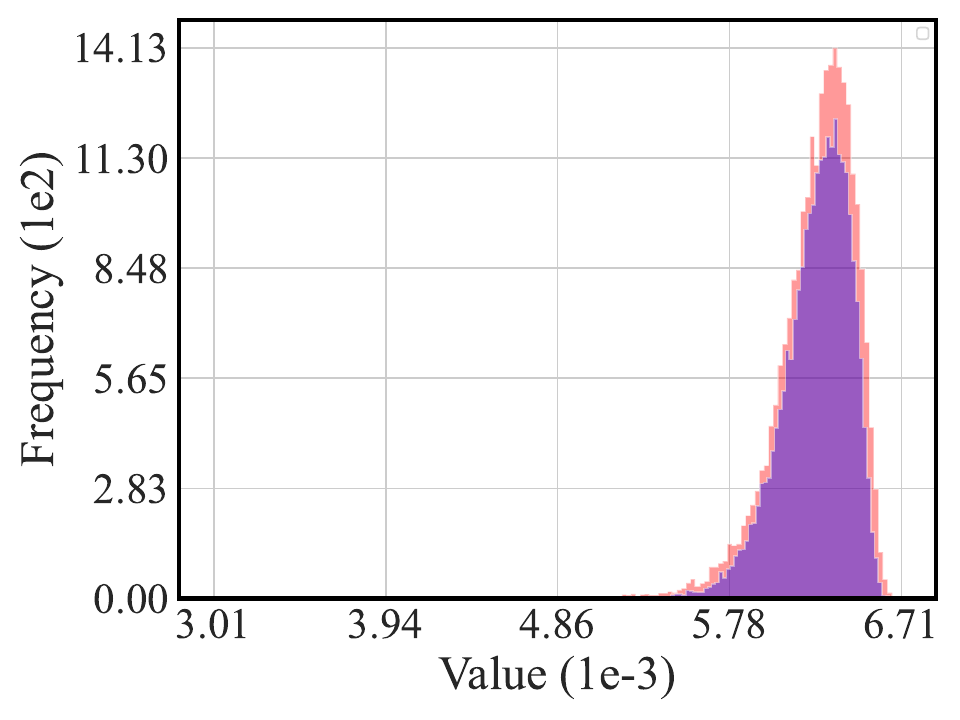}%
            }
            \subfigure[\shortstack{\phantom{(a)} LTV (\(T_\mathbf{v} = 201\)) \\ \phantom{(a)} $D_\text{KL} = 0.038$}]{
                \includegraphics[width=0.19\linewidth]{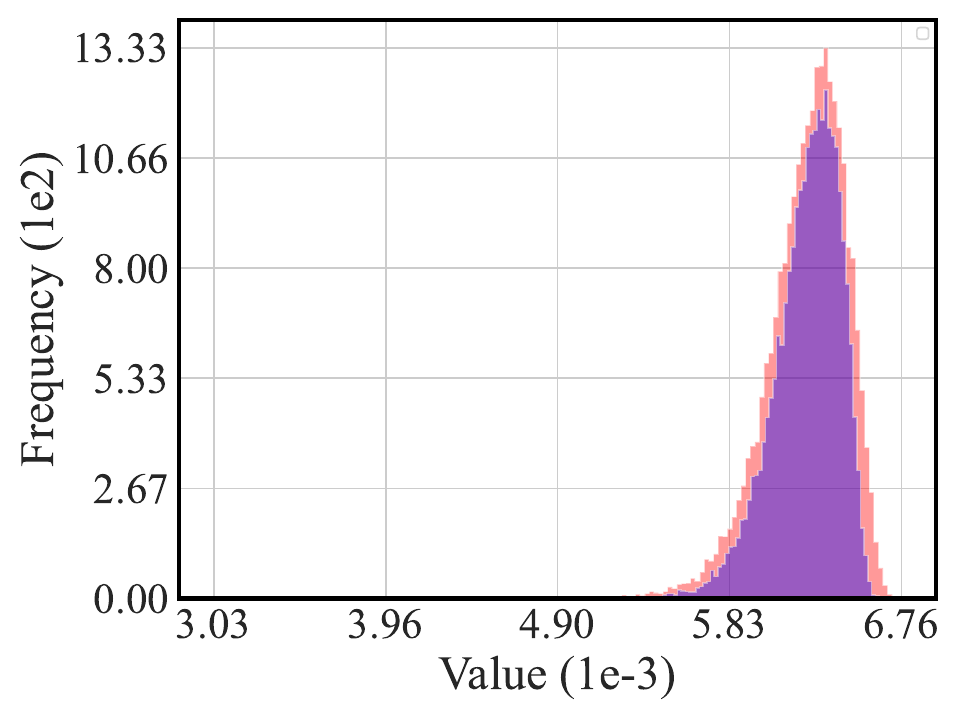}%
            }
        }
    \caption{Histograms of the empirical distribution of the last hidden states for the vanilla transformer collected at \(T_{\text{train}}\) and the tested configuration at the maximum length \(T_{\text{max}} = 201\) under 2-layer ReLU neural networks. These histograms are generated using a dataset of 25,600 samples and correspond to the KL divergence scores reported in Table \ref{tab:ablations}.}
    \label{fig:ablation_histograms_relu_nets}
\end{figure*}

\end{document}